\documentclass{article}

\usepackage[T1]{fontenc}       % Use modern font encoding
\usepackage[utf8]{inputenc}    % Allow UTF-8 input
\usepackage[english]{babel}    % Language
\usepackage{lmodern}           % Latin Modern fonts (fixes cmtt OMS warnings)
\usepackage[margin=1in]{geometry} % Page layout

\usepackage{amsmath,amssymb,amsthm}

\usepackage{makecell}
\usepackage{graphicx}
\usepackage{booktabs}
\usepackage{tabularx}
\usepackage{nicefrac}
\usepackage{array} 
\usepackage[ruled,vlined,linesnumbered]{algorithm2e}
\usepackage{tikz}
\usepackage{siunitx}
\usetikzlibrary{arrows.meta, positioning}
\usepackage{longtable}
\usepackage{ltablex}
\usepackage{xcolor}
\usepackage{soul}
\sethlcolor{yellow!35}
\keepXColumns 
\usepackage{tabularx,ragged2e,booktabs,xcolor,soul}
\sethlcolor{yellow!35}
\newcolumntype{P}{>{\RaggedRight\arraybackslash\hspace{0pt}}p{0.485\linewidth}}
\usepackage{multirow}

\usepackage{arydshln}

\usepackage{array,ragged2e,booktabs,makecell}
\newcolumntype{L}[1]{>{\RaggedRight\arraybackslash}m{#1}}
\SetKwComment{tcc}{\scriptsize/* }{ */}
\SetKwComment{tcp}{\scriptsize// }{}
\RestyleAlgo{ruled}                  % keep the ruled style
\usepackage[authoryear, round]{natbib}

\usepackage{hyperref}
\usepackage{fancyvrb}
\usepackage{newunicodechar}
\newunicodechar{≈}{\approx}
\usepackage{enumitem}
\usepackage{listings}
\usepackage{array} 
\usepackage{ragged2e}
\usepackage{float}

\usepackage[dvipsnames]{xcolor}     % single consolidated xcolor load
\usepackage[most]{tcolorbox} % loads all common tcolorbox libraries
\usepackage{xcolor}          % for named colors

\usepackage{listings}
\usepackage{listingsutf8}
\usepackage{tcolorbox}
\usepackage{upquote}

\definecolor{RoyalBlue}{RGB}{65,105,225}
\definecolor{mygray}{rgb}{0.96,0.96,0.96}
\definecolor{CodeStr}{RGB}{78,154,6}      % dark green strings
\definecolor{CodeKw}{RGB}{52,101,164}     % blue keywords
\definecolor{CodeCom}{RGB}{143,89,2}      % brown comments
\definecolor{CodeNum}{RGB}{196,63,63}     % red numbers (contrast)
\definecolor{CodeId}{RGB}{32,74,135}      % identifiers

\lstdefinestyle{mintlike}{
  basicstyle=\ttfamily\small,
  backgroundcolor=\color{mygray},
  showstringspaces=false,
  breaklines=true,
  columns=fullflexible,
  keepspaces=true,
  upquote=true,
  keywordstyle=\color{CodeKw}\bfseries,
  commentstyle=\itshape\color{CodeCom},
  stringstyle=\color{CodeStr},
  identifierstyle=\color{CodeId},
  frame=none,
  numbers=none,
  literate=
   *{0}{{{\color{CodeNum}0}}}{1}
    {1}{{{\color{CodeNum}1}}}{1}
    {2}{{{\color{CodeNum}2}}}{1}
    {3}{{{\color{CodeNum}3}}}{1}
    {4}{{{\color{CodeNum}4}}}{1}
    {5}{{{\color{CodeNum}5}}}{1}
    {6}{{{\color{CodeNum}6}}}{1}
    {7}{{{\color{CodeNum}7}}}{1}
    {8}{{{\color{CodeNum}8}}}{1}
    {9}{{{\color{CodeNum}9}}}{1},
}

\lstdefinelanguage{json}{
  morestring=[b]",
  stringstyle=\color{CodeStr},
  morekeywords={true,false,null},
  keywordstyle=\color{CodeKw}\bfseries,
  comment=[l]{//},
  commentstyle=\itshape\color{CodeCom},
  literate=
   *{:}{{:}}{1}
    {,}{{,}}{1}
    {\{}{{\{}}{1}
    {\}}{{\}}}{1}
    {[}{{[}}{1}
    {]}{{]}}{1}
    {0}{{{\color{CodeNum}0}}}{1}
    {1}{{{\color{CodeNum}1}}}{1}
    {2}{{{\color{CodeNum}2}}}{1}
    {3}{{{\color{CodeNum}3}}}{1}
    {4}{{{\color{CodeNum}4}}}{1}
    {5}{{{\color{CodeNum}5}}}{1}
    {6}{{{\color{CodeNum}6}}}{1}
    {7}{{{\color{CodeNum}7}}}{1}
    {8}{{{\color{CodeNum}8}}}{1}
    {9}{{{\color{CodeNum}9}}}{1},
}

\newtcblisting{pythoncode}{
  listing engine=listings,
  listing only,
  boxrule=0.6pt,
  colback=mygray, colframe=RoyalBlue,
  left=5pt,right=5pt,top=3pt,bottom=3pt,
  enhanced, sharp corners,
  listing options={language=Python, style=mintlike}
}

\newtcblisting{jsoncode}{
  listing engine=listings,
  listing only,
  boxrule=0.6pt,
  colback=mygray, colframe=RoyalBlue,
  left=5pt,right=5pt,top=3pt,bottom=3pt,
  enhanced, sharp corners,
  listing options={language=json, style=mintlike}
}

\newtcblisting{textcode}{
  listing engine=listings,
  listing only,
  boxrule=0.5pt,
  colback=mygray, colframe=RoyalBlue,
  left=1mm, right=1mm, top=1mm, bottom=1mm,
  arc=2pt, enhanced, sharp corners,
  listing options={
    style=mintlike,
    language=,          % plain text (no lexer)
    escapeinside=||,    % enable |...| LaTeX escapes
    basicstyle=\ttfamily\small
  }
}

\title{TimeStampEval: A Simple LLM Eval and a Little Fuzzy Matching Trick to Improve Search Accuracy}
\author{James McCammon\\
\small Independent Researcher\\
\small\texttt{james.j.mccammon@gmail.com}}
\date{October 27, 2025}

\begin{document}

\maketitle

\begin{abstract}
Traditional fuzzy matching fails when searching for quotes that appear semantically identical but syntactically different across documents—a common problem when matching official written records against speech-to-text transcriptions. We introduce TimeStampEval, a benchmark for retrieving precise millisecond timestamps from long transcripts when given potentially non-verbatim quotes, and demonstrate a simple two-stage approach that dramatically improves accuracy while cutting costs by over 90\%.

Our motivating use case is an automated long-form podcast that stitches Congressional Record clips into AI-hosted narration, a sample episode of which is available on \href{https://open.spotify.com/episode/5SPtYgWfY8yGhr7Un0qBPL}{Spotify}. The core technical challenge: given a sentence-timestamped transcript and a target quote that may have syntactic drift from transcription errors or editorial changes, return the exact \texttt{start\_ms}/\texttt{end\_ms} boundaries. While standard algorithms handle verbatim matching perfectly, real production systems face fuzzy variants that break traditional approaches.

We evaluated six modern LLMs on a 2,800-sentence Congressional Record transcript (120k tokens), finding four key results from our full run. First, simple prompt engineering dominates model selection: placing the query before the transcript and using a compact text-first format improved accuracy by 3–20 percentage points while reducing tokens by 30–42\%. Second, off-by-one errors form a distinct regime between complete failures and exact matches, suggesting models understand the task but struggle with precise boundaries. Third, a modest reasoning budget (600–850 tokens) rescues weak configurations from ~37\% to ~77\% accuracy and pushes strong ones to mid-90s. Fourth, our two-stage ``Assisted Fuzzy'' approach—using RapidFuzz to narrow candidates then LLM verification on short snippets—improved fuzzy matching accuracy by up to 50 percentage points while reducing latency by half and cost-per-correct by 87–96\%.

A focused deep dive with Gemini Flash across ten transcripts (50k–900k tokens, spanning 1989–2025) confirmed three patterns: (1) the method scales robustly to ~400k tokens with gradual degradation approaching one million, (2) accuracy remains stable across decades despite vocabulary and speaking style changes, and (3) the system correctly rejects absent targets with 95–100\% accuracy, a critical production safeguard. The resulting recipe is practical: for verbatim quotes, skip LLMs entirely; for fuzzy matches, pre-narrow with traditional fuzzing then verify with an LLM on just the relevant snippet.
\end{abstract}

\section{Summary of Results}
\label{sec:summary-of-results}

% --- Run definitions shown above the table ---
\subsection{Experimental setups used in this paper}
A total of three separate experimental setups were used in this paper.

\begin{enumerate}[leftmargin=1.5em, itemsep=8pt, topsep=2pt]
  \item \textbf{Full Run} —  
  This was the primary experiment used for cross-model comparison. It covered six models in total, drawn from both the Gemini and GPT-5 families. Claude was excluded at this stage due to cost constraints. The design built directly on insights from the Flash deep dive: it focused on query placement, span length, and transcript thirds while standardizing on the two most informative transcript formats (JSON and Text First Top). The full run represents the most balanced and comprehensive evaluation in the paper, combining accuracy, efficiency, and cost. Each model completed 540 trials per condition, enabling direct comparison across providers and sizes.  
  \begin{itemize}[leftmargin=2em, itemsep=2pt, topsep=2pt]
    \item Gemini-2.5 Flash
    \item Gemini-2.5 Flash-Lite
    \item Gemini-2.5-Pro
    \item GPT-5
    \item GPT-5-mini
    \item GPT-5-nano
  \end{itemize}

  \item \textbf{Pilot} —  
  This was an early exploratory run conducted midway through the project, before GPT-5 had been released. The pilot included a broader but less systematically controlled set of models, covering multiple Gemini, GPT-4.1, Claude, and O-series variants. The design was lighter: JSON transcripts with query-after placement, no reasoning tokens enabled, and smaller sample sizes (e.g., $n=9$ per length for exact match, $n=18$ for fuzzy). While limited in generalization, the pilot was essential for refining the experimental setup. It revealed several patterns that motivated the more structured deep dive and full run: query-last inflates difficulty and variance, last-third targets are disproportionately hard, off-by-one behaves as a transitional error class, and the cost–accuracy frontier separates into ``value'' models versus ``ceiling'' models. In short, the pilot served as a bridge—broad but noisy—between initial exploration and the larger-scale evaluations that anchor the paper.  
  \begin{itemize}[leftmargin=2em, itemsep=2pt, topsep=2pt]
    \item Gemini-2.5-Pro
    \item Gemini-2.5 Flash
    \item Gemini-2.5 Flash-Lite (Preview 06-17)
    \item GPT-4.1
    \item GPT-4.1 mini
    \item GPT-4.1 nano 
    \item Claude 3.5 Haiku
    \item Claude 3.7 Sonnet
    \item Claude Sonnet 4.0
    \item o3
    \item o3-mini
    \item o4-mini
  \end{itemize}

  \item \textbf{Flash Deep Dive} —  
  This was an intermediate run conducted with Gemini 2.5 Flash only. The purpose was to explore the parameter space in detail and generate hypotheses for the larger experiments that followed. In this run we systematically varied target placement across transcript thirds, query position (before vs.\ after), and span length $\ell \in \{1,2,3,5,10\}$. We also measured efficiency variables such as latency and reasoning token usage. Beyond these placement and length manipulations, the deep dive extended into long-context testing, with inputs ranging from 100k up to 900k tokens, and explored five different text formats. These included the standard JSON transcript, Text First Top, and several alternative layouts that shifted the ordering of sentence, start\_ms, and end\_ms markers. The deep dive provided the first clear evidence that input format and token length strongly affect retrieval performance, shaping the hypotheses carried forward into the full run.  
  \begin{itemize}[leftmargin=2em, itemsep=2pt, topsep=2pt]
    \item Gemini-2.5 Flash
  \end{itemize}
  
\end{enumerate}
\vspace{0.25em}

\subsection{Glossary of terms}
\begin{table}[H]
\centering
\small
\setlength{\tabcolsep}{6pt}
\renewcommand{\arraystretch}{1.15}
\begin{tabular}{@{}P P@{}}
\toprule
\textbf{Term} & \textbf{What I mean (plain English)} \\
\midrule
Query & The part of model prompt instruction that describes which target is being searched for in the transcript. \\ \midrule
Q↑ / Q↓ & Shorthand for query placement: Q↑ = query before the transcript; Q↓ = query after the transcript. \\ \midrule
Text First Top (TFT) & Transcript layout that puts the sentence \emph{before} its timestamps and the query \emph{before} the transcript. \\ \midrule
JSON & Native speech-to-text (STT) format from the provider (AssemblyAI); JSON format with the start and end timestamps for each sentence spoken, the order is \texttt{start\_ms}, \texttt{end\_ms}, \texttt{sentence}. \\ \midrule
Off-by-one & When the model returns start or end millisecond timestamps that are shifted by exactly one sentence from the true target. \\ \midrule
Adjusted accuracy & Accuracy that counts off-by-one as “good enough.” \\ \midrule
Span length (\(\ell\)) & Number of consecutive sentences (\(\ell\in\{1,2,3,5,10\}\)). \\ \midrule
Transcript thirds & First / middle / last third of the transcript. \\ \midrule
Exact match & Predicted \texttt{start\_ms}/\texttt{end\_ms} exactly equal the reference pair. \\ \midrule
Fuzzy match & The model searches the transcript for a quote that does not appear verbatim in the transcript, for example the query may include ``U.S.'' while the target spells it out as ``United States''; in the benchmark we use synthetically generated perturbed queries to assess fuzzy matching accuracy. \\ \midrule
Unassisted Fuzzy & Model searches the whole transcript for a perturbed quote. \\ \midrule
Assisted Fuzzy (hybrid) & RapidFuzz algorithm first narrows down to a short snippet; the model then returns timestamps from that window. \\ \midrule
Snippet control & Exact-match using dynamic snippets only (isolate context-length effects). \\ \midrule
Cost per correct (CPC) & Dollars spent divided by the number of correct timestamp returns. \\ \midrule
Label-free TFT & TFT with bare numbers for \texttt{start\_ms}/\texttt{end\_ms}. \\ \midrule
Long-context buckets & Concatenated transcripts of length 100k--900k tokens to test robustness. \\ \midrule
\bottomrule
\end{tabular}
\caption{Quick glossary for the Summary of Results.}
\label{tab:summary-glossary}
\end{table}

\subsection{Main findings}
Below is a summary of main findings.

\begin{enumerate}[leftmargin=1.5em, itemsep=8pt, topsep=2pt]

  \item \textbf{Placing the query before the transcript improves matching accuracy.}
  In the JSON condition, placing the query before the transcript outperformed placing it after by roughly 14–16 percentage points on 2–5 sentence spans and by about 3 points on single sentences; the 10‑sentence span also gained about 12.5 points. Average latency dropped from 21.2\,s (query after) to 14.4\,s (query before), with Text First Top at 13.7\,s. Median reasoning tokens were lower when the query came first and scaled with span length; when the query came after, medians were higher and largely flat with span length. See Table~\ref{tab:exact-timestamp-by-length}, Table~\ref{tab:exact-match-per-model-latency}, and \S\ref{sec:timestamps}.

  \item \textbf{Transforming the native STT JSON reduced tokens by 30–42\% and improved accuracy.}
  Converting the provider’s JSON into a text format that puts the sentence before its timestamps cut prompt tokens by about 30\%, and removing field labels cut by about 42\%. Across the exact‑match baselines this transformation improved accuracy by roughly 1–15 percentage points and, critically, collapsed off‑by‑one near misses: Text First Top showed about a 5\% off‑by‑one rate versus 12–16\% in JSON with the query after. On the needle controls, text formats were 2.7–2.9$\times$ faster than JSON at essentially the same accuracy. See \S\ref{sec:text-field-order-variations}, Table~\ref{tab:gemini-flash-json-vs-text-results}, Table~\ref{tab:offby1-by-format}, and Table~\ref{tab:latency}.

  \item \textbf{Unassisted fuzzy matched below exact by 4–10 points; the assisted fuzzy pipeline added 4–50 points, halved latency, and cut cost per correct by at least $10\times$.}
  Without assistance, fuzzy matching lagged exact by 4–10 points on average, with the largest drops in smaller models. Adding a simple RapidFuzz step to carve a short snippet and then sending only that snippet to the model increased accuracy by 4–50 points in the main run (and up to 70–90 points in the pilot), reduced average latency from about 16.1\,s to about 8.1\,s, and lowered dollars‑per‑correct from \$0.0547 to \$0.0045 on average. The savings came from sending about 99\% fewer input sentences and from fewer reasoning tokens. See Table~\ref{tab:opttext-exact-vs-fuzzy}, Table~\ref{tab:fuzzy-vs-assisted-fuzzy}, Table~\ref{tab:pilot-fuzzy-assisted}, Table~\ref{tab:latency-fuzzy-assisted-seconds}, and Table~\ref{tab:new-run-cost-fuzzy-assisted-pct}.

  \item \textbf{Off‑by‑one concentrated under JSON with the query after and shrank under Text First Top with the query before.}
  JSON with the query after produced the highest off‑by‑one rates, while Text First Top with the query before reduced them to about 5\%. As model size increased, errors moved along a clear ladder from miss to off‑by‑one to exact. See Table~\ref{tab:json-text-error-breakdown}, Table~\ref{tab:offby1-by-format}, and Figure~\ref{fig:offbyone-transition}.

  \item \textbf{Enabling model reasoning raised the worst layout from about 37\% to about 72–77\% exact and nudged the best layout to about 95–96\%.}
  Allowing thinking on the weakest prompt configuration (text after timestamps, query after) lifted exact‑match accuracy from roughly 37\% to roughly 72–77\%. On the strongest configuration (Text First Top with the query before), accuracy rose from about 89\% to about 95–96\% (near ceiling). Typical reasoning usage was small—about 600–850 tokens—which kept the cost impact negligible. See Table~\ref{tab:budget-sweep} and \S\ref{sec:thinking-budget}.

  \item \textbf{Span length followed an inverted‑U: 2–5 sentences were easiest; 1 and 10 dipped, most visibly in smaller models.}
  The inverted‑U shape was consistent across formats and providers, with the dip at 1 and 10 sentences strongest in smaller models; larger models were closer to ceiling across lengths. See Table~\ref{tab:exact-timestamp-by-length} and the per‑model tables that follow.

  \item \textbf{Transcript position effects varied by setup: the full run showed the last third as lowest; a ten‑transcript sweep with Text First Top showed small gaps.}
  In the full run, the last third trailed the first and middle across formats (for example, Text First Top averaged 79.2\% in the last third versus 88.1\% in the first). In a separate sweep over ten Congressional Record days using Text First Top, accuracy remained high across thirds (about 99\% first, 95.4\% middle, 97.0\% last) with tight confidence intervals. See Table~\ref{tab:thirds-accuracy} and \S\ref{sec:gemini-flash-confidence-intervals}.

  \item \textbf{Latency and tokens scaled with task complexity; Gemini was faster on average; OpenAI was higher on fuzzy accuracy on average.}
  Gemini models were consistently faster and scaled latency more predictably (for example, Flash at roughly 5–7\,s on many exact tasks), while OpenAI models ran slower but averaged about 8.5 points higher accuracy on the fuzzy tasks in the full run. Fuzzy matching increased median reasoning tokens by about 25\% overall, with the steepest increases in Flash‑Lite. See Table~\ref{tab:exact-match-per-model-latency}, Table~\ref{tab:opt-vs-fuzzy}, Table~\ref{tab:thinking-tokens-exact-vs-fuzzy-median}, and \S\ref{sec:fuzzy}.

  \item \textbf{Accuracy held through roughly 400k tokens; beyond that it declined gradually while latency and cost grew roughly linearly.}
  On the long‑context sweep, exact‑match averaged about 87\% at 500k tokens and about 84\% at 900k; fuzzy averaged about 69\% at 900k. Adjusted accuracy that credits off‑by‑one remained high until the very top end. See Figure~\ref{fig:flash25-length-sweep}.

  \item \textbf{Exact‑match was a benchmark, not a recommendation.}
  When passages are truly verbatim, deterministic matching is cheaper and faster than using a model. We used exact‑match to isolate format and placement effects before moving to the fuzzy tasks, which reflect the production scenario. See \S\ref{sec:timestamps} and \S\ref{sec:fuzzy}.

\end{enumerate}

\subsection{Findings grouped by what was compared}
% ─────────────────────────────────────────────────────────────────────────────
\begin{table}[H]
\centering
\small
\setlength\tabcolsep{6pt}
\renewcommand{\arraystretch}{1.15}
\begin{tabularx}{\textwidth}{@{}l l l X@{}}
\toprule
\textbf{Prompt / Layout Factor} & \textbf{Scope} & \textbf{Sections} & \textbf{Observed effect (direction \& notes)}\\
\midrule
\makecell[l]{\textbf{Query placement} \\ Q$\uparrow$ vs.\ Q$\downarrow$} &
Full; Deep Dive; Pilot &
\makecell[l]{\S\ref{sec:gemini-flash-json-vs-text},\\ \S\ref{sec:timestamps}} &
Q$\uparrow$ $>$ Q$\downarrow$ by $\sim$3–20 pp depending on format/model; Q$\uparrow$ also lowers latency and median thinking tokens. Q$\downarrow$ interacts with model size (small models struggle).\\
\cmidrule(lr){1-4}
\makecell[l]{\textbf{Format} \\ TFT vs.\ JSON} &
Full; Deep Dive &
\makecell[l]{\S\ref{sec:gemini-flash-json-vs-text},\\ \S\ref{sec:timestamps}} &
TFT reduces input tokens $\sim$30–42\%; improves exact accuracy by $\sim$1–15 pp by converting off‑by‑one to exact. JSON is viable but more error‑prone, esp.\ under Q$\downarrow$.\\
\cmidrule(lr){1-4}
\makecell[l]{\textbf{Sentence–timestamp order (text)} \\ Text First / Middle / End} &
Deep Dive &
\S\ref{sec:text-field-order-variations} &
\emph{Text First} dominates; \emph{Text End} + Q$\downarrow$ is the hardest baseline (38.9\% exact; rises to $\sim$72–77\% with reasoning).\\
\cmidrule(lr){1-4}
\makecell[l]{\textbf{Transcript third} \\ (first, middle, last)} &
Full; Deep Dive &
\makecell[l]{\S\ref{sec:timestamps}} &
Last third is consistently hardest; penalty is amplified for Q$\downarrow$ and small models (Flash‑Lite, GPT‑5 nano).\\
\bottomrule
\end{tabularx}
\caption{Prompt \& layout factors. Abbrev.: Q$\uparrow$/Q$\downarrow$ as defined above; TFT = Text First Top.}
\label{tab:summary-prompt-layout}
\end{table}

% ─────────────────────────────────────────────────────────────────────────────
\begin{table}[H]
\centering
\small
\setlength\tabcolsep{6pt}
\renewcommand{\arraystretch}{1.15}
\begin{tabularx}{\textwidth}{@{}l l l X@{}}
\toprule
\textbf{Task / Difficulty Factor} & \textbf{Scope} & \textbf{Sections} & \textbf{Observed effect (direction \& notes)}\\
\midrule
\makecell[l]{\textbf{Span length} \\ $\ell\in\{1,2,3,5,10\}$} &
Full; Deep Dive &
\S\ref{sec:timestamps} &
Inverted‑U: best at 2–5; dips at 1 \& 10. Effect strongest in small models; large models near ceiling.\\
\cmidrule(lr){1-4}
\makecell[l]{\textbf{Fuzzy vs.\ Exact}} &
Full; Pilot &
\S\ref{sec:fuzzy} &
Unassisted fuzzy is harder by $\sim$4–10 pp on average; drop steepest for small models (nano/Lite).\\
\cmidrule(lr){1-4}
\makecell[l]{\textbf{Off‑by‑one dynamics}} &
Full; Deep Dive &
\makecell[l]{\S\ref{sec:off-by-one-error-description},\\ \S\ref{sec:timestamps}} &
Off‑by‑one rises under JSON (esp.\ Q$\downarrow$). ``Ladder'' effect under Q$\downarrow$: miss $\rightarrow$ off‑by‑one $\rightarrow$ exact as model size increases; TFT + Q$\uparrow$ collapses most near‑misses to exact.\\
\cmidrule(lr){1-4}
\makecell[l]{\textbf{Absent‑target rejection}} &
Full (light) &
\S\ref{sec:timestamps} &
False positives are rare; zero for GPT‑5 family in our test; slightly higher for very long spans on some Gemini models.\\
\bottomrule
\end{tabularx}
\caption{Task factors: lengths, fuzziness, error modes, and absent‑target behavior.}
\label{tab:summary-task}
\end{table}

% ─────────────────────────────────────────────────────────────────────────────
\begin{table}[H]
\centering
\small
\setlength\tabcolsep{6pt}
\renewcommand{\arraystretch}{1.15}
\begin{tabularx}{\textwidth}{@{}l l l X@{}}
\toprule
\textbf{Model / Provider Factor} & \textbf{Scope} & \textbf{Sections} & \textbf{Observed effect (direction \& notes)}\\
\midrule
\makecell[l]{\textbf{Model size / scale}} &
Full; Pilot &
\makecell[l]{\S\ref{sec:timestamps},\\ \S\ref{sec:fuzzy}} &
Scale improves exact \& fuzzy accuracy; robustness to length \& last‑third improves with size; cost rises with scale. Clear ``value'' vs.\ ``ceiling'' frontier.\\
\cmidrule(lr){1-4}
\makecell[l]{\textbf{Provider effects}} &
Full; Pilot &
\makecell[l]{\S\ref{sec:fuzzy},\\ \S\ref{tab:opt-vs-fuzzy}} &
Gemini $\rightarrow$ lower latency, smoother scaling; OpenAI $\rightarrow$ higher fuzzy accuracy on average (by $\sim$8.5 pp in our full run).\\
\bottomrule
\end{tabularx}
\caption{Model and provider patterns.}
\label{tab:summary-model-provider}
\end{table}

% ─────────────────────────────────────────────────────────────────────────────
\begin{table}[H]
\centering
\small
\setlength\tabcolsep{6pt}
\renewcommand{\arraystretch}{1.15}
\begin{tabularx}{\textwidth}{@{}l l l X@{}}
\toprule
\textbf{Efficiency Metric} & \textbf{Scope} & \textbf{Sections} & \textbf{Observed effect (direction \& notes)}\\
\midrule
\makecell[l]{\textbf{Reasoning tokens}} &
Full; Deep Dive &
\makecell[l]{\S\ref{sec:thinking-budget},\\ \S\ref{sec:timestamps}} &
Under Q$\uparrow$ (TFT or JSON), medians grow $\sim$+100 tokens per added sentence. Under Q$\downarrow$, medians are higher but flat vs.\ span length (fixed traversal cost). Fuzzy adds $\sim$25\% median tokens; Flash‑Lite has steepest increase.\\
\cmidrule(lr){1-4}
\makecell[l]{\textbf{Latency}} &
Full; Deep Dive &
\makecell[l]{\S\ref{sec:timestamps},\\ \S\ref{tab:opt-vs-fuzzy}} &
JSON Q$\uparrow$ is slowest on average (e.g., $\sim$21.2s); TFT is lowest ($\sim$13.7s). Fuzzy adds $\sim$3s vs.\ TFT exact. Gemini‑Flash is fastest; GPT‑5 slowest.\\
\cmidrule(lr){1-4}
\makecell[l]{\textbf{Cost \& caching}} &
Full &
\makecell[l]{\S\ref{sec:timestamps}} &
Provider prefix‑caching can make Q$\downarrow$ \emph{appear} cheaper when the fixed transcript sits at the prompt prefix; caching is provider‑dependent and not transferable across models. Cost per correct favors small models when accuracy is adequate.\\
\bottomrule
\end{tabularx}
\caption{Efficiency: tokens, latency, and cost.}
\label{tab:summary-efficiency}
\end{table}

% ─────────────────────────────────────────────────────────────────────────────
\begin{table}[H]
\centering
\small
\setlength\tabcolsep{6pt}
\renewcommand{\arraystretch}{1.15}
\begin{tabularx}{\textwidth}{@{}l l l X@{}}
\toprule
\textbf{Assistance / Variants} & \textbf{Scope} & \textbf{Sections} & \textbf{Observed effect (direction \& notes)}\\
\midrule
\makecell[l]{\textbf{Assisted Exact (snippet control)}} &
Full (light) &
\makecell[l]{\S\ref{sec:fuzzy} (Assisted Control)} &
With dynamic snippets, exact‑match goes near‑ceiling even for weak models; shows input length is a first‑order driver of errors.\\
\cmidrule(lr){1-4}
\makecell[l]{\textbf{Assisted Fuzzy (hybrid)}} &
Full; Pilot &
\makecell[l]{\S\ref{sec:fuzzy},\\ \S\ref{sec:assisted-fuzzy-model-response-times}, \S\ref{sec:cpc-assisted-fuzzy}} &
Hybrid (RapidFuzz narrowing + LLM) boosts accuracy by +4 to +50 pp (pilot: up to +70–90 pp), halves latency, and cuts cost per correct by $\gtrsim$10$\times$. Biggest gains for weaker models.\\
\cmidrule(lr){1-4}
\makecell[l]{\textbf{Thinking budget sweep}} &
Deep Dive &
\S\ref{sec:thinking-budget} &
Granting a large budget (model self‑uses $\sim$600–850 tokens) dramatically rescues poor layouts (e.g., +36 pp) at negligible cost; smaller gains on strong setups (ceiling).\\
\cmidrule(lr){1-4}
\makecell[l]{\textbf{Label‑free TFT variant}} &
Deep Dive &
\S\ref{sec:gemini-flash-use-larger-model} &
Removing \texttt{start\_ms}/\texttt{end\_ms} labels (keeping only numbers) preserves accuracy while trimming tokens further (up to $\sim$42\% vs.\ JSON).\\
\bottomrule
\end{tabularx}
\caption{Assisted methods and variants.}
\label{tab:summary-assistance}
\end{table}

% ─────────────────────────────────────────────────────────────────────────────
\begin{table}[H]
\centering
\small
\setlength\tabcolsep{6pt}
\renewcommand{\arraystretch}{1.15}
\begin{tabularx}{\textwidth}{@{}l l l X@{}}
\toprule
\textbf{Long‑Context Robustness} & \textbf{Scope} & \textbf{Sections} & \textbf{Observed effect (direction \& notes)}\\
\midrule
\makecell[l]{\textbf{100k–900k tokens}} &
Deep Dive &
\S\ref{sec:accuracy-improv-tests} &
Accuracy stable through $\sim$400k; gradual decline thereafter (exact $\sim$87\% at 500k, $\sim$84\% at 900k; fuzzy $\sim$69\% at 900k). Adjusted accuracy stays high until top end. Latency \& cost scale near‑linearly with length.\\
\bottomrule
\end{tabularx}
\caption{Robustness across very long transcripts.}
\label{tab:summary-long-context}
\end{table}

%╔══════════════════════════════════════════════════════════════════════════════╗
%║  ███╗   ██╗███████╗██╗    ██╗    ███████╗███████╗ ██████╗                    ║
%║  ████╗  ██║██╔════╝██║    ██║    ██╔════╝██╔════╝██╔════╝                    ║
%║  ██╔██╗ ██║█████╗  ██║ █╗ ██║    ███████╗█████╗  ██║                         ║
%║  ██║╚██╗██║██╔══╝  ██║███╗██║    ╚════██║██╔══╝  ██║                         ║
%║  ██║ ╚████║███████╗╚███╔███╔╝    ███████║███████╗╚██████╗                    ║
%║  ╚═╝  ╚═══╝╚══════╝ ╚══╝╚══╝     ╚══════╝╚══════╝ ╚═════╝                    ║
%╚══════════════════════════════════════════════════════════════════════════════╝

%-----------------------------------------------------------------------------
% Motivation — Why ``Quote–to–Timestamp'' Matters
%-----------------------------------------------------------------------------
\section{Motivation — Why ``Quote--to--Timestamp'' Matters}

\subsection{A concrete production scenario}
\begin{figure}[H]
    \centering
    \includegraphics[width=1\linewidth]{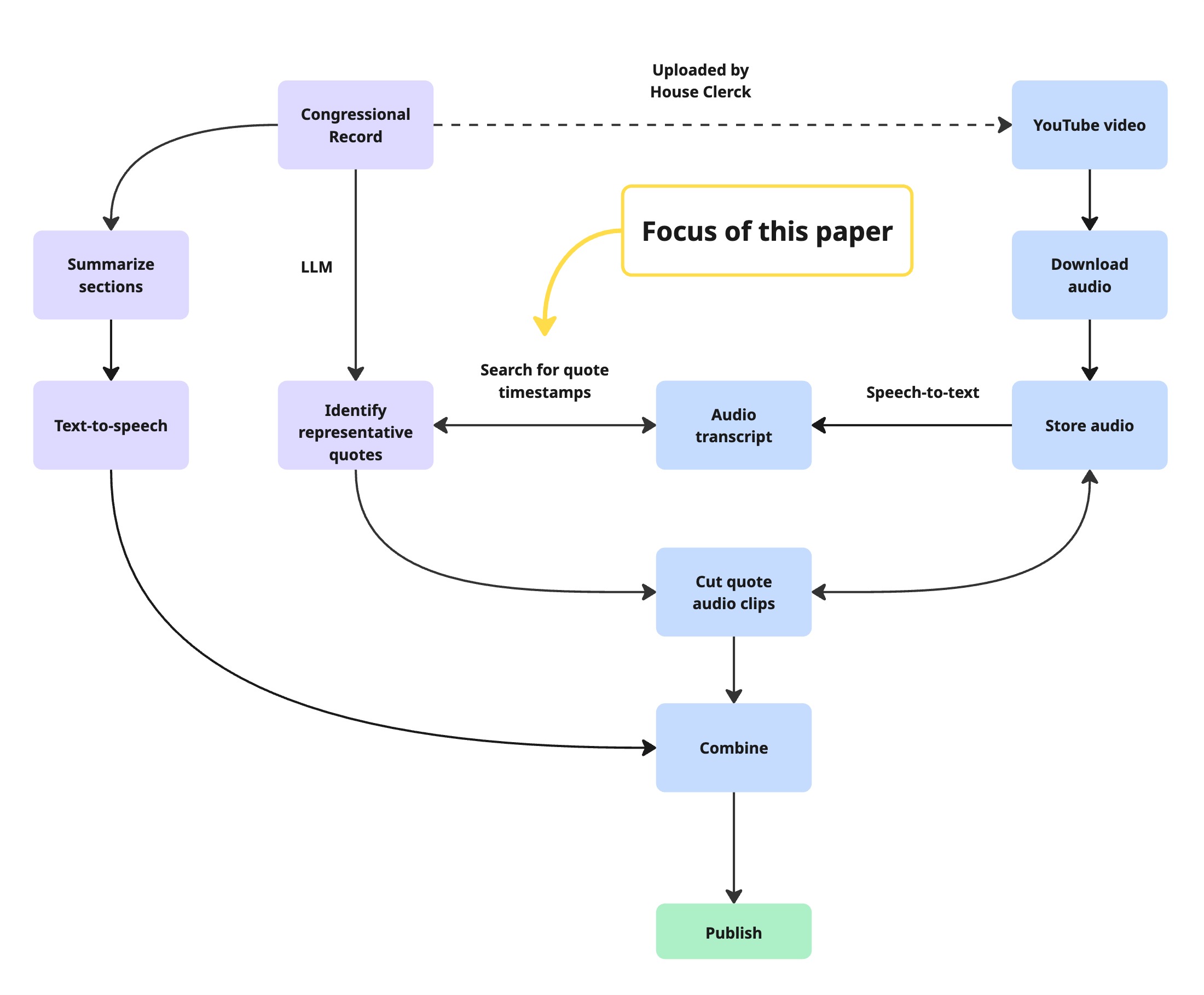}
    \caption{AI-powered podcast production workflow}
    \label{fig:podcast-workflow}
\end{figure}

Consider the practical task of constructing an automated, AI-produced long-form political podcast as described in Figure~\ref{fig:podcast-workflow}. This workflow allows for the insertion of short speech segments from the United States congressional floor, similar to human-edited news podcasts. The workflow begins with the \textit{Congressional Record of the United States} (CREC)—a lightly edited, speaker-attributed transcript of congressional floor proceedings that includes rich metadata such as member names, roll-call results, and session details.

As part of the workflow, an LLM curates representative quotes from the CREC. Separately, audio from the congressional floor is downloaded from YouTube videos of the floor proceedings and transcribed using speech-to-text (STT). Crucially, an automated process must search the transcribed proceedings and find the timestamps associated with the LLM's curated quotes (the focus of this paper). The quotes are then inserted as sound bites into the main podcast narration, which is generated using a text-to-speech (TTS) model based on an LLM-generated script. The production pipeline therefore has to align two slightly different strings for every quote:

\begin{enumerate}
    \item The quote from the official \textit{Congressional Record} returned by the LLM (perhaps not exactly verbatim).
    \item The quote as transcribed by speech-to-text (STT) technology from the raw floor audio.
\end{enumerate}

These two representations are not identical, yet it is impractical to rely solely on the transcribed floor proceedings. Using the CREC as the primary source for quote identification provides key advantages—official accuracy, completeness, and key metadata like speaker identification. By contrast, official audio and video archives—such as those broadcast by the Cable-Satellite Public Affairs Network (C-SPAN)—are not restricted to live floor debate. They may intersperse other governmental content, including coverage of committee hearings or replays of earlier proceedings for reference. As a result, the audio/video record cannot reliably be treated as a verbatim, continuous record of the floor.

Minor variations in actual speech delivery (such as pronunciation differences or microphone interference) can cause discrepancies between the audio transcript and the CREC text. Moreover, sentence boundaries differ between the edited CREC and unedited speech transcripts: the CREC employs punctuation marks such as colons and semicolons, which do not clearly correspond to spoken pauses or boundaries in STT-generated transcripts. Additionally, even when instructed to return quotes verbatim, the LLM may introduce slight textual variations, including altered punctuation, expansion or collapsing of abbreviations or contractions, or minor wording changes.

Because a single congressional sitting typically spans over three hours—more than 2,500 sentences, or roughly 120,000 text tokens—aligning these variants manually is infeasible. Extreme cases amplify this challenge: for instance, Senator Cory Booker’s 25-hour filibuster from 7 p.m. EDT on 31 March 2025 to 8:05 p.m. EDT on 1 April generated 13,000 sentences, and more than 640,000 tokens. \footnote{For this project Booker’s entire speech was transcribed using AssemblyAI, and token counts were calculated with the Google Gemini token counter. These counts include questions and remarks from other members of Congress to whom Booker yielded time.}

At such scales, manual matching would require days of human effort. Thus, an automated retrieval system capable of rapidly locating the best-matching STT sentence—despite punctuation differences, transcription errors, and paraphrasing—is essential. Fuzzy string matching and semantic search techniques consequently become mandatory components of a robust production pipeline.

Even projects that rely on a single source of truth transcript still face a timestamp-resolution problem. Downstream tools—podcast cutters, highlight generators, search widgets—need an accurate \verb|start_ms| / \verb|end_ms| pair for each quoted passage. Locating that window quickly and reliably is essential for fully automated, low-latency production.

Zooming beyond this specific workflow, accurate and efficient quote-to-timestamp matching is increasingly important in many other automated media scenarios. Even when there is only a single source of truth—such as scraping timestamped transcripts directly from video platforms—rapid, low-cost alignment remains crucial for scaling highlight extraction, interactive search, and content summarization. Moreover, robust timestamp matching supports fact-checking, citation validation, and real-time captioning applications, all of which rely heavily on precise temporal indexing to deliver credible, user-friendly experiences.

\subsection{Timestamp retrieval}
Many audio pipelines can benefit from an automated timestamp retrieval step that, given a quotation, locates the best-matching STT sentence and returns its associated \verb|start_ms|/\verb|end_ms| timestamps. Successful retrieval ensures that downstream tools—audio splicers, highlight generators, or interactive search widgets—operate confidently and accurately, pointing each textual quotation precisely to the corresponding slice of audio.

To achieve timestamp retrieval, pipelines depend upon high-quality, sentence-level transcripts generated by speech-to-text (STT) services such as \href{https://github.com/m-bain/whisperX}{WhisperX}, \href{https://www.assemblyai.com}{AssemblyAI}, \href{https://www.rev.ai}{Rev AI}, and \href{https://aws.amazon.com/transcribe/}{Amazon Transcribe}. These services typically output structured JSON data annotating each sentence with millisecond-level boundaries:

\begin{jsoncode}
[
    {
    "start_ms": 13361710,
    "end_ms": 13371520,
    "text": "Make no mistake, these tariffs are taxes."
    },
...
]
\end{jsoncode}

Increasing reliance on such transcripts underscores the criticality of accurate retrieval, as recent work has shown that automated pipelines like WhisperX require precise timestamped alignments to enable reliable indexing, summarization, and retrieval of audio segments \citep{Bain2023WhisperX, Hazen2006AutomaticAlignment}. While structured, these STT-produced transcripts introduce specific challenges, making naive exact-string matching approaches impractical.

\subsection{Why naive exact matching fails}
\label{sec:why-exact-match-fails}
Returning to the workflow from Figure ~\ref{fig:podcast-workflow}, simple exact-string matching methods—where a textual quotation from the CREC or LLM output is compared directly with STT output—quickly become unworkable, even when augmented by basic normalization rules such as substituting ``U.S.'' with ``United States''. Such business logic rapidly expands into unmaintainable complexity, as it cannot realistically anticipate all lexical variations or transcription idiosyncrasies. Four primary challenges illustrate why exact matching consistently fails in practice:

\paragraph{1. STT segmentation artifacts.}
Sentence segmentation by STT engines is frequently based on periods. Consequently, common abbreviations like ``Mr.'', ``Ms.'', or ``Inc.'' often mistakenly trigger artificial sentence boundaries. For example, a segment from the June 11, 2025, Congressional session transcript produced by AssemblyAI demonstrates this fragmentation clearly:

\begin{jsoncode}
[
    {
    "start_ms": 429150,
    "end_ms": 429550,
    "text": "Mr."
    },
    {
    "start_ms": 429550,
    "end_ms": 433350,
    "text": "Speaker, I rise today to honor an extraordinary woman, Ms."
    },
    {
    "start_ms": 433350,
    "end_ms": 442510,
    "text": "Marianne Murphy, maternity director ..."
    }
]
\end{jsoncode}

These segmentation artifacts significantly reduce the effectiveness of exact-match retrieval.

\paragraph{2. Lexical drift across sources.}
Minor textual differences introduced by both STT and LLM systems defeat exact-string matching. Common forms of lexical drift include:

\begin{itemize}[leftmargin=1.7em]
\item \textbf{Lexical variations}: contractions expanded or collapsed ("they're" $\leftrightarrow$ "they are"), abbreviations ("U.S." $\leftrightarrow$ "United States").
\item \textbf{Sentence-boundary shifts}: clauses merged or split due to punctuation-based segmentation, particularly involving abbreviations (e.g., ``Inc.'', ``Ms.'', ``Mr.'').
\item \textbf{Near-duplicate phrasing}: minor substitutions or paraphrasing (``costing jobs'' vs. ``destroying jobs'').
\item \textbf{SST noise}: insertion, deletion, or substitution errors—particularly homophones or dropped low-confidence words.
\item \textbf{Missing tokens}: STT output often omits low-confidence words, creating gaps compared to original text.
\item \textbf{Punctuation differences}: inconsistent punctuation choices between edited CREC text and raw STT output.
\end{itemize}

\paragraph{3. Technical‐term ambiguity.}
STT often struggles with legal or highly specialized language common in official records. Examples include numerical and alphanumeric ambiguities (``Clause 8 of Rule XX'' transcribed as ``Clause 8 of Rule 20'') or abbreviations expanded into unintended phonetic variants (``LGBTQ+'' transcribed as ``L, G, B, T, Q, plus''). While custom dictionaries can help, they are impractical for diverse or unpredictable settings like congressional debates, where specialized terms constantly change based on topic and speaker. Moreover, certain spoken acronyms (e.g., ``DOGE'' for ``Department Of Government Efficiency'') may be ambiguous, causing further divergence between official text and spoken form.

\paragraph{4. LLM paraphrase bias.}
Even when explicitly prompted for verbatim quotes from official records, even state-of-the-art LLMs may introduce subtle textual alterations. Small stylistic changes—such as contractions (``will not'' to ``won't''), slight rewordings, or altered punctuation—compound existing STT mismatches, further eroding exact-match accuracy.

\par\vspace{1em}

Taken together, these four challenges illustrate why straightforward exact-match retrieval methods consistently fail. Effective retrieval pipelines therefore must incorporate fuzzy lexical matching, semantic search techniques, and robust normalization strategies to accurately bridge the lexical gap between official transcripts, STT outputs, and LLM-selected quotes.

\subsection{Illustrative failure cases}
In this section, we highlight two examples that demonstrate how even seemingly straightforward retrieval tasks can lead to model errors, even for state-of-the-art models like GPT-5.

The first example comes from a separate, but related project where GPT-5 was asked to return the final three sentences of a speaker’s passage. This was intended to support boundary identification within a larger transcript, a task conceptually similar to returning quotes verbatim. While the model has a task completion rate above 90\%, there are still instances it does not return verbatim passages. As shown in Table~\ref{tab:failure-gpt5}, the model struggled with this passage: it produced sentences that matched portions of the reference text, but the returned material was not contiguous, complicating downstream use.

% =========================
% Case 1: GPT-5 (standardized)
% =========================
\begin{table}[H]
\centering
\caption{Non-contiguous return: GPT-5 was asked for the final three sentences. Highlighted spans match the source but are stitched from non-adjacent positions, breaking exact retrieval.}
\vspace{0.3em}
\begingroup\setlength{\emergencystretch}{2em}\sloppy
\begin{tabularx}{\textwidth}{@{}P P@{}}
\toprule
\textbf{Model return (GPT-5)} & \textbf{Reference passage} \\
\midrule
``\hl{Those families lost their husbands, fathers, and sons.} \hl{They simply want to have their sacrifice and law enforcement recognized by the people who were rescued.} \hl{That is not happening, and that is a shame.}'' &
``To this day, Officer Sicknick's family--there were three officers who eventually died of causes related to the attack on the Capitol on January 6. Those families would hope that the people who were saved that day might be grateful. \hl{Those families lost their husbands, fathers, and sons.} \hl{They simply want to have their sacrifice and law enforcement recognized by the people who were rescued.} I know there are a number of my colleagues on the other side who had exemplary careers in law enforcement. They understand better than any that when you have a son, daughter, husband, or wife who gives their life in service of protecting the people in this Chamber, you would hope for at least just a small gesture, a plaque to commemorate that heroism and service to their country. \hl{That is not happening, and that is a shame.}'' \\
\bottomrule
\end{tabularx}
\endgroup
\label{tab:failure-gpt5}
\end{table}

The second case, while hypothetical, highlights potential failure modes that emerge when LLM errors combine with discrepancies between a primary written source and a separately transcribed spoken record. In such situations, small deviations introduced by automatic transcription systems—for example, in punctuation, contractions, or disfluencies—interact with model errors. When the LLM is then asked to return exact sentences, this combination can easily produce mismatched or incomplete spans.

% =============================================
% Case 2: Lexically divergent but same meaning
% =============================================
\begin{table}[H]
\centering
\caption{Semantically identical but lexically divergent sentences. A quote–to–timestamp retriever must tolerate all highlighted differences.}
\vspace{0.3em}
\begingroup\setlength{\emergencystretch}{2em}\sloppy
\begin{tabularx}{\textwidth}{@{}P P@{}}
\toprule
\textbf{Target quote (LLM)} & \textbf{Closest transcript span (STT)} \\
\midrule
``Make no mistake. These tariffs are taxes; and \hl{they're} costing jobs here in the \hl{U.S.}.'' &
``\hl{Colleagues}, make no mistake, these tariffs are taxes. And \hl{they are} costing jobs here in the \hl{United States}.'' \\
\bottomrule
\end{tabularx}
\endgroup
\label{tab:failure-transcript}
\end{table}

These cases underscore that accurate, contiguous passage retrieval remains nontrivial, whether due to boundary-detection errors in powerful models like GPT-5 or the compounding effects of smaller models and noisy transcripts.

\subsection{Operational stakes}
These practical challenges motivate a specialized search task: matching semantically identical but syntactically distinct sentences across different textual sources. However, most real-world production workflows do not end at merely locating a textual quote—they require identifying the exact millisecond boundaries to cut, replay, or audit media clips. Thus, an effective solution must accomplish two things simultaneously: \textbf{(i)} locate the correct sentence(s) despite paraphrasing and transcription noise, and then \textbf{(ii)} precisely output the corresponding \verb|start_ms|–\verb|end_ms| timestamps.

Existing long-context evaluations of large language models — such as OpenAI’s million-token needle-in-a-haystack test or Anthropic’s Claude-3 recall demonstration — only require textual retrieval \citep{OpenAI2025LongContext, Anthropic2024NeedleRecall}. No publicly available benchmark currently tasks the model with returning numeric timestamps, yet timestamps are precisely what downstream applications (clip extraction tools, searchable archives, regulatory audits) fundamentally rely upon.

Accurate quote-to-timestamp retrieval unlocks important downstream capabilities:
\begin{itemize}[leftmargin=*]
\item \textbf{Frame-accurate editing.} Enables automated extraction of precise audio clips without manual scrubbing through multi-hour recordings.
\item \textbf{Source-linked journalism.} Allows fact-checkers to navigate directly from textual claims to their corresponding audio or video evidence.
\item \textbf{Searchable archives.} Makes lengthy media assets (hearings, lectures, podcasts) navigable via natural-language queries, jumping directly to the relevant timestamp.
\item \textbf{Regulatory compliance.} Ensures timestamp-exact evidence, protecting financial, legal, or compliance teams from the risks associated with misaligned audits.
\end{itemize}

In the remainder of this paper, we first review existing literature on ``needle-in-a-haystack'' retrieval tasks, focusing on their limitations with respect to timestamp extraction. We then introduce a simple benchmark task explicitly designed to evaluate timestamp retrieval, including the critical concept of an ``off-by-one'' timestamp error. Finally, we propose and test a hybrid search approach that combines fuzzy lexical matching—via RapidFuzz’s Levenshtein partial match—with an LLM-based refinement step, demonstrating significant performance improvements, especially for smaller language models.

\section{Literature Review: Needle-in-a-Haystack Retrieval in LLMs}

--------
Large context windows theoretically eliminate the need for external retrieval. Nevertheless, Lost in the Middle \citep{liu2024lost} showed that models struggle to exploit information buried mid-prompt, with accuracy dropping 30–50\% when relevant spans move away from the ends of the context. Model providers have since responded: GPT-4.1 advertises 100\% recall of needles anywhere inside a 1M-token prompt \citep{openai2025}, Claude 3 reports 99+\% on a strengthened Needle-in-a-Haystack evaluation \citep{anthropic2024claude3}, and Gemini-1.5 Pro reaches 99.7\% recall across up to 1M tokens \citep{google2024needle}.

None of these evaluations require mapping the retrieved text to precise temporal indices—a necessity for audiovisual navigation, meeting playback, or auditability. Chapter-Llama \cite{ventura2025chapter} trains an LLM to emit timestamps at paragraph granularity, supervised during finetuning. Textual domains lack analogous open tasks.
--------

Results do not necessarily match previous results in the literature. For example, ``Solar Ranking'' was the most accurately answered but was in the last third of the transcript---contrasting with prior ``lost-in-the-middle'' findings. As reported by \citet{hsieh2024ruler}, ``distraction needles'' reduce performance. The ``Funding Amount'' question had the most distraction needles (e.g., \$9.4 billion, which appeared eight times throughout the transcript), likely explaining its lower accuracy.

This doesn't necessarily match model providers. For instance, in their GPT-4.1 announcement \citet{openai2024gpt41} claims that ``GPT‑4.1, GPT‑4.1 mini, and GPT‑4.1 nano can process up to 1 million tokens of context...[including the] ability to retrieve a small hidden piece of information (a ``needle'') positioned at various points within the context window...In our internal needle in a haystack eval, GPT‑4.1, GPT‑4.1 mini, and GPT 4.1 nano are all able to retrieve the needle at all positions in the context up to 1M.'' However, in this simple control test the same models did not consistently retrieve the needle consistently across formats and questions. \citet{anthropic2024claude3}.

\citet{liu2024lost} found that language models exhibit a U-shaped performance curve when retrieving information from different positions in long contexts, specifically GPT-3.5-Turbo's performance can drop by more than 20 percentage points when relevant information is placed in the middle of the context.

\citet{he2024neverlost} constructed a question answering task where answers were contained within a set of multiple documents. In one version, documents were sorted by relevance, so that documents containing the answer were near the beginning of the context window. In the second version documents were sorted randomly so that the answer-containing documents were more likely to be at the middle or end of the context window. Performance dropped by up to 17.3\% for models like GPT-3.5-Turbo-16K and ChatGLM2-6B-32K when using random ordering compared to relevance-based ordering. This demonstrates that these models are implicitly best at accessing information near the beginning of their context window.

Recent work by \citet{wang2025reasoning} found that needle placement within the context (ranging from 2.5\% to 97.5\% depth) showed no significant impact on accuracy for multi-hop reasoning questions that require information from multiple supporting documents.

\citet{kuratov2024babilong} demonstrated that current LLMs effectively utilize only 10-20\% of their context window and that retrieval performance varies significantly based on fact location, with facts positioned in the middle of the context being most difficult to retrieve.

Graphs presented in \citet{zhao2024longagent} suggest that for single-document question answering some popular models accuracy degrades when retrieving needles in the middle of the context (Claude 2.1 200k, OpenAI GPT-4 128K) although performance can be improved via fine-tuning (LLaMA-2 7B, GPT-3.5 16k).

\subsection{Background and Challenges}
Needle-in-a-haystack tasks test an LLM’s ability to locate a small but critical piece of information (``the needle'') buried in a large context (``the haystack'').  
Early work revealed a strong \emph{position bias}. Liu et al.~(2023) showed that models perform best when the relevant information is at the beginning or end of the input, but accuracy drops sharply for information placed in the middle, a phenomenon dubbed ``lost-in-the-middle.'' Similar trends were reported by Kamradt~(2023), where ChatGPT-4’s recall declined once context length exceeded roughly 64 K tokens, especially for mid-document needles. Anthropic’s Claude 2.1 exhibited only 27 \% recall under similar conditions, again with markedly better performance when the needle appeared at the top or bottom of the prompt. Collectively, these studies underscored that early long-context LLMs often ``forgot'' details buried in the center of very long inputs.

\subsection{Advances in Long-Context Models}
Newer models have substantially mitigated position bias. OpenAI’s GPT-4.1 (2024) reports consistent 100 \% recall for needles located anywhere in a 1 M-token context, including its smaller ``Mini’’ and ``Nano’’ variants. Anthropic’s Claude 3 Opus (200 K+ window) reaches \(>\!99\%\) accuracy on random-sentence needle tests and can even flag sentences that appear out of place, indicating near-perfect recall plus a measure of verification. Both OpenAI and Anthropic credit architectural and training changes that explicitly optimized for needle-in-a-haystack behavior.

\subsection{Structured vs.\ Unstructured Input Approaches}
Needle retrieval is studied in both plain-text and structured formats. Key-value retrieval tasks (e.g., long JSON lists) constitute a structured variant where the needle is a specific key–value pair. Liu et al.~(2023) tested up to 300 pairs and again found position effects: LLaMA-2 7B excelled when the target pair was at the start or end, but struggled mid-context. GPT-3.5, however, showed strong retrieval across its entire 4 K-token window, suggesting robustness at shorter lengths.  
Benchmarks such as \textsc{DENIAHL} varied value types (numbers, letters, mixed) and broke simple patterns to test true recall versus pattern-completion. Results indicate that structured input does not guarantee better recall; performance depends on both data layout and the presence (or absence) of exploitable regularities.

\subsection{Question-Answering vs.\ Exact Recall Formulations}
Task framing also matters. In open-ended QA settings, a model may locate the needle but paraphrase it. For instance, ``Lost in the Middle’’ found GPT-3.5 could answer correctly yet paraphrased sentences, lowering lexical-match scores despite perfect semantic recall. In contrast, LLaMA-2 tended to quote verbatim. Prompting can shift behavior: adding ``return the most relevant sentence'' boosted Claude 2.1’s exact-sentence recall dramatically. Recent provider documentation (e.g., GPT-4.1) emphasizes format-following and structured extraction to ensure precise outputs when needed.

\subsection{Mitigation Strategies: Retrieval-Augmentation and Depth Restriction}
Before ultra-long context models, Retrieval-Augmented Generation (RAG) was the standard mitigation: a separate search component fetched a short snippet that was then fed to the LLM. Debate persists over RAG versus pure long-context methods. Anthropic argues that supplying the \emph{entire} document to a high-capacity model can outperform coarse retrieval for nuanced queries, while a 2025 Unstructured study found RAG more complete (and cheaper) than Gemini 2.0’s 1 M-token context on financial filings.  
Hybrid pipelines have emerged: a lightweight fuzzy search narrows candidate locations, and the LLM confirms or extracts the answer from a smaller window, effectively ``lifting'' the needle closer to the prompt’s surface.  

\subsection{Synthesis}
Needle-in-a-haystack retrieval has progressed from a well-documented failure mode to a near-solved capability in the latest generation of LLMs. Yet outcomes still hinge on choices such as structured vs.\ plain text, RAG vs.\ full-context, and exact-match vs.\ QA formulations. Our present study extends this line of work by systematically comparing JSON-structured prompts to raw text, varying snippet lengths, and evaluating fuzzy-matching assistance to identify best-practice combinations for robust, efficient needle retrieval.
See Table~\ref{tab:related}.

\begin{table}[h]
\centering
\begin{tabular}{@{}l l l l@{}}
\toprule
Problem & Representative Work & Granularity & Exact timestamps?\\
\midrule
Needle-in-a-Haystack & OpenAI, Anthropic, Google & Sentence & No\\
Key-value retrieval & Liu et al. (Lost-Middle) & Pair & No\\
Video temporal grounding & Chapter-Llama, TimeLoc & Seconds & Yes (multimodal)\\
Forced subtitle alignment & WhisperX, Kaldi & Word & Traditional STT\\
Fuzzy + LLM tooling & Bavaro 2023 blog & Sentence & Informal\\
\bottomrule
\end{tabular}
\caption{Comparison of related tasks.}
\label{tab:related}
\end{table}

%╔══════════════════════════════════════════════════════════════════════════════╗
%║  ███╗   ██╗███████╗██╗    ██╗    ███████╗███████╗ ██████╗                    ║
%║  ████╗  ██║██╔════╝██║    ██║    ██╔════╝██╔════╝██╔════╝                    ║
%║  ██╔██╗ ██║█████╗  ██║ █╗ ██║    ███████╗█████╗  ██║                         ║
%║  ██║╚██╗██║██╔══╝  ██║███╗██║    ╚════██║██╔══╝  ██║                         ║
%║  ██║ ╚████║███████╗╚███╔███╔╝    ███████║███████╗╚██████╗                    ║
%║  ╚═╝  ╚═══╝╚══════╝ ╚══╝╚══╝     ╚══════╝╚══════╝ ╚═════╝                    ║
%╚══════════════════════════════════════════════════════════════════════════════╝

%-----------------------------------------------------------------------------
% Methodology
%-----------------------------------------------------------------------------

\section{Methodology}
\label{sec:methods}

\subsection{Models Evaluated}
\label{models-evaluated}
Our experiments evaluate six core models from two major providers—Google and OpenAI—alongside nine additional models tested in a smaller pilot run. Google’s evaluation set includes the latest Gemini~2.5 models spanning Pro, Flash, and Flash-Lite. The OpenAI line-up reflects the company's GPT-5 series, which—according to a recent announcement \citep{openai2025gpt5}—will replace earlier GPT-4–class and o3/o4 models in the API. These older models were included in the early pilot test before OpenAI's deprecation announcement while the full experimental setup was still under development. The pilot test also includes Anthropic’s  Claude 4.0 and Claude 3.7 Sonnet variants as well as the lighter Claude 3.5 Haiku. After substantial testing Claude models were not included in the full run due to cost concerns. Models included in the full run had reasoning enabled (also called ``thinking'' by some providers), while for the pilot run reasoning was disabled. The reasoning setup for each model is discussed further in Seceion~\ref{sec:reasoning-setup}. The full model lineup is included in Table~\ref{tab:models-used}.

\begin{table}[H]
\centering
\small
\setlength\tabcolsep{8pt}
\begin{tabular}{@{}l l l l@{}}
\toprule
\textbf{Provider} & \textbf{Model} & \textbf{Coverage} & \textbf{Reasoning} \\
\midrule
OpenAI\footnotemark[1] & GPT-5        & Full  & Enabled \\
OpenAI                 & GPT-5 mini   & Full  & Enabled \\
OpenAI                 & GPT-5 nano   & Full  & Enabled \\
\midrule % <-- single line to separate full vs condensed
Google                 & Gemini-2.5 Pro        & Full   & Enabled \\
Google                 & Gemini-2.5 Flash      & Full   & Enabled \\
Google                 & Gemini-2.5 Flash-Lite & Full   & Enabled \\
\midrule\midrule % <-- double line for provider boundary
OpenAI                 & GPT-4.1      & Pilot & Disabled \\
OpenAI                 & GPT-4.1 mini & Pilot & Disabled \\
OpenAI                 & GPT-4.1 nano & Pilot & Disabled \\
OpenAI                 & o3           & Pilot & Disabled \\
OpenAI                 & o3-Mini      & Pilot & Enabled \\
OpenAI                 & o4-Mini      & Pilot & Disabled \\
\midrule % <-- single line to separate full vs condensed
Google                 & Gemini-2.5 Pro        & Pilot  & Enabled \\
Google                 & Gemini-2.5 Flash      & Pilot  & Enabled \\
Google                 & Gemini-2.5 Flash-Lite & Pilot  & Disabled \\
\midrule % <-- single line to separate full vs condensed
Anthropic              & Claude Sonnet 4.0 & Pilot & Disabled \\
Anthropic              & Claude 3.7 Sonnet & Pilot & Disabled \\
Anthropic              & Claude 3.5 Haiku  & Pilot & Disabled \\
\bottomrule
\end{tabular}
\caption{Models included in the benchmark. OpenAI announced plans to deprecate non–GPT-5 models in the API \citep{openai2025gpt5}. Models marked \emph{Pilot} were evaluated on a limited set of early tests and are summarized throughout the remainder of this paper where applicable. Reasoning status reflects whether the model’s default settings enabled internal reasoning during those runs.}
\label{tab:models-used}
\end{table}

\subsection{Evaluation Loop, Caching, and Task Inventory}
\label{sec:evaluation-loop-caching-task}
\subsubsection{Tasks Overview}
We evaluate all models under identical conditions and prompt formats, using a single 2{,}772-sentence transcript. In a limited number of tests, other transcripts were also used; these cases are explicitly marked. Requests are dispatched sequentially within each \emph{provider/model} block to maximize provider-side prefix caching and keep key--value state warm (see Section~\ref{sec:caching}).

We call the ``query'' the varying part of the prompt that instructs the model to find a particular span of consecutive sentences (e.g., a Congress member’s quote or short speech snippet). The span of consecutive sentences itself—appearing in the transcript—is the ``target''.

We evaluate two main formats: (i) a JSON format (the native output of AssemblyAI’s STT API) and (ii) an Text First Top format motivated by our format experiments (see Sections \ref{sec:gemini-flash-json-vs-text} and \ref{sec:text-field-order-variations}. In almost all experiments the query is placed before the transcript, with a single JSON baseline variant serving as the exception where it is placed after.

We organize tasks into three families: (1) information-retrieval controls that bridge prior needle-in-a-haystack work and timestamp retrieval; (2) exact-match grounding, where spans of $L \in \{1,2,3,5,10\}$ consecutive sentences are evaluated; and (3) fuzzy-match variants that pressure-test production scenarios, including an Assisted Fuzzy pipeline. For the exact-match family, we sample non-overlapping targets evenly across transcript thirds for each span of consecutive sentences. Unless otherwise noted, each configuration is run 3 times.

\subsubsection{Dataset}
We use a single floor-speech session of the U.S.\ House of
Representatives (4 h 04 m of audio) published by the
\emph{House Clerk} on YouTube\footnote{\url{https://www.youtube.com/watch?v=kuQUuFt8ugs}} and
therefore in the U.S.\ public domain.  Audio was transcribed by
AssemblyAI (paid tier), yielding a 2{,}772-sentence JSON transcript
(AssemblyAI’s sentence splits are occasionally coarse or misplaced).

\subsubsection{Reasoning setup}
\label{sec:reasoning-setup}
We define two experimental configurations: \textit{reasoning enabled} and \textit{reasoning disabled}. The Full coverage experiments enabled reasoning as preliminary testing showed that, especially with the GPT-5 class of models, reasoning was required for normal model operation. In the Pilot experiments reasoning was disabled.

\paragraph{Thinking.}  
The following settings were used to enable model reasoning:  
\begin{itemize}
    \item \textbf{OpenAI GPT-5 series:} Reasoning was set to \emph{medium effort}. We tested other effort levels and found medium to be the most appropriate balance.  
    \item \textbf{Google Gemini models:} Reasoning was set to \emph{dynamic thinking}, meaning the model automatically chose how many reasoning tokens to use.  
\end{itemize}

OpenAI exposes three main reasoning effort levels: \texttt{"low"}, \texttt{"medium"}, and \texttt{"high"}.  
For Gemini, setting \texttt{"thinking"}~=~\verb|-1| as the thinking budget enables \emph{dynamic thinking}, where the model automatically decides how many reasoning tokens to generate. In our experiments, this typically resulted in fewer than 1{,}000 tokens.  

For the early pilot tests, all models were left on default reasoning settings. For Gemini 2.5, the defaults are dynamic thinking for Pro and Flash and thinking off for Flash-Lite default~\citep{google2025models,google2025thinking}. For o-series models, only o3-mini exposes a reasoning effort parameter with options low, medium, and high; when not specified the default is medium~\citep{openai2025reasoningeffort}. The larger o3 and o4-mini do not expose a user-settable reasoning effort parameter. GPT-4.1 is not a reasoning model (it does not produce an internal chain-of-thought, though planning can be prompted)~\citep{openai2025gpt41guide}. For Anthropic, extended thinking is off by default and requires specifying a thinking budget (minimum 1,024 tokens); this was not enabled in early tests, so responses used the standard mode without extended~\citep{anthropic2025messages,anthropic2025extendedthinking}.

\begin{table}[H]
\centering
\footnotesize
\begin{tabularx}{\textwidth}{@{} l X X @{}}
\toprule
\textbf{Coverage} & \textbf{Models} & \textbf{Reasoning} \\
\midrule
Full (New runs) &
\textbf{OpenAI}: GPT-5; GPT-5-mini; GPT-5 nano \newline
\textbf{Google}: Gemini-2.5 Pro; Gemini-2.5 Flash; Gemini-2.5 Flash-Lite &
GPT-5 reasoning set to "medium"; Gemini set to dynamic thinking \\
\midrule
Pilot (Early runs) &
\textbf{OpenAI}: GPT-4.1; GPT-4.1 mini; GPT-4.1 nano; o3; o3-mini; o4-mini \newline
\textbf{Anthropic}: Claude-3.5 Haiku; Claude-3.7 Sonnet; Claude Sonnet 4.0 \newline
\textbf{Google}: Gemini-2.5 Pro; Gemini-2.5 Flash; Gemini-2.5 Flash-Lite &
All models were left at default settings. o3-mini defaults to \emph{medium}; o3 and o4-mini have no user-settable control and run with full internal reasoning; GPT-4.1 is not a reasoning model (no internal chain-of-thought / no thinking budget); Anthropic extended thinking off by default (not enabled here); Gemini defaults: 2.5 Pro and 2.5 Flash dynamic on by default, Flash-Lite off. \\
\bottomrule
\end{tabularx}
\caption{Models included in both the full experiment set and early pilot grouped, with their reasoning status used.}
\label{tab:reasoning-matrix}
\end{table}

\subsubsection{Evaluation Footprint}
\label{sec:evaluation-footprint}

We organize the benchmark into three broad tiers, progressing from targeted controls to the full timestamp–retrieval task, and finally to fuzzy matching. The Text First Top format is described in Section~\ref{sec:optimized-text-format-description-basic}.

\begin{enumerate}[label=(\roman*), itemsep=2pt, leftmargin=14pt]
    \item \textbf{Needle controls (Tasks 1a–1c)} — Three ``needle-in-a-haystack'' questions positioned in the first, middle, and last thirds of the transcript. Each question is sent to the model three times (\(N_{\text{try}}{=}3\)). Each is evaluated in three transcript formats:
    \begin{enumerate}[label=\alph*)]
        \item \textbf{Task 1a (JSON)} — Transcript provided in full JSON format with millisecond timestamps attached to each sentence.
        \item \textbf{Task 1b (Plain Text)} — Transcript provided in plain text with no timestamps. Serves as a baseline, reproducing standard needle-in-a-haystack setups without temporal grounding.
        \item \textbf{Task 1c (Text First Top)} — Transcript provided in an OPT format derived from JSON, placing sentence text before its timestamps to reduce input length while preserving timing boundaries.
    \end{enumerate}
    These controls anchor the benchmark by reproducing standard retrieval evaluations in a transcript setting before moving to timestamp-specific tasks.

    \item \textbf{Exact–match passage grounding (Tasks 2a–2c)} — Models are required to return the exact \texttt{start\_ms} and \texttt{end\_ms} boundaries for a target passage that appears verbatim in the transcript. Note that in this paper exact match retrieval acts as a testing ground and baseline for the more important fuzzy matching; if retrieval is known to be exact-match only then existing non-LLM matching algorthims are cheaper, faster, and more accurate. Five passage lengths are tested, \(L \in \{1,2,3,5,10\}\). For each length, \(S=12\) non-overlapping passages are sampled evenly across the first, middle, and last thirds of the transcript. Each passage is sent to the model three times (\(N_{\text{try}}{=}3\)). Details are further discussed in Table~\ref{tab:call-budget-exact} We evaluate three transcript–query configurations:
    \begin{enumerate}[label=\alph*)]
        \item \textbf{Task 2a (JSON, Query Top)} — Query placed after a JSON transcript, which includes full sentence–timestamp structure. This setup increases difficulty by requiring the model to process the entire transcript before receiving the task.
        \item \textbf{Task 2b (JSON, Query Bottom)} — Query placed before a JSON transcript, giving the model access to the task instruction upfront.
        \item \textbf{Task 2c (Text First Top, Query Bottom)} — Query placed before an OPT transcript, that places the sentence text before its millisecond timestamps, reducing input length while preserving boundaries.
    \end{enumerate}

    \par\vspace{1em}

    For example, when \(L=5\), we sample 12 passages of five consecutive sentences: four from the first third of the transcript, four from the middle third, and four from the final third. Each passage is evaluated three times, giving the model three independent opportunities to return the correct \texttt{start\_ms} and \texttt{end\_ms}. This procedure is applied for all tested passage lengths and across all three transcript–query formats (2a–2c). This same setup applies to all other passage lengths.

    \item \textbf{Fuzzy–match passage retrieval (Tasks 3a–3c)} — These tasks extend the exact–match setting to cases where the target passage has lexical perturbations or format drift, mimicking real-world conditions such as editorial edits or STT artifacts. All perturbations are synthetic, generated by Gemini~2.5~Pro using six few-shot examples of rephrasing (punctuation changes, contractions, synonym swaps, and light rewrites). Importantly, the same 60 base passages used in Tasks~2 are reused here as the starting point for perturbations, ensuring direct comparability. As before, five passage lengths \(L \in \{1,2,3,5,10\}\) and three transcript regions (first, middle, last third) are tested, with three independent tries per case (\(N_{\text{try}}{=}3\)). Details are further discussed in Table~\ref{tab:call-budget-fuzzy}. We evaluate three setups:
    
    \begin{enumerate}[label=\alph*)]
        \item \textbf{Task 3a (UnAssisted Fuzzy, Text First Top)} — Perturbed passages are searched against the full transcript. This setting tests whether models can directly tolerate lexical drift without additional assistance.
        
        \item \textbf{Task 3b (Baseline Snippet Control, Text First Top)} — 
        An exact–match control where passages are cropped to short context windows around their center. 
        Five snippet lengths are tested $L \in \{1,2,3,5,10\}$, with $S=3$ passages per length, positioned across the first, middle, and last thirds of the transcript. 
        This setup isolates the effect of dynamic snippets (as used in Task~3c) from the fuzziness itself. 
        Results from Tasks~2 (exact matches over the full transcript) provide a complementary comparison.
        
        \item \textbf{Task 3c (Assisted Fuzzy, Text First Top)} — Perturbed passages are first processed by our hybrid fuzzy–match algorithm (Section~\ref{sec:fuzzy-matching}), which uses RapidFuzz partial ratios to narrow the search to a dynamically extracted snippet. The snippet is then passed to the model for timestamp prediction. Compared to 3a, this tests the efficacy of assisted retrieval under more limited context windows, closer to production workflows.
    \end{enumerate}
\end{enumerate}

\subsubsection{Text First Top format}
\label{sec:optimized-text-format-description-basic}
Via testing outlined in Section~\ref{sec:accuracy-improv-tests} we found that translating JSON into a text format that places the sentence \textit{before} the timestamps reduces transcript token count by almost 30\% and improves performance. An example of this format is shown below. See Section~\ref{sec:text-field-order-variations} for testing details. This setup is referred to as the Text First Top format or TFT when abbreviated. The Text First Top format is used in Task 1c, Task 2c, and Tasks 2a-2c.

\vspace{0.5em}
\noindent\textbf{(a) Text before millisecond timestamps}\\
Sentence text comes first, then the start and end timestamps.
\begin{tcolorbox}[colback=gray!10, colframe=RoyalBlue, arc=2pt, boxrule=0.5pt, left=1mm, right=1mm, top=1mm, bottom=1mm, fontupper=\ttfamily]
\textbf{\textcolor{BrickRed}{The House will be in order.}}
 start\_ms: 34090, end\_ms: 35050;
\\
\ldots
\\
\textbf{\textcolor{BrickRed}{It's clear that we must slash this tranche of wasteful spending and continue down a path to fully restore fiscal sanity in our nation.}}
 start\_ms: 9323830, end\_ms: 9334470;
\\
\ldots
\\
\textbf{\textcolor{BrickRed}{The House stands adjourned until 10am tomorrow morning for morning hour debate.}}
 start\_ms: 14682790, end\_ms: 14685470;
\end{tcolorbox}

\subsubsection{Transcript–wide Distribution of passages}
\label{sec:transcript-dist-of-passages}
As described above, for the exact-match experiments we generate five sentence–length
conditions \(L=\{1,2,3,5,10\}\).  
For each length \(\ell\) we create \(S=12\) \emph{consecutive-sentence} sets that
evenly partition the transcript:

\begin{enumerate}[label=\alph*)]
  \item four set sampled from the \emph{first third}\,(\(0 \le i < \nicefrac{1}{3}N\)),  
  \item four from the \emph{middle third}\,(\(\nicefrac{1}{3}N \le i < \nicefrac{2}{3}N\)),  
  \item four from the \emph{last third}\,(\(\nicefrac{2}{3}N \le i < N\)),  
\end{enumerate}

where \(N\) is the total sentence count.  
Within each third we force at least \emph{one} set to lie within
\(\pm200\) sentence indices of its ``prototype'' anchor—index \(0\), the exact center, or \(N-\ell\) respectively—ensuring coverage of the extreme and middle-of-context positions. These generated sentences are reused in Tasks 2a and 2c as the base sentences which are then perturbed.

\subsubsection{Provider-specific structured output}
\begin{itemize}[itemsep=2pt,leftmargin=10pt]
  \item \textbf{OpenAI \& Google.}  Each call supplies a JSON Schema compiled
        from a \texttt{pydantic} model, enabling strict parsing
        (\href{https://platform.openai.com/docs/guides/prompt-caching}{OpenAI},
        \href{https://ai.google.dev/gemini-api/docs/caching?lang=python}{Gemini}).
  \item \textbf{Anthropic.}  Claude does not yet enforce schemas;
        we request \emph{``Respond in pure JSON''} and extract the first
        \texttt{\{} … \texttt{\}} span
        (\href{https://docs.anthropic.com/en/docs/build-with-claude/prompt-caching}{docs}).
        This method worked well and produced zero decode errors.
\end{itemize}

\subsubsection{Call Budget Per Model and Task Details}
\label{sec:call-budget-and-task-details}
The total number of evaluation calls per model is:

{\small
\[
\underbrace{9 \times 3}_{\substack{\text{Tasks 1a–1c} \\ \text{Needle Controls}}} +
\underbrace{5 \times 12 \times 3}_{\substack{\text{Task 2a} \\ \text{JSON, Top}}} +
\underbrace{5 \times 12 \times 3}_{\substack{\text{Task 2b} \\ \text{JSON, Bottom}}} +
\underbrace{5 \times 12 \times 3}_{\substack{\text{Task 2c} \\ \text{TFT, Top}}} +
\underbrace{5 \times 12 \times 3}_{\substack{\text{Task 3a} \\ \text{UnAssisted Fuzzy}}} +
\underbrace{5 \times 3 \times 3}_{\substack{\text{Task 3b} \\ \text{Snippet Control}}} +
\underbrace{5 \times 12 \times 3}_{\substack{\text{Task 3c} \\ \text{Assisted Fuzzy}}}
= \mathbf{1{,}035\ \text{API calls}}
\]
}

All calls are issued sequentially per model. The control and Task tables below specify the task details.

\par\vspace{1em}
\par\vspace{1em}

%---------------------- Controls Table ----------------------
\begin{table}[H]
\centering
\small
\setlength\tabcolsep{6pt}
\renewcommand{\arraystretch}{1.15} % uniform row height
\begin{tabular}{@{} L{4.4cm} L{8.0cm} c @{}}
\toprule
\multicolumn{3}{c}{\textbf{Needle Controls}} \\
\midrule
\textbf{Task family} & \textbf{Purpose and set–up} & \textbf{Calls / model} \\
\midrule\midrule

\makecell{\textbf{Task 1a: JSON}} &
Three ``needle-in-a-haystack’’ questions—(i) a budget figure, (ii) the day’s presiding–officer name \& date, and (iii) the change in U.S.\ solar-capacity rank—positioned respectively in the first, middle, and last third of the transcript. The prompt contains the full JSON + timestamp transcript. Each question is sent to the model three times for evaluation, $N_{\text{try}}{=}3$. This control helps bridge the gap between previous ``needle-in-the-haystack" metrics and the main timestamp task. &
$3_{\text{questions}}\times3_{\text{tries}} = 9$ \\
\cmidrule(lr){1-3}

\makecell{\textbf{Task 1b: Plain Text}} &
Same three questions as above, but the transcript is now in plain text format with no timestamps.
Each question is sent to the model three times for evaluation, $N_{\text{try}}{=}3$. This control helps bridge the gap between previous ``needle-in-the-haystack" metrics and the main timestamp task. &
$3_{\text{questions}}\times3_{\text{tries}} = 9$ \\
\cmidrule(lr){1-3}

\makecell{\textbf{Task 1c: Text First Top}} &
Same three questions as above, but the prompt now includes the Text First Top transcript format, a transformation of the JSON: Query Top, then ``[text], start\_ms: X, end\_ms: Y;''\}. Each question is sent to the model three times for evaluation, $N_{\text{try}}{=}3$. This control helps bridge the gap between previous ``needle-in-the-haystack" metrics and the main timestamp task. &
$3_{\text{questions}}\times3_{\text{tries}} = 9$ \\
\midrule\midrule

\multicolumn{2}{r}{\textbf{Total calls per model}} & \textbf{27} \\
\bottomrule
\end{tabular}
\caption{Per-model evaluation budget — controls.}
\label{tab:call-budget-controls}
\end{table}
%---------------------- End Controls Table ----------------------

%---------------------- Exact Match Baseline Table ----------------------
\begin{table}[H]
\centering
\small
\setlength\tabcolsep{6pt}
\renewcommand{\arraystretch}{1.15} % uniform row height
\begin{tabular}{@{} L{4.4cm} L{8.0cm} c @{}}
\toprule
\multicolumn{3}{c}{\textbf{Exact Match Passage Retrieval Baseline}} \\
\midrule
\textbf{Task family} & \textbf{Purpose and set–up} & \textbf{Calls / model} \\
\midrule\midrule

\makecell{\textbf{Task 2a: JSON}\\\textbf{(Query Top)}} &
Timestamp grounding for five passage lengths $L=\{1,2,3,5,10\}$ consecutive sentences.
For each length we sample $S=12$ non-overlapping passages, positioned evenly across the first, middle, and last thirds of the transcript.
The query (target sentence[s] and instruction) is placed \emph{after} a JSON transcript that includes full sentence–timestamp structure, increasing difficulty by requiring the model to process the transcript before receiving the task.
Each passage is sent to the model three times for evaluation, $N_{\text{try}}{=}3$. &
\(|L| \times S \times 3 = 5 \times 12 \times 3 = 180\) \\
\cmidrule(lr){1-3}

\makecell{\textbf{Task 2b: JSON}\\\textbf{(Query Bottom)}} &
Same setup as above (five passage lengths, $S=12$ passages per length across transcript thirds), but here the query is placed \emph{before} the JSON transcript, giving the model access to the task instruction upfront.
Each passage is sent to the model three times for evaluation, $N_{\text{try}}{=}3$. &
\(|L| \times S \times 3 = 5 \times 12 \times 3 = 180\) \\
\cmidrule(lr){1-3}

\makecell{\textbf{Task 2c: Text First Top}\\\textbf{(Query Bottom)}} &
Same setup as above (five passage lengths, $S=12$ passages per length across transcript thirds), but using the Text First Top transcript format: the sentence text appears before its millisecond timestamps (``[text], start\_ms: X, end\_ms: Y;''), reducing input length while preserving boundaries. The query is placed \emph{before} the transcript.
Each passage is sent to the model three times for evaluation, $N_{\text{try}}{=}3$. &
\(|L| \times S \times 3 = 5 \times 12 \times 3 = 180\) \\

\midrule\midrule
\multicolumn{2}{r}{\textbf{Total calls per model}} & \textbf{540} \\
\bottomrule
\end{tabular}
\caption{Per-model evaluation budget — exact-match baseline.}
\label{tab:call-budget-exact}
\end{table}
%---------------------- End Exact Match Baseline Table ----------------------

%---------------------- Fuzzy Passage Retrieval Table ----------------------
\begin{table}[H]
\centering
\small
\setlength\tabcolsep{6pt}
\renewcommand{\arraystretch}{1.15} % uniform row height
\begin{tabular}{@{} L{4.4cm} L{8.0cm} c @{}}
\toprule
\multicolumn{3}{c}{\textbf{Fuzzy Match Passage Retrieval}} \\
\midrule
\textbf{Task family} & \textbf{Purpose and set–up} & \textbf{Calls / model} \\
\midrule\midrule
\makecell{\textbf{Task 3a: UnAssisted Fuzzy}\\\textbf{(Text First Top)}} &
Perturbed passages are searched against the full transcript. Perturbations are synthetic, generated by Gemini~2.5~Pro with few-shot rephrasing (punctuation changes, contractions, synonym swaps, and light rewrites), simulating real-world conditions such as editorial edits or STT artifacts.  
Five passage lengths ($L=\{1,2,3,5,10\}$) and three transcript regions (first, middle, last) are tested, with $S=12$ passages per case and three tries per passage. &
\(|L| \times S \times 3 = 5 \times 12 \times 3 = 180\) \\
\cmidrule(lr){1-3}

\makecell{\textbf{Task 3b: Snippet Control}\\\textbf{(Text First Top)}} &
Exact–match control with snippets cropped around each passage’s center. 
Five snippet lengths ($L=\{1,2,3,5,10\}$) are tested, with $S=3$ non–overlapping passages per length, positioned across transcript thirds. 
This isolates the effect of dynamic snippets (as used in 3c) from fuzziness itself, while holding the input exact. 
Each passage is sent to the model three times ($N_{\text{try}}{=}3$). &
\(|L| \times S \times 3 = 5 \times 3 \times 3 = 45\) \\
\cmidrule(lr){1-3}

\makecell{\textbf{Task 3c: Assisted Fuzzy}\\\textbf{(Text First Top)}} &
Same perturbed passages as 3a, but a hybrid fuzzy–match algorithm (Section~\ref{sec:fuzzy-matching}) first uses RapidFuzz partial ratios to narrow context to a dynamically extracted snippet. The snippet is then passed to the model for timestamp prediction, testing assisted retrieval under tighter windows closer to production.  
Five passage lengths, $S=12$ per length, three transcript regions, and three tries per passage. &
\(|L| \times S \times 3 = 5 \times 12 \times 3 = 180\) \\

\midrule\midrule
\multicolumn{2}{r}{\textbf{Total calls per model}} & \textbf{405} \\
\bottomrule
\end{tabular}
\caption{Per-model evaluation budget — fuzzy passage retrieval.}
\label{tab:call-budget-fuzzy}
\end{table}
%---------------------- End Fuzzy Passage Retrieval Table ----------------------

\subsection{Evaluation Loop and Caching Strategy}
\label{sec:caching}

\label{sec:evaluation-loop}
We evaluate three providers—Google and OpenAI as part of the Full coverage experiments and Anthropic as part of the Pilot—all of which expose some form of prompt caching:

\begin{itemize}
  \item OpenAI documents prefix-caching
        (\href{https://platform.openai.com/docs/guides/prompt-caching}{docs}), but in practice hit-rates were low
        (\(\approx\!15\%\)) on our workload.
  \item Google Gemini caches requests whose \emph{prefix} is identical across calls
        (\href{https://ai.google.dev/gemini-api/docs/caching?lang=python}{docs}).  
  \item Anthropic Claude offers an explicit TTL–based cache
        (\href{https://docs.anthropic.com/en/docs/build-with-claude/prompt-caching}{docs}); we kept the default
        \texttt{5\,min}.  
\end{itemize}

%---------------------- New Algo Section ----------------------
\begin{algorithm}[H]
\DontPrintSemicolon
\SetKwInOut{KwData}{Input}
\KwData{providers $\mathcal{P}$; models per provider $\mathcal{M}_p$ \\
        lengths $L=\{1,2,3,5,10\}$; snippet lengths $L_{\text{snip}}=\{1,2,3,5,10\}$ \\
        formats $\mathit{fmt}\in\{\text{JSON},\text{TFT},\text{PLAIN}\}$; control questions $Q_c=3$ \\
        fuzzy passages $F$; sets per length $S=6$; tries $N_{\text{try}}=3$ \\
        $\text{TARGET\_POS}\in\{\text{Top},\text{Bottom}\}$; thinking flag $\Theta\in\{\text{Disabled},\text{Enabled}\}$}
\BlankLine

% ---- Thinking mode switch (outer guard) ----
\If{$\Theta=\text{Enabled}$}{
  \textsc{SetThinking}$(\text{on})$;
}\Else{
  \textsc{SetThinking}$(\text{off})$;
}
\BlankLine

% ---------------- Controls ----------------
\ForEach{$p \in \mathcal{P}$}{
  \ForEach{$\mathit{fmt} \in \{\text{JSON},\text{TFT},\text{PLAIN}\}$}{
    \textsc{SetPromptCache}$(\textit{system}_{\text{ctrl}},\,\textit{instr}_{\text{ctrl}}(\mathit{fmt},\textit{transcript}))$\;
    \ForEach{$m \in \mathcal{M}_p$}{
      \For{$q \gets 1$ \KwTo $Q_c$}{
        \For{$t \gets 1$ \KwTo $N_{\text{try}}$}{
          \textsc{RunControl}$(m[\Theta],\,q,\,\mathit{fmt})$;
        }
      }
    }
  }
}

% -------- Sentence-length JSON & OPT (cache based solely on TARGET_POS, then loop) --------
\ForEach{$p \in \mathcal{P}$}{
  \ForEach{$\mathit{fmt} \in \{\text{JSON},\text{TFT}\}$}{
    \If{$\text{TARGET\_POS}=\text{Bottom}$}{
      \textsc{SetPromptCache}$(\textit{system}_{\text{exact}},\,\textit{instr}_{\text{exact}}(\mathit{fmt},\textit{transcript}))$;
    }
    \ForEach{$m \in \mathcal{M}_p$}{
      \ForEach{$\ell \in L$}{
        \For{$s \gets 1$ \KwTo $S$}{
          \For{$t \gets 1$ \KwTo $N_{\text{try}}$}{
            \textsc{RunExact}$(m[\Theta],\,\ell,\,s,\,\mathit{fmt})$;
          }
        }
      }
    }
  }
}

% ---------------- Snippet controls ----------------
\ForEach{$p \in \mathcal{P}$}{
  \ForEach{$m \in \mathcal{M}_p$}{
    \ForEach{$\ell \in L_{\text{snip}}$}{
      \For{$t \gets 1$ \KwTo $N_{\text{try}}$}{
        \textsc{RunSnippetCtrl}$(m[\Theta],\,\ell)$;
      }
    }
  }
}

% ---------------- Fuzzy Full ----------------
\ForEach{$p \in \mathcal{P}$}{
  \ForEach{$m \in \mathcal{M}_p$}{
    \For{$f \gets 1$ \KwTo $F$}{
      \For{$t \gets 1$ \KwTo $N_{\text{try}}$}{
        \textsc{RunFuzzy}$(m[\Theta],\,f,\,\text{Full})$;
      }
    }
  }
}

% ---------------- Fuzzy Assist ----------------
\ForEach{$p \in \mathcal{P}$}{
  \ForEach{$m \in \mathcal{M}_p$}{
    \For{$f \gets 1$ \KwTo $F$}{
      \For{$t \gets 1$ \KwTo $N_{\text{try}}$}{
        \textsc{RunFuzzy}$(m[\Theta],\,f,\,\text{Assist})$;
      }
    }
  }
}

\caption{Phase-first execution with optional thinking. Controls: cache once per provider/format; run $Q_c$ questions with $N_{\text{try}}$ tries. Sentence-length: loop over $\mathit{fmt}\in\{\texttt{JSON},\texttt{TFT}\}$; if \texttt{TARGET\_POS=Bottom} then cache and execute. Snippet and fuzzy blocks iterate providers $\rightarrow$ models with $N_{\text{try}}$ tries, caching across tries. The term $m[\Theta]$ denotes model $m$ with reasoning $\Theta$ (on/off).}
\label{alg:phase-first-eval}
\end{algorithm}
%---------------------- End Algo Section ----------------------

\subsection{Hardware \& runtime.}
All experiments were executed from a single \textbf{2023 MacBook Air M2}
(8-core CPU, 8 GB RAM) over the public APIs; no GPU was involved.

%╔══════════════════════════════════════════════════════════════════════════════╗
%║  ███╗   ██╗███████╗██╗    ██╗    ███████╗███████╗ ██████╗                    ║
%║  ████╗  ██║██╔════╝██║    ██║    ██╔════╝██╔════╝██╔════╝                    ║
%║  ██╔██╗ ██║█████╗  ██║ █╗ ██║    ███████╗█████╗  ██║                         ║
%║  ██║╚██╗██║██╔══╝  ██║███╗██║    ╚════██║██╔══╝  ██║                         ║
%║  ██║ ╚████║███████╗╚███╔███╔╝    ███████║███████╗╚██████╗                    ║
%║  ╚═╝  ╚═══╝╚══════╝ ╚══╝╚══╝     ╚══════╝╚══════╝ ╚═════╝                    ║
%╚══════════════════════════════════════════════════════════════════════════════╝

\section{Off-by-one errors}
\label{sec:off-by-one-error-description}
To understand task results, we first introduce the concept of an \textit{off-by-one error}, a particular timestamp misalignment where the model's predicted start or end timestamp is shifted by exactly one sentence boundary compared to the actual target timestamps. Specifically, there are four categories of off-by-one errors:

\begin{itemize}
    \item \textbf{Start too early}: Model includes an extra preceding sentence.
    \item \textbf{Start too late}: Model omits the first sentence of the target passage.
    \item \textbf{End too early}: Model omits the final sentence of the target passage.
    \item \textbf{End too late}: Model includes an additional following sentence.
\end{itemize}

Although seemingly minor, off-by-one errors significantly affect quality in fully automated workflows. For example, in AI-produced podcasts, a model that returns a start timestamp one sentence too late may inadvertently omit crucial context at the beginning of a quotation, abruptly interrupting narrative flow. Conversely, a model predicting an end timestamp one sentence too late may inadvertently include unrelated or disruptive content, compromising the coherence of the audio snippet.

However, the severity of an off-by-one error varies with the intended use case. For workflows involving manual review—such as compliance audits or interactive search tools—an off-by-one error may be acceptable, especially if the system reliably locates timestamps that closely approximate the desired boundaries, significantly reducing manual labor.

The following JSON snippet illustrates a concrete example of an off-by-one error, where the model incorrectly returns a starting timestamp one sentence too early. Here, by selecting the start timestamp at 7{,}558{,}890 milliseconds instead of the correct 7{,}544{,}330 milliseconds, the model omitted a sentence that may give crucial context to the main intended point, potentially disrupting the narrative flow when presented in audio form.

Off-by-one errors are quantified with respect to our timestamp retrieval tasks througout the remainder of this paper.

\begin{jsoncode}
[
    {
      "start_ms": 7544330, // Actual target start
      "end_ms": 7558090,
      "text": "The substance of HR56 is irrelevant since there is no justification for Congress to legislate on local DC matters, but I will briefly discuss it."
    },
    {
      "start_ms": 7558890, // Model chooses this as start timestamp (off-by-one error)
      "end_ms": 7577080,
      "text": "Consistent with federal law, the position of the Major Cities Chiefs association and DC's values, DC limits cooperation with federal immigration agencies."
    },
    {
      "start_ms": 7578120,
      "end_ms": 7600700,
      "text": "DC concluded that cooperating with federal immigration agencies would make DC less safe for all residents by diverting police resources and discouraging immigrants from interacting with the police department and other government agencies."
    },
    {
      "start_ms": 7601500,
      "end_ms": 7606460,
      "text": "Many states, cities and counties have reached the same conclusion."
    },
    {
      "start_ms": 7606780,
      "end_ms": 7610860,
      "text": "I urge members to respect the will of D.C."
    },
    {
      "start_ms": 7610860,
      "end_ms": 7616940, // Actual target end
      "text": "residents by voting no no on this bill."
    },
    {
      "start_ms": 7617100,
      "end_ms": 7618140,
      "text": "And I yield back."
    }
]
\end{jsoncode}

\subsection{Literature Review}

Few existing studies explicitly address off-by-one errors in retrieval contexts similar to ours. However, some recent literature has observed related phenomena. 

For example, \citet{hsieh2024ruler} 
conducted a needle-in-a-haystack evaluation involving both a target key-value pair and distraction key-value pairs. When tasked to retrieve a target value based on a key, the Yi-34B-200K model by \href{https://www.01.ai/}{01.AI} frequently selected a nearby incorrect value instead. As the authors observed:

\begin{quote}
``In the extreme version, Yi often returns values from the vicinity of the target, suggesting a coarse match of the range but the lack of precision to locate the key when the target is in-distribution of the noises.''
\end{quote}

In another context, \citet{castillo2024beyond} test LLM memory in multi-turn conversations between a user and a model. One of these tests is ``Prospective Memory,'' which evaluates an agent's ability to execute instructions at a specified future point. In this test, the model is presented with a quote from one of nine historical figures (e.g., Benjamin Franklin, Eleanor Roosevelt, and Aristotle). The system randomly selects both a quote and a target response position (between the 2nd and 8th future responses). After presenting the quote, the model is instructed to append it to a specific future response, with the counting beginning from the current interaction. The authors observe that this task proved particularly challenging, noting that ``the most common failure was in the agent not being able to count its statement correctly, usually resulting in an off-by-one error.''

However, while these studies recognize off-by-one errors as important failure modes, neither quantifies their frequency nor systematically investigates their occurrence as a central research topic. 

In our own timestamp retrieval experiments, we observed a consistent tendency for models to produce predictions shifted by exactly one sentence boundary from the correct timestamps. To the best of our knowledge, this paper presents the first detailed quantitative investigation into the phenomenon of off-by-one errors. This specific focus is particularly relevant to timestamp-based retrieval tasks, which uniquely demand precise alignment between textual predictions and sentence-level timestamp boundaries. In contrast, most traditional needle-in-a-haystack evaluations primarily assess the model’s ability to recall or locate particular pieces of information without explicitly measuring such fine-grained positional misalignment.

%╔══════════════════════════════════════════════════════════════════════════════╗
%║  ███╗   ██╗███████╗██╗    ██╗    ███████╗███████╗ ██████╗                    ║
%║  ████╗  ██║██╔════╝██║    ██║    ██╔════╝██╔════╝██╔════╝                    ║
%║  ██╔██╗ ██║█████╗  ██║ █╗ ██║    ███████╗█████╗  ██║                         ║
%║  ██║╚██╗██║██╔══╝  ██║███╗██║    ╚════██║██╔══╝  ██║                         ║
%║  ██║ ╚████║███████╗╚███╔███╔╝    ███████║███████╗╚██████╗                    ║
%║  ╚═╝  ╚═══╝╚══════╝ ╚══╝╚══╝     ╚══════╝╚══════╝ ╚═════╝                    ║
%╚══════════════════════════════════════════════════════════════════════════════╝

%-----------------------------------------------------------------------------
% Accuracy Improvement Tests
%-----------------------------------------------------------------------------
\section{Google Gemini 2.5 Flash Deep Dive}
\label{sec:accuracy-improv-tests}
We begin with a single mid-sized model—Gemini 2.5 Flash—and conduct a comprehensive analysis of the factors that affect timestamp-retrieval accuracy. We chose Gemini 2.5 Flash for two reasons: (1) as a mid-sized model, it offers a strong performance–cost balance; and (2) it supports fine-tuning on transcript data, enabling additional performance tests. At project start, Anthropic did not permit fine-tuning, and only later enabled it in limited circumstances \citep{anthropic2024-haiku-ga}. We also evaluated OpenAI’s GPT-series models; although our training data passed the Moderations API when submitted directly, fine-tuning jobs were nevertheless flagged. Constraining ourselves to the three major providers while requiring fine-tuning support thus made Gemini 2.5 Flash the most practical choice.

This deep dive provides a useful set of baselines to help inform more limited testing done on a wider range of models later in the paper.

\subsection{Prompt Formats}
\label{sec:prompt-format-examples}
Our experiments test two main prompt formats—JSON and plain text—and further vary the position of text fields and target placement. For clarity, we organize this section into the following parts: (1) JSON vs. text format, (2) sentence field order within the text format, and (3) position of the target sentence within the prompt.

\subsubsection{JSON Format}
\label{sec:json-format-description}
The baseline format presents each sentence as a JSON object with \texttt{start\_ms}, \texttt{end\_ms}, and \texttt{text} fields. This is the native format of our STT provider, AssemblyAI and similar to outputs of other providers.

\begin{jsoncode}
{
  "start_ms": 9323830,
  "end_ms": 9334470,
  "text": "It's clear that we must slash this tranche of wasteful spending and continue down a path to fully restore fiscal sanity in our nation."
},
\end{jsoncode}

\subsubsection{Text Format: Field Order Variations}
\label{sec:text-field-order-variations}
We also ``umpack'' the JSON format into a text format with several variations described below. The text format offers a significant efficiency advantage: it is on average about 30\% shorter in token count compared to the equivalent JSON structure (exact reduction depends on provider-specific tokenization). This savings comes from eliminating required JSON syntax—such as braces (\texttt{{} \texttt{}}) and quotation marks around keys in key–value pairs. This directly reduces API cost and latency for large-context retrieval, especially at scale.

To evaluate the impact of field ordering within each sentence, we systematically test three patterns. Each pattern is illustrated below by a transcript snippet with the opening sentence, the central test sentence, and the closing sentence. Quoted text is in bold red; ellipses indicate omitted lines.

\vspace{0.5em}
\noindent\textbf{(a) Text before millisecond timestamps}\\
Sentence text comes first, then the start and end timestamps.
\begin{tcolorbox}[colback=gray!10, colframe=RoyalBlue, arc=2pt, boxrule=0.5pt, left=1mm, right=1mm, top=1mm, bottom=1mm, fontupper=\ttfamily]
\textbf{\textcolor{BrickRed}{The House will be in order.}}
 start\_ms: 34090, end\_ms: 35050;
\\
\ldots
\\
\textbf{\textcolor{BrickRed}{It's clear that we must slash this tranche of wasteful spending and continue down a path to fully restore fiscal sanity in our nation.}}
 start\_ms: 9323830, end\_ms: 9334470;
\\
\ldots
\\
\textbf{\textcolor{BrickRed}{The House stands adjourned until 10am tomorrow morning for morning hour debate.}}
 start\_ms: 14682790, end\_ms: 14685470;
\end{tcolorbox}

\vspace{0.5em}
\noindent\textbf{(b) Text between millisecond timestamps}\\
Start timestamp first, then sentence text, then end timestamp.
\begin{tcolorbox}[colback=gray!10, colframe=RoyalBlue, arc=2pt, boxrule=0.5pt, left=1mm, right=1mm, top=1mm, bottom=1mm, fontupper=\ttfamily]
start\_ms: 34090,
\textbf{\textcolor{BrickRed}{The House will be in order.}}
 end\_ms: 35050;
\\
\ldots
\\
start\_ms: 9323830,
\textbf{\textcolor{BrickRed}{It's clear that we must slash this tranche of wasteful spending and continue down a path to fully restore fiscal sanity in our nation.}}
 end\_ms: 9334470;
\\
\ldots
\\
start\_ms: 14682790,
\textbf{\textcolor{BrickRed}{The House stands adjourned until 10am tomorrow morning for morning hour debate.}}
 end\_ms: 14685470;
\end{tcolorbox}

\vspace{0.5em}
\noindent\textbf{(c) Text after millisecond timestamps}\\
Start and end timestamps come first, then the sentence text.
\begin{tcolorbox}[colback=gray!10, colframe=RoyalBlue, arc=2pt, boxrule=0.5pt, left=1mm, right=1mm, top=1mm, bottom=1mm, fontupper=\ttfamily]
start\_ms: 34090, end\_ms: 35050,
\textbf{\textcolor{BrickRed}{The House will be in order.}}
\\
\ldots
\\
start\_ms: 9323830, end\_ms: 9334470,
\textbf{\textcolor{BrickRed}{It's clear that we must slash this tranche of wasteful spending and continue down a path to fully restore fiscal sanity in our nation.}}
\\
\ldots
\\
start\_ms: 14682790, end\_ms: 14685470,
\textbf{\textcolor{BrickRed}{The House stands adjourned until 10am tomorrow morning for morning hour debate.}}
\end{tcolorbox}

\subsubsection{Query Placement}
\label{sec:query-placement}
We also vary whether the instructions including the target passage (the ``query'') appears at the top (before the transcript) or at the bottom (after the transcript) of the prompt.

\vspace{0.5em}
\noindent\textbf{Query at top}\\
The query sentence is shown first, followed by the transcript.

\begin{tcolorbox}[
  colback=gray!10, colframe=RoyalBlue, boxrule=0.5pt, arc=2pt,
  left=1mm, right=1mm, top=1mm, bottom=1mm,
  enhanced,
  listing engine=listings, listing only,
  listing options={
    style=mintlike,
    basicstyle=\ttfamily\small
  }
]
You are an expert at finding exact text matches in timestamped transcripts. \\

Your task is to find the start and end timestamps (in milliseconds) for a given sentence within transcribed audio data. \\

[Detailed instructions continue...] \\

TARGET SENTENCE: \textbf{\textcolor{BrickRed}{\{target\_passage\}}}  \\

TRANSCRIPT DATA: \textbf{\textcolor{BrickRed}{\{transcript\}}} 
\end{tcolorbox}

\vspace{0.5em}
\noindent\textbf{Query at bottom}\\
The query is shown after the transcript.

\begin{tcolorbox}[
  colback=gray!10, colframe=RoyalBlue, boxrule=0.5pt, arc=2pt,
  left=1mm, right=1mm, top=1mm, bottom=1mm,
  enhanced,
  listing engine=listings, listing only,
  listing options={
    style=mintlike,
    basicstyle=\ttfamily\small
  }
]
You are an expert at finding exact text matches in timestamped transcripts. \\

Your task is to find the start and end timestamps (in milliseconds) for a given sentence within transcribed audio data. \\

[Detailed instructions continue...] \\

TRANSCRIPT DATA: {transcript} \\

TARGET SENTENCE: \textbf{\textcolor{BrickRed}{\{target\_passage\}}}
\end{tcolorbox}

\subsection{Experimental Design: JSON and Text Format Variations}
\label{sec:exp-design-text-json}
Each trial uses a passage of consecutive sentences selected from the transcript. For this evaluation, we test five passage lengths $L = \{1, 2, 3, 5, 10\}$. For example, for $L=3$, transcript sentences 500--502 might be selected as a target passage; the model receives these as the query and must identify the corresponding timestamps within the full transcript.

For each length, 12 non-overlapping passages are sampled—4 from the first third, 4 from the middle third, and 4 from the last third—ensuring balanced coverage across the document (\S\ref{sec:transcript-dist-of-passages}).

For each trial, the model receives both a set of instructions including the target passage (this is called the ``query'') and the full transcript, provided either as structured JSON (\S\ref{sec:json-format-description}) or structured text (\S\ref{sec:text-field-order-variations}). The order of presentation is varied: in some prompts, the query appears before the transcript, and in others, after (\S\ref{sec:query-placement}). 

Note that in this experiment the query is an exact match of the target sentence. No fuzzy matching is used.

In the text condition, we further permute the order of the fields within each sentence, varying whether the sentence text appears before, between, or after the start and end timestamp fields (\S\ref{sec:text-field-order-variations}). 

Each configuration is evaluated three times, yielding a fully crossed grid of prompt format, target placement, sentence length, transcript region, and sentence field order, for a total of 1,440 trials. All results in this section use a thinking budget of zero. Table~\ref{tab:gemini-flash-experiment-grid} summarizes this setup.

This sentence-set configuration (five lengths of 1, 2, 3, 5, and 10 sentences, 12 sets per length, three trials each) will be used for the remainder of this section unless otherwise specified.

\begin{table}[H]
\centering
\footnotesize
\begin{tabular}{llcccS[table-format=3.0]}
\toprule
\textbf{Format} & \textbf{Query placement} & \textbf{Sentence field order} & \textbf{Lengths} & \textbf{Num. Passages} & {\textbf{Total trials}} \\
\midrule
JSON    & Before transcript & --                      & 1,2,3,5,10    & 12 per length & 180 \\
JSON    & After transcript  & --                      & 1,2,3,5,10    & 12 per length & 180 \\
Text    & Before transcript & Text before timestamps  & 1,2,3,5,10    & 12 per length & 180 \\
Text    & After transcript  & Text before timestamps  & 1,2,3,5,10    & 12 per length & 180 \\
Text    & Before transcript & Text between timestamps & 1,2,3,5,10    & 12 per length & 180 \\
Text    & After transcript  & Text between timestamps & 1,2,3,5,10    & 12 per length & 180 \\
Text    & Before transcript & Text after timestamps   & 1,2,3,5,10    & 12 per length & 180 \\
Text    & After transcript  & Text after timestamps   & 1,2,3,5,10    & 12 per length & 180 \\
\midrule
\multicolumn{5}{r}{\textbf{Total across all setups:}} & 1,440 \\
\bottomrule
\end{tabular}
\caption{Experimental grid for the ``no reasoning'' LLM setup. Each row represents 180 prompt trials (12 passages $\times$ 5 lengths $\times$ 3 tries), varying JSON and text layouts, quote placement, and field order within each sentence.} This is an exact matching task.
\label{tab:gemini-flash-experiment-grid}
\end{table}

\subsection{Results: JSON and Text Format Variations}
\label{sec:gemini-flash-json-vs-text}

We now present a head-to-head comparison of JSON and text transcript formats for the timestamp retrieval task. Table~\ref{tab:gemini-flash-json-vs-text-results} summarizes baseline results for all configurations, using Gemini 2.5 Flash and a thinking budget of zero tokens. For each, we report both exact-match accuracy and the adjusted accuracy that counts off-by-one errors as correct. See Section~\ref{sec:off-by-one-error-description} for detailed discussion of off-by-one errors.

\vspace{0.5em}
\noindent
For clarity, we introduce terminology that describes the key dimensions of each configuration: the transcript format (native JSON format from AssemblyAI or transformed text format), the position of the sentence text field relative to its timestamp fields (first, middle, or end), and the position of the query within the prompt (top/before or bottom/after the transcript). For example:
\begin{itemize}
  \item \textbf{Text First Top}: Transcript in text format, with the sentence text field appearing \emph{before} the timestamp fields (\texttt{start\_ms, end\_ms}) within each transcript entry, and the target passage placed \emph{before} the transcript in the prompt.
  \item \textbf{Text End Bottom}: Transcript in text format, with the sentence text field appearing \emph{after} the timestamp fields within each transcript entry, and the target passage placed \emph{after} the transcript in the prompt.
\end{itemize}

\noindent
This notation generalizes to all tested configurations (e.g., ``Text Middle Top,'' ``JSON Top,'' etc.), providing a concise way to reference each experimental condition throughout the analysis.

\subsubsection{Overview of Results}
\label{sec:gemini-2-5-results-overview}
For a fixed query position (e.g. after the transcript), the text format generally outperforms JSON. The largest gains are seen when (1) the query is placed at the top (before the transcript), and (2) the sentence text appears before the timestamp fields (for the text format). The best baseline accuracy, Text First Top, reaches 91.1\% (96.1\% adjusted), while the worst, Text End Bottom, achieves only 38.9\% (75.6\% adjusted). Even in the JSON format, placing the target block before the transcript delivers the best results, but performance still lags behind text-based configurations. Adjusted accuracy refers to counting off-by-one errors as correct responses.

\vspace{0.5em}
\noindent
The chart above summarizes the most substantial gains achieved through a series of prompt and transcript format modifications. In the following sections, we dive deeper into the error patterns and configuration-specific results that underlie these headline improvements.

\subsubsection{Detailed Results by Configuration}
\label{sec:detailed-config-results}

\begin{figure}[H]
    \centering
    \includegraphics[width=0.8\textwidth]{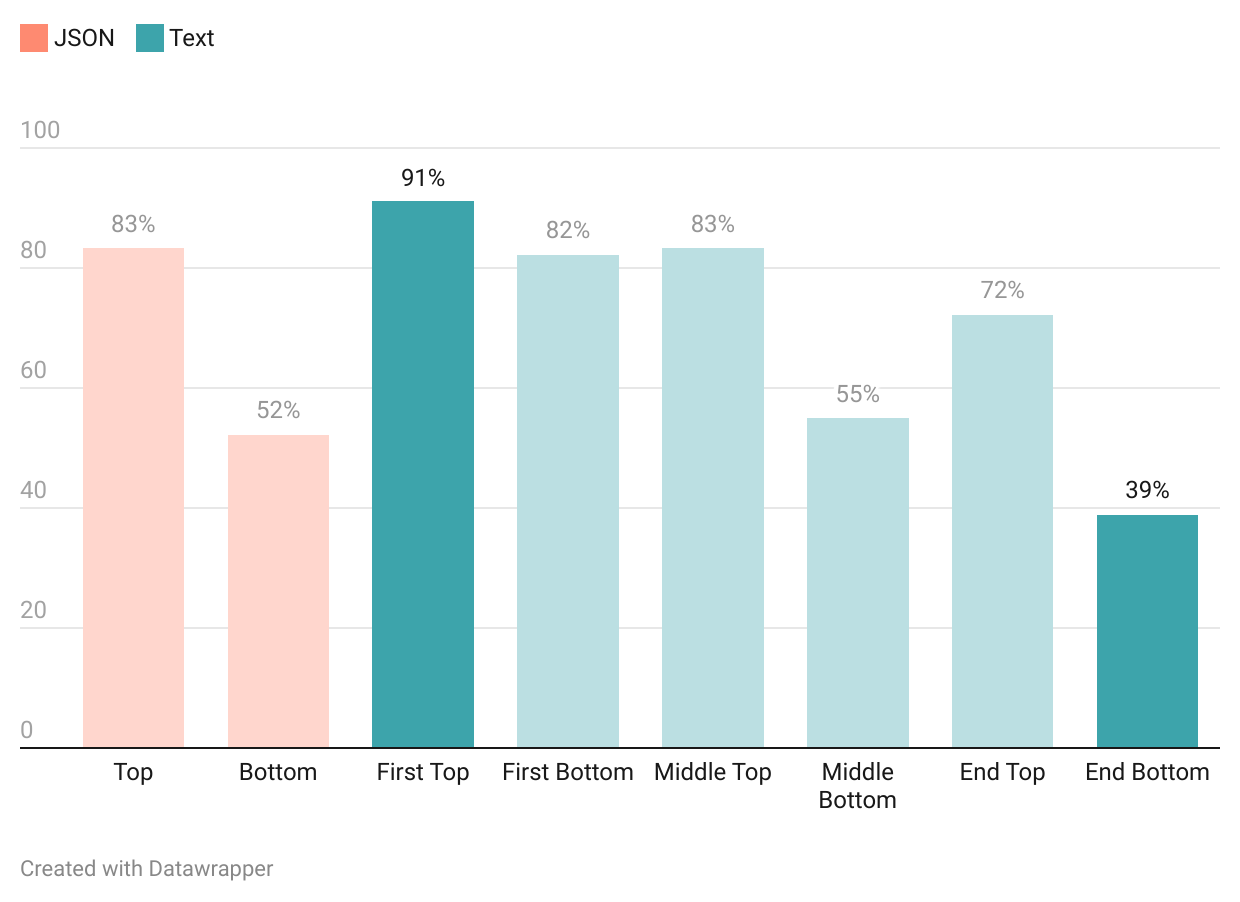}
    \caption{Performance comparison of Google’s Flash 2.5 model across 8 different prompt configurations, testing JSON vs text formats with varying query placements (Top/Bottom) and sentence locations (First/Middle/End). Results show text format significantly outperforms JSON, with Text First Top achieving the highest accuracy at 91\% and Text End Bottom performing worst at 39\%.}
    \label{fig:gemini-flash-format-config-results}
\end{figure}

To better understand the effects of each prompt and transcript format, we break down performance across all tested configurations. Figure~\ref{fig:gemini-flash-format-config-results} shows the exact-match accuracy for each combination of format, target placement, and sentence position. The accompanying table reports raw counts and percentages for both metrics, highlighting how each design choice impacts model accuracy.

\subsubsection{Error Types}
Most failures in both formats are off-by-one errors—predictions shifted by exactly one sentence. Counting off-by-one as correct nearly doubles effective accuracy in lower-performing configurations, particularly for Text End Bottom and JSON Bottom. Whether off-by-one \textit{should} be counted as correct depends on the application. In fully automated workflows, off-by-one errors may be treated as critical failures, whereas in human-in-the-loop applications getting ``close enough'' may be sufficient.

\vspace{1em}
\begin{table}[H]
\centering
\small
\begin{tabularx}{\textwidth}{l l l l l l l}
\toprule
\textbf{Configuration} & \textbf{Format} & \textbf{Target} & \textbf{Text Pos.} & \textbf{Exact} & \textbf{Adj.} & \textbf{Tests} \\
\midrule
Json Top              & JSON & Top    & --      & 150/180 (83.3\%) & 170/180 (94.4\%) & 180 \\
Json Bottom           & JSON & Bottom & --      & 94/180 (52.2\%)  & 145/180 (80.6\%) & 180 \\
\midrule
Text First Top        & Text & Top    & First   & 164/180 (91.1\%) & 173/180 (96.1\%) & 180 \\
Text First Bottom     & Text & Bottom & First   & 148/180 (82.2\%) & 169/180 (93.9\%) & 180 \\
Text Middle Top       & Text & Top    & Middle  & 150/180 (83.3\%) & 173/180 (96.1\%) & 180 \\
Text Middle Bottom    & Text & Bottom & Middle  & 99/180 (55.0\%)  & 137/180 (76.1\%) & 180 \\
Text End Top          & Text & Top    & End     & 130/180 (72.2\%) & 167/180 (92.8\%) & 180 \\
Text End Bottom       & Text & Bottom & End     & 70/180 (38.9\%)  & 136/180 (75.6\%) & 180 \\
\bottomrule
\end{tabularx}
\caption{JSON vs. Text format: Exact and adjusted (off-by-one counted correct) accuracy across prompt styles. Text Fist Top outperforms all other formats.}
\label{tab:gemini-flash-json-vs-text-results}
\end{table}

\subsection{Experimental Design: Thinking Budget Sweep}
We rerun only the best and worst configurations from the above grid with five different thinking budgets (zero, dynamic, 1{,}000 tokens, 5{,}000 tokens, 20{,}000 tokens). For each thinking budget, all five sentence lengths are evaluated for both prompt types.
As discussed in more detail in Section~\ref{sec:gemini-2-5-results-overview}, the best-performing configuration is the text format with sentence text \emph{before} the timestamp markers and the query \emph{before} the transcript. The worst-performing configuration is the text format with sentence text \emph{after} both timestamp markers and the query \emph{after} the transcript. 

\begin{table}[H]
\centering
\footnotesize
\begin{tabular}{lllcc}
\toprule
\textbf{Prompt format} & \textbf{Query placement} & \textbf{Sentence field order} & \textbf{Thinking budget} & \textbf{Total trials} \\
\midrule
Text & Before transcript & Text before timestamps (best) & Zero & 180 \\
Text & Before transcript & Text before timestamps (best) & Dynamic & 180 \\
Text & Before transcript & Text before timestamps (best) & 1{,}000 tokens & 180 \\
Text & Before transcript & Text before timestamps (best) & 5{,}000 tokens & 180 \\
Text & Before transcript & Text before timestamps (best) & 20{,}000 tokens & 180 \\
\midrule
Text & After transcript & Text after timestamps (worst) & Zero & 180 \\
Text & After transcript & Text after timestamps (worst) & Dynamic & 180 \\
Text & After transcript & Text after timestamps (worst) & 1{,}000 tokens & 180 \\
Text & After transcript & Text after timestamps (worst) & 5{,}000 tokens & 180 \\
Text & After transcript & Text after timestamps (worst) & 20{,}000 tokens & 180 \\
\midrule
\multicolumn{4}{r}{\textbf{Total for budget sweep:}} & 1,800 \\
\bottomrule
\end{tabular}
\caption{
Configuration grid for the thinking budget sweep. Each row represents 180 prompt trials ($5$ lengths $\times$ $12$ sets $\times$ $3$ tries), across five thinking budgets: no thinking, dynamic, 1{,}000, 5{,}000, and 20{,}000 tokens. This is an exact matching task.
}
\label{tab:gemini-flash-budget-experiment-grid}
\end{table}

\subsection{Results: Thinking Budget Sweep}
\label{sec:thinking-budget}

Beyond prompt format and layout, another crucial variable is the \emph{thinking budget}: the explicit allowance of extra tokens for the model to ``think'' or reason before responding. In this context, ``thinking'' refers to granting the model additional space to reflect or plan, rather than immediately generating a prediction. Recent advances have suggested that structured reasoning—whether by explicit chain-of-thought, system-instructed planning, or simple token allocation—can improve model accuracy, particularly on tasks that benefit from stepwise analysis.

To quantify these effects, we systematically varied the thinking budget across both the best-performing (Text First Top) and worst-performing (Text End Bottom) configurations. For each, we measured accuracy, average tokens consumed, and cost.

\medskip
\noindent
Note that the no-thinking baseline for each configuration was re-tested in this experiment. For Text First Top, accuracy was 91\% in the initial experiment and 89\% in the thinking-budget retest; for Text End Bottom, the scores were 39\% and 37\%, respectively. These reproduced figures match what we might expect given the findings in Section~\ref{sec:gemini-flash-confidence-intervals}.

\medskip
\noindent
The results (Figure~\ref{fig:gemini-flash-format-config-results} and Figure~\ref{fig:gemini-flash-successive-configs}) reveal several key findings:

\begin{itemize}
    \item \textbf{Thinking delivers dramatic gains for weak prompts at negligible cost.} Allocating a large thinking budget (up to 20,000 tokens per trial) enables even the worst prompt configuration (Text End Bottom) to leap from just 37\% accuracy to as high as 77\%—a near-doubling of performance—while the best configuration (Text First Top) rises from 89\% to 96\%. Despite these large allocations, Gemini Flash 2.5 typically uses only 600--850 tokens for reasoning, so the actual marginal cost per trial remains extremely low: for example, an 800-token thinking step costs just \$0.002 (two tenths of a cent) at \$2.50 per million output tokens. This means outsized accuracy gains can be achieved for poor prompt designs with virtually no practical cost penalty, at least for exact match timestamp retrieval tasks. Though note that in this paper exact match retrieval acts as a testing ground and baseline for the more important fuzzy matching; if retrieval is known to be exact-match only then existing non-LLM matching algorthims are cheaper, faster, and more accurate.
    \item \textbf{Thinking helps bad configurations more than good ones, but this is partly due to ceiling effects.} Poorly-structured prompts, like Text End Bottom, see large apparent accuracy gains (up to +36\%) when granted even modest thinking budgets, primarily because they have more room for improvement. In contrast, strong configurations (Text First Top) start near the ceiling, so further gains are necessarily smaller—not because they do not benefit from reasoning, but because their performance is already high and cannot exceed 100\%. Thus, the diminishing returns in high-performing settings reflect a ceiling effect, not an absence of benefit from increased thinking.
    \item \textbf{Token usage is self-regulating.} Regardless of the budget allocated (from 1,000 to 20,000 tokens), the model typically uses only what it ``needs''—around 600--850 tokens for reasoning—suggesting an intrinsic convergence to an optimal reasoning length.
    \item \textbf{Diminishing returns and efficiency.} Very large thinking budgets are rarely necessary: near-maximum accuracy is achieved with moderate allocations, and excess budget is simply unused. This has practical implications for cost and model-caching efficiency.
\end{itemize}

\begin{figure}[H]
    \centering
    \includegraphics[width=0.8\textwidth]{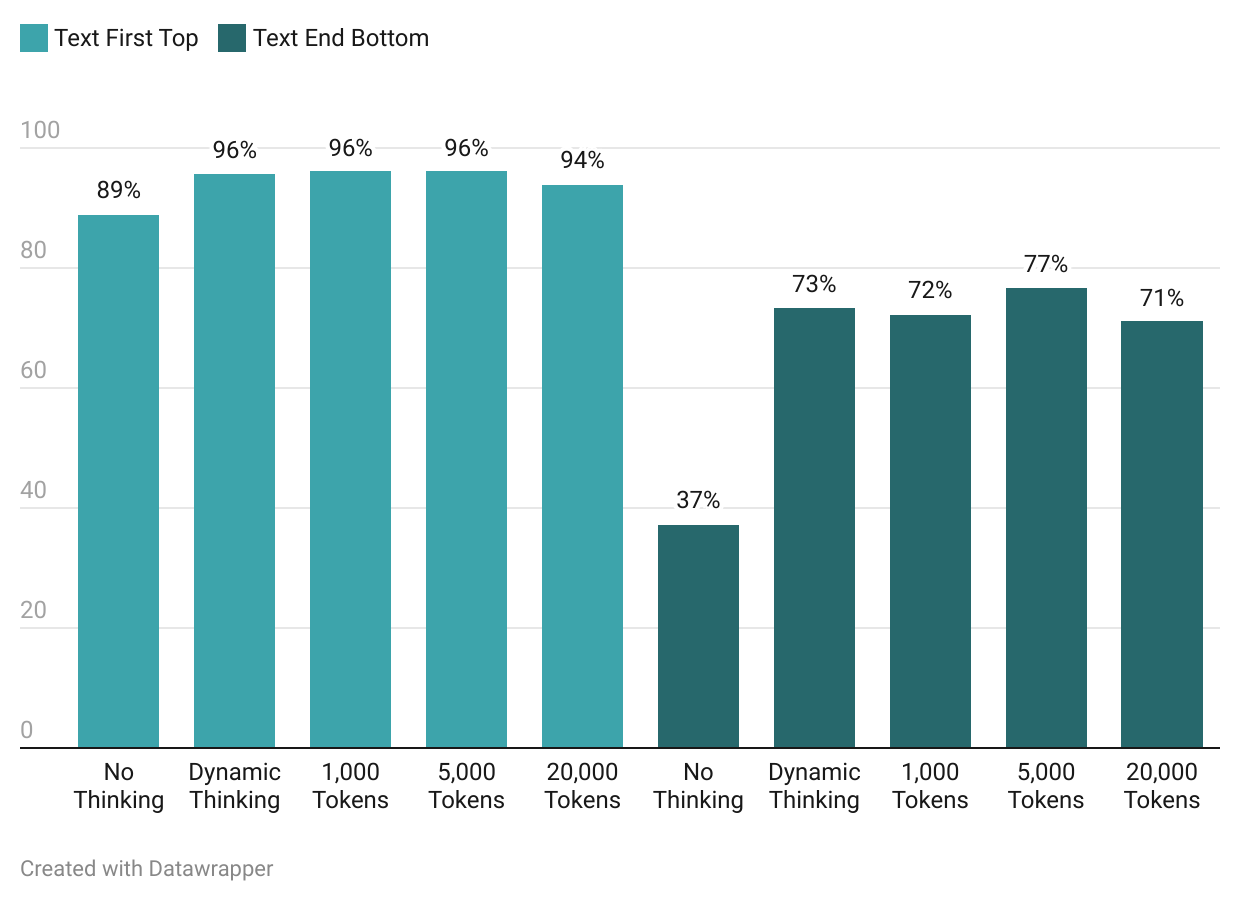}
    \caption{Performance comparison of Google's Flash 2.5 model when provided different thinking token budgets. No thinking performs worse regardless of format, but the effect is mitigated in the Text First Top format with the query at the top and the sentence text before timestamp markers. Once allowed to think, increasing the model's thinking budget has no noticeable impact on performance, likely because despite the larger budgets the model typically thought for only 1,000 tokens or less.}
    \label{fig:flash25-thinking-strategy}
\end{figure}

\begin{table}[H]
\centering
\small
\begin{tabular}{l l c c c}
\toprule
\textbf{Configuration} & \textbf{Thinking Budget} & \textbf{Accuracy} & \textbf{Avg. Tokens} & \textbf{Tests} \\
\midrule
\multicolumn{5}{l}{\textit{Best: Text First Top}} \\
\midrule
Text First Top   & No thinking      & 88.9\% & --   & 180 \\
Text First Top   & Dynamic          & 95.6\% & 712  & 180 \\
Text First Top   & 1,000 tokens     & 96.1\% & 721  & 180 \\
Text First Top   & 5,000 tokens     & 96.1\% & 653  & 180 \\
Text First Top   & 20,000 tokens    & 93.9\% & 849  & 180 \\
\midrule
\multicolumn{5}{l}{\textit{Worst: Text End Bottom}} \\
\midrule
Text End Bottom  & No thinking      & 37.2\% & --   & 180 \\
Text End Bottom  & Dynamic          & 73.3\% & 687  & 180 \\
Text End Bottom  & 1,000 tokens     & 72.2\% & 589  & 180 \\
Text End Bottom  & 5,000 tokens     & 76.7\% & 710  & 180 \\
Text End Bottom  & 20,000 tokens    & 71.1\% & 751  & 180 \\
\bottomrule
\end{tabular}
\caption{Best and worst baseline prompt configurations under different thinking budgets. Accuracy is exact-match. Dynamic means adaptive ``let-the-model-decide'' budget.}
\label{tab:budget-sweep}
\end{table}

\subsection{Increasing Exact Match Accuracy from 52\% to 98\%}
\begin{figure}[H]
    \centering
    \includegraphics[width=0.8\textwidth]{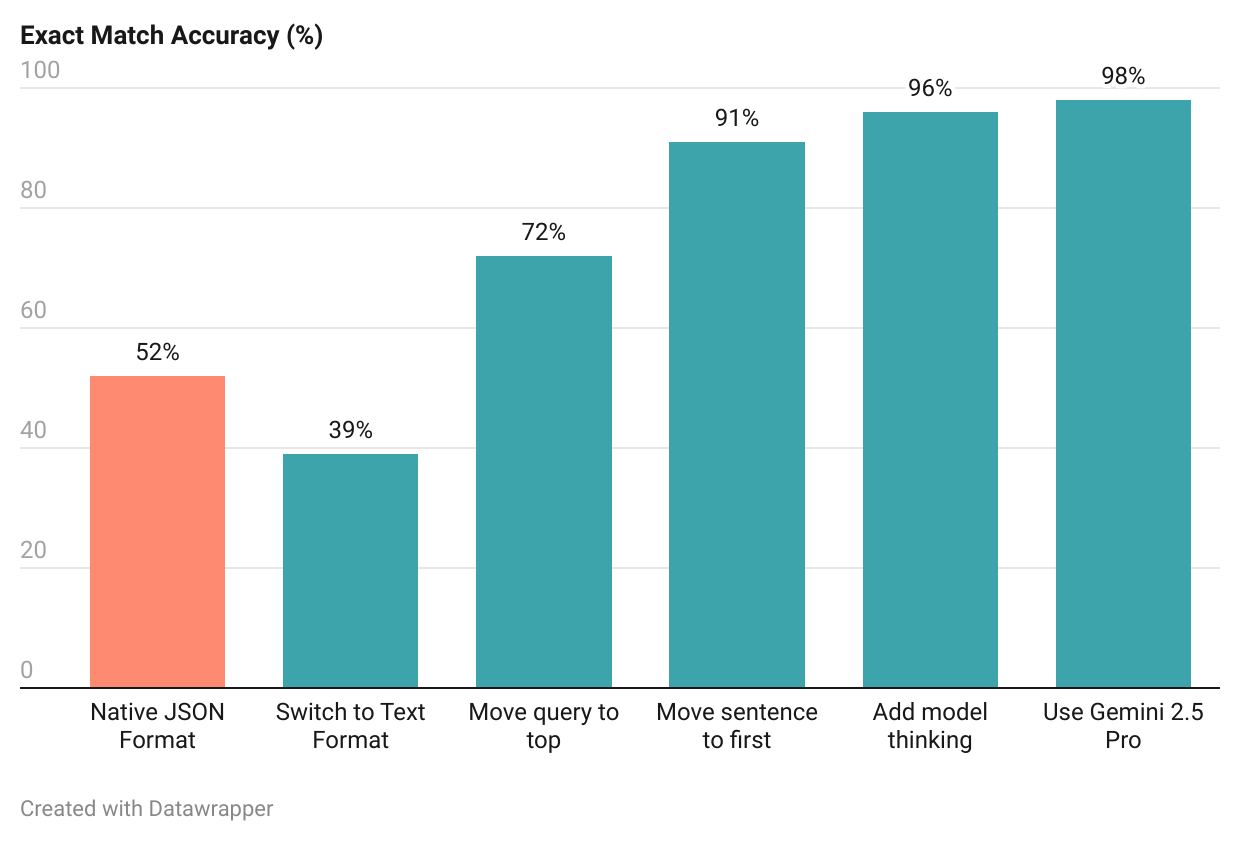}
    \caption{This chart shows how successive changes to prompt structure and transcript formatting dramatically increase exact-match accuracy for Google’s Gemini Flash 2.5 model on the timestamp retrieval task. Moving from the native JSON format of the Speech-to-Text provider to a text format reduces token count by 30\%, but also reduces accuracy by ~13 percentage points. Beginning with the weakest text format (``Text End Bottom''), each bar represents an improvement: (1) moving the query from the bottom to the top of the prompt, (2) placing the sentence text before the timestamp fields (instead of after), and (3) enabling model thinking. Each adjustment leads to a significant gain, with accuracy rising from 39\% to 96\%. Moving to a more powerful (but also slower and more costly model, Gemini 2.5 pro, further increases accuracy, but only by a few percentage points. These results highlight the outsized impact of thoughtful prompt and data formatting—well before model-level changes or tuning}
    \label{fig:gemini-flash-successive-configs}
\end{figure}

Figure~\ref{fig:gemini-flash-successive-configs} compiles the key results from Sections \ref{sec:gemini-flash-json-vs-text} and \ref{sec:thinking-budget}, showing how successive changes to the transcript format, query placement, and sentence placement (for the text format) gradually improves performance. This includes the impact of moving from Gemini 2.5 Flash to Gemini 2.5 Pro, which is tested in Section~\ref{sec:gemini-flash-use-larger-model}. Note this chart includes point estimates only and does not included expected variations due to the stochastic nature of model responses.

\subsection{Off-by-one Errors and Conversion to Exact Matches}

While adjusted accuracy (\S\ref{sec:gemini-flash-json-vs-text}) counts off-by-one predictions as correct, it is useful to examine the \emph{magnitude} of errors directly. Table~\ref{tab:json-text-error-breakdown} breaks results into three categories: 
\begin{enumerate}
    \item \textbf{Exact matches}. Start and end boundaries match the reference passage exactly.
    \item \textbf{Off-by-one}. Predicted boundaries are shifted by exactly one sentence earlier or later.
    \item \textbf{Major shifts}. Predicted boundaries are shifted by two or more sentences from the correct target.
\end{enumerate}
These categories sum to the total number of tests for each configuration.

\begin{table}[H]
\centering
\small
\begin{tabularx}{\textwidth}{l c c c c}
\toprule
\textbf{Configuration} &
\textbf{Exact} &
\textbf{Off-by-1} &
\textbf{Major shifts} &
\textbf{Tests} \\
\midrule
Json Top           & 150 (83.3\%) & 20 (11.1\%) & 10 (5.6\%)  & 180 \\
Json Bottom        & 94 (52.2\%)  & 51 (28.3\%) & 35 (19.4\%) & 180 \\
\midrule
Text First Top     & 164 (91.1\%) & 9 (5.0\%)   & 7 (3.9\%)   & 180 \\
Text First Bottom  & 148 (82.2\%) & 21 (11.7\%) & 11 (6.1\%)  & 180 \\
Text Middle Top    & 150 (83.3\%) & 23 (12.8\%) & 7 (3.9\%)   & 180 \\
Text Middle Bottom & 99 (55.0\%)  & 38 (21.1\%) & 43 (23.9\%) & 180 \\
Text End Top       & 130 (72.2\%) & 37 (20.6\%) & 13 (7.2\%)  & 180 \\
Text End Bottom    & 70 (38.9\%)  & 66 (36.7\%) & 44 (24.4\%) & 180 \\
\bottomrule
\end{tabularx}
\caption{Error breakdown for JSON vs.\ text configurations (Gemini 2.5 Flash, zero thinking). Columns report counts with percentages; ``Major shifts'' denotes predictions shifted by two or more sentences from the correct target. Exact + Off-by-1 + Major shifts $=180$ for every row.}
\label{tab:json-text-error-breakdown}
\end{table}

A clear pattern emerges: the majority of differences between configurations come from the reduction of off-by-one errors, though there is also a reduction in major shifts. In the worst-performing setup (Text End Bottom), off-by-one errors make up over a third of all trials (36.7\%), and major shifts account for nearly a quarter (24.4\%). By contrast, the best-performing setup (Text First Top) reduces off-by-one errors to just 5\% and major shifts to under 4\%. 

Quantitatively, 63\% of the improvement from Text End Bottom to Text First Top comes from converting off-by-one predictions into exact matches, with 90\% of all off-by-one errors present in Text End Bottom resolved as exact matches under the Text First Top setup. The remaining 36\% of the improvement results from the correction of major shifts. This trend mirrors the broader finding from Section~\ref{sec:detailed-config-results}: as prompt design approaches the optimal configuration, both the frequency and the magnitude of errors drop sharply.

\subsection{A Larger Model - Testing Gemini 2.5 Pro with the Optimal Configuration}
\label{sec:gemini-flash-use-larger-model}
Gemini 2.5 Pro was also tested using the 180-tests configuration described in Section~\ref{sec:exp-design-text-json} using the optimal configuration of Text First Top. Exact timestamp match accuracy increased from a maximum of 96.1\% for Gemini 2.5 Flash with thinking to 98.3\%. Notably, sentences of length one proved the most difficult, with the model achieving an accuracy of only 91.7\% on lengths of 1 sentence. For all other sentence lengths Gemini 2.5 Pro scored 100\% accuracy.

\subsection{Further Text Compression - Removing Timestamp Labels}
Transcript inputs can be compressed further by removing the \texttt{start\_ms} and \texttt{end\_ms} field labels from each sentence, leaving only the millisecond values (see below). In the input prompt, the model is then explicitly told the order of these values and that they represent the start and end times in milliseconds.

Using the 180-test setup described in Section~\ref{sec:exp-design-text-json} with the optimal \textbf{Text First Top} configuration, Gemini 2.5 Flash achieved the same accuracy as in earlier experiments with labeled timestamps: \textbf{88.3\%} without thinking and \textbf{95.0\%} with thinking.

\vspace{0.5em}
\begin{tcolorbox}[colback=gray!10, colframe=RoyalBlue, arc=2pt, boxrule=0.5pt, left=1mm, right=1mm, top=1mm, bottom=1mm, enhanced, sharp corners, fontupper=\ttfamily]
\textbf{\textcolor{BrickRed}{The House will be in order.}}, 34090, 35050;
\\
\ldots
\\
\textbf{\textcolor{BrickRed}{It's clear that we must slash this tranche of wasteful spending and continue down a path to fully restore fiscal sanity in our nation.}}, 9323830, 9334470;
\\
\ldots
\\
\textbf{\textcolor{BrickRed}{The House stands adjourned until 10am tomorrow morning for morning hour debate.}}, 14682790, 14685470;
\end{tcolorbox}

This ``no labels'' variant reduces token count by 42\% compared to the native AssemblyAI JSON output---likely the most compact format that still preserves the full transcript in a human-readable form. However, as discussed in Section~\ref{sec:fuzzy-matching}, the most effective strategy---especially when exact matches are uncertain---is to run a fuzzy search first, then pass high-probability candidate snippets to the model. A fully compressed transcript remains useful as a fallback when fuzzy-match retrieval fails.

\subsection{Stability and Confidence Intervals}
\label{sec:gemini-flash-confidence-intervals}
A natural question is how much variation there is in evaluation results and how confident we can be in those numbers. To determine this we followed Anthropic’s statistical guidance~\citep{anthropic2024stat}, we compute \emph{sentence-set-pooled} 95\% confidence intervals (CIs) to obtain an estimate of how the evaluation results would be expected to perform on new transcripts from the same Congressional Record.

It was cost-prohibitive to rerun every model-configuration pair, so we continued to use Gemini 2.5 Flash and focused on the Text First Top format. For this evaluation we constructed a new dataset of 10 Congressional Record transcripts, five for the U.S. House of Representatives and five for the U.S. Senate. The ensure a mix of different transcripts we pick various dates, starting in 1989 and ending in 2025. Likewise the length was variable from just 44,000 tokens up to 860,000 tokens.

\begin{table}[H]
\centering
\begin{tabular}{lrrrr}
\toprule
\textbf{Chamber} & \textbf{Date} & \textbf{Year} & \textbf{Transcript} & \textbf{Token length} \\
\midrule
House & September 28 & 2006 &  & 515{,}920 \\
House & April 16 & 2021 &  & 162{,}404 \\
House & June 12 & 2025 &  & 44{,}094 \\
House & July 3 & 2025 &  & 254{,}943 \\
House & July 17 & 2025 &  & 212{,}347 \\
Senate & February 7 & 1989 &  & 75{,}744 \\
Senate & October 15 & 1990 &  & 82{,}602 \\
Senate & July 18 & 1994 &  & 78{,}274 \\
Senate & January 1 & 2013 &  & 122{,}956 \\
Senate & March 31 & 2025 &  & 859{,}082 \\
\midrule
Average & — & — & — & 240{,}836 \\
\bottomrule
\end{tabular}
\caption{Token lengths for selected House and Senate transcripts.}
\end{table}

We ran a multi‑transcript sweep with the Text First Top setup on ten Congressional Record transcripts. For each transcript we sampled $9$ passages of sentence lengths $L = \{1,2,3,5,10\}$ in $4$ separate runs, with only $1$ try per passage. That’s $n{=}180$ passages per transcript (and $n{=}1800$ overall when pooling all ten), which is enough to put tight error bars around the mean. When sampling from the transcript each transcript-run pair used a separate random seed to help improve generalization.

\begin{table}[H]
\centering
\label{tab:total_trials}
\begin{tabular}{lrrrrr}
\toprule
 & \textbf{Transcripts} & \textbf{Passages} & \textbf{Tries per set} & \textbf{Runs} & \textbf{Total trials} \\
\midrule
Per configuration & 10 & 45 & 1 & 4 & 1800 \\
\bottomrule
\end{tabular}
\caption{Total trials per configuration in the run-to-run stability experiment.}
\end{table}

\subsubsection{Exact Match Accuracy}

We start with the pooled accuracy across all ten transcripts, and then break it out by sentence length and transcript third. Error bars are tight overall (about $\pm 0.8$ pp), with predictable variation by position in the transcript; early segments are easiest, the middle is slightly harder.

\begin{table}[H]
\centering
\caption{Overall accuracy across transcripts (Text First Top).}
\begin{tabular}{lrrrr}
\toprule
\textbf{Scope} & \textbf{n (questions)} & \textbf{Accuracy} & \textbf{95\% CI} \\
\midrule
Overall (all transcripts) & 1{,}800 & 97.1\% & [96.3, 97.9]\% \\
House                      &   900   & 95.7\% & [94.3, 97.0]\% \\
Senate                     &   900   & 98.6\% & [97.8, 99.3]\% \\
\bottomrule
\end{tabular}
\end{table}

Accuracy is high and consistent across sentence lengths (1–10 sentences), with small, non‑systematic differences. 

\begin{table}[H]
\centering
\caption{Accuracy by sentence length (overall).}
\begin{tabular}{rrrr}
\toprule
\textbf{Length} & \textbf{n} & \textbf{Accuracy} & \textbf{95\% CI} \\
\midrule
1   & 360 & 95.0\% & [92.7, 97.3]\% \\
2   & 360 & 97.8\% & [96.3, 99.3]\% \\
3   & 360 & 97.8\% & [96.3, 99.3]\% \\
5   & 360 & 97.2\% & [95.5, 98.9]\% \\
10  & 360 & 97.8\% & [96.3, 99.3]\% \\
\bottomrule
\end{tabular}
\end{table}

By transcript region, the first third is easiest (~99\%), the middle third dips modestly (~95.4\%), and the last third rebounds (~97.0\%). Note that while the passage finding algorithm attempts to find a unique passage of the given length by \emph{starting} in a given third of the transcript, in rare cases it wanders outside of that third to ensure it meets all conditions (unique and of the given length, with each sentence at least 20 characters long).

\begin{table}[H]
\centering
\caption{Accuracy by transcript third (overall).}
\begin{tabular}{lrrr}
\toprule
\textbf{Region} & \textbf{n} & \textbf{Accuracy} & \textbf{95\% CI} \\
\midrule
First third   & 589 & 99.0\% & [98.2, 99.8]\% \\
Middle third  & 612 & 95.4\% & [93.8, 97.1]\% \\
Last third    & 599 & 97.0\% & [95.6, 98.4]\% \\
\bottomrule
\end{tabular}
\end{table}

\subsubsection{Latency}

Average latency is also presented for readers who are interested. Short, single‑sentence queries show higher average latency (likely fixed overhead dominates), while lengths 2–5 are faster on average.

\begin{table}[H]
\centering
\caption{Latency (milliseconds). Overall includes p90; bins report mean.}
\begin{tabular}{lrrr}
\toprule
\textbf{Group} & \textbf{n} & \textbf{Avg ms}\\
\midrule
Overall             & 1{,}800 & 9{,}725\\
\midrule
Length 1            & 360 & 14{,}913 \\
Length 2            & 360 & 8{,}436 \\
Length 3            & 360 & 7{,}116 \\
Length 5            & 360 & 8{,}107 \\
Length 10           & 360 & 10{,}053 \\
\midrule
First third         & 589 & 7{,}158 \\
Middle third        & 612 & 10{,}173 \\
Last third          & 599 & 11{,}792 \\
\bottomrule
\end{tabular}
\end{table}

%------------------------------------
\subsubsection{Interpretation (strict) vs.\ expectation (relaxed)}
\label{sec:ci-interpretation-relaxed}

The confidence intervals (CIs) in Section~\ref{sec:gemini-flash-confidence-intervals} were calculated using $n=1800$ distinct passages
($S=9$ passages $\times$ $|L|=5$ lengths $\times$ $R=4$ runs $\times$ $T=10$ transcripts), pooled across transcripts and runs.
Index transcripts by $j=1,\dots,T$, runs by $r=1,\dots,R$, lengths by $\ell\in L=\{1,2,3,5,10\}$, and passages-within-length by $s=1,\dots,S$.
For each passage we run $N_{\text{try}}=1$ trial, so the per-passage accuracy is
\[
\hat{p}_{j r \ell s}
= \frac{1}{N_{\text{try}}}\sum_{t=1}^{N_{\text{try}}} x_{j r \ell s t}
= x_{j r \ell s 1},\qquad x_{j r \ell s t}\in\{0,1\}.
\]
The pooled mean accuracy is
\[
\hat{\mu}
= \frac{1}{n} \sum_{j=1}^{T} \sum_{r=1}^{R} \sum_{\ell\in L} \sum_{s=1}^{S} \hat{p}_{j r \ell s},
\qquad n = T R |L| S = 1800.
\]
The pooled standard error is
\[
SE
= \frac{\sigma_{\hat{p}}}{\sqrt{n}},\qquad
\sigma_{\hat{p}}
= \sqrt{\frac{1}{n-1}\sum_{j=1}^{T} \sum_{r=1}^{R} \sum_{\ell\in L} \sum_{s=1}^{S}
\left(\hat{p}_{j r \ell s}-\hat{\mu}\right)^2 }.
\]
The 95\% confidence interval is
\[
\hat{\mu} \pm 1.96 \times SE.
\]

Strictly speaking, this CI applies to exact match accuracy estimates using the same data-generating process used to build them:
(i) pick a CREC transcript,
(ii) transcribe with the same STT provider (AssemblyAI),
(iii) convert to Text First Top,
(iv) sample passages with the {same scheme
($ L \in\{1,2,3,5,10\}$; 9 passages balanced by transcript third),
(v) run Gemini 2.5 Flash with thinking on once per passage,
(vi) repeat with independent passage draws (4 runs/transcript),
and (vii) pool passage-level outcomes across transcripts.
The CI then provide a range of expected exact match accuracies answering the question:
``If you rerun this exact recipe on a new but similar transcript, where
should the accuracy land?'' These intervals are wider than the
within-transcript spreads because they include both within-day sampling
variation (different passages) and across-day difficulty differences.

However, you will note that exact match point estimates for Gemini 2.5 Flash in other parts of this paper, fall within, or very close to, this same range despite using a slightly different data generating process. More casually, you can think of the CIs the way a
practitioner would: ``Given a new CREC transcript, using the same Text First Top + thinking recipe, expect exact
matches in the mid‑90s; a good single‑number guess is about 97\%, with a
few percentage points of wiggle room.'' That relaxed
summary lines up with the point values we saw elsewhere in the paper:
the single‑transcript thinking sweep in
Section~\ref{sec:thinking-budget} reports 95.6\%, 96.1\%, and 93.9\% and the broader exact‑match results in
Section~\ref{sec:exact-match-timestamp-results} keep the Text First Top
condition in that same band.

\subsubsection{Assumptions for Extending to Other Transcripts}
The confidence intervals in Section~\ref{sec:gemini-flash-confidence-intervals} apply to the same
data-generating process we used to construct them. Extending these intervals beyond that setting
requires a few concrete assumptions:

\begin{enumerate}[itemsep=2pt,topsep=2pt,leftmargin=1.25em]

\item \textbf{Transcripts must be similar.} The transcripts in our sample are drawn from the \emph{Congressional Record} (CREC). It is likely that non-CREC transcripts that vary significantly in tone, topic, length, complexity, or other characteristics will perform differently. On the other hand, it is reasonable to expect that other CREC transcripts will perform similarly, since we intentionally included a diverse sample that spaned decades, included both long (~45k tokens) and short ( ~800k tokens) transcripts, with an even distribution across chambers (House and Senate). However, outlier CREC transcripts could in principle fall outside this range.

\item \textbf{Speech-to-text quality must be comparable.} All transcripts were generated using AssemblyAI. If another provider produces markedly noisier or cleaner output, measured accuracy could shift, since the benchmark task depends directly on sentence segmentation and lexical fidelity.

\item \textbf{Passage lengths must remain in range.} The evaluation was limited to passages of length $1$--$10$ consecutive sentences. The intervals may extend to slightly longer spans, but they should not be assumed valid for passages that are dozens or hundreds of sentences long.

\item \textbf{Enough passages must be averaged.} The reported confidence intervals were calculated at scale ($n=1800$ pooled passages). With a much smaller evaluation sample, results may deviate from the $\sim97\%$ overall accuracy (mid-90s). Averaging across more passages will stabilize performance within that range.

\end{enumerate}

Most importantly, note that these CIs apply to the exact match task accuracy, a task that is best suited for standard matching algorithms rather than LLMs. The exact match task is used as a benchmark for the more realistic fuzzy matching task in Section~\ref{sec:fuzzy}.

\subsection{Exact and Fuzzy Retrieval Accuracy Across Transcript Lengths}
To measure how retrieval accuracy scales with transcript length, we constructed a synthetic dataset of nearly one million tokens by concatenating three complete Congressional Record sessions: 
\begin{enumerate}[itemsep=0pt,topsep=2pt]
    \item The March~31--April~1, 2025, Senate filibuster led by Cory Booker (D--NJ), a 25-hour session in which he frequently yielded for questions and comments from other members;
    \item A July~3, 2025 House session that included an 8-hour floor speech from Hakeem Jeffries (D--NY);
    \item A standard House session on July~17, 2025.
\end{enumerate}
Full sentences were retained, and we trimmed at bucket boundaries to come as close as possible to target cutoffs. Buckets were generated at 50{,}000-token intervals up to 950{,}000 tokens, though for cost reasons we report results only for the 100k--900k range.

The claimed context limit for Gemini 2.5 Flash is 1{,}048{,}576 tokens. We therefore extended the sweep up to 900{,}000 tokens, covering most of the usable range while keeping the experiment computationally and financially tractable. Constructing each bucket required appending and trimming the composite transcript to the desired length, producing token counts that are within a few dozen tokens of the targets.

\subsubsection{Experimental Setup}
The context length sweep was structured as follows. The experimental setup was largely driven by cost, given the scale of the test: at these lengths, each run costs approximately \$70. Running five replicates across multiple models, passage lengths, reasoning settings, and query placements would have been cost-prohibitive. Accordingly, we adopted a lighter setup than usual to balance rigor with feasibility.

\begin{itemize}[itemsep=4pt, topsep=2pt]

    \item \textbf{Nine buckets.} We tested nine transcript lengths in 100k increments, from 100k up to 900k tokens.

    \item \textbf{Standard sentence lengths.} Within each bucket we evaluated the standard five sentence lengths: $L \in \{1,2,3,5,10\}$.

    \item \textbf{Only six passages per length.} Each length used \emph{six} non-overlapping sentence sets (rather than the standard twelve), with the usual three independent trials per passage.

    \item \textbf{Text Firt Top.} We fixed the prompt to the best-performing design identified earlier (Text First Top).

    \item \textbf{Reasoning tokens.} Thinking was explicitly disabled to avoid additional costs from reasoning output tokens.

    \item \textbf{Independent runs.} For the configuration outlined in the above bullets, we conducted two complete runs, each covering all buckets and sentence lengths. Unlike earlier experiments in this paper, which typically reused the same passages across conditions to maximize comparability, here we intentionally treated the two runs as independent samples. Each run drew its own sets of sentences within each bucket. This design provided a stronger estimate of variance, but at the expense of direct one-to-one comparability across runs. This design yields 90 tests per bucket per run, or 180 tests per bucket across two runs, for a total of 1{,}620 trials. This setup was repeated for both exact match and fuzzy match tasks (two runs for exact match and two runs for fuzzy match).

    \item \textbf{Exact and Fuzzy match tasks.} Both exact match and fuzzy match conditions were tested under the same setup, yielding four total runs (two per mode).

\end{itemize}

\subsubsection{Results}
Figure~\ref{fig:flash25-length-sweep} summarizes the averaged performance across the two runs for each task. Accuracy for the exact match task is generally higher than for the fuzzy match task, except at 500k input tokens, though this is likely due to random model variation than a repeatable artifact.

Accuracy is stable through the mid-range, with exact-match scores above 95\% through 400k input tokens and fuzzy match about 85\%. Beyond this, exact match performance declines gradually: 87\% at 500k input tokens and 84\% at 900k. Fuzzy match performance declines after 500k input tokens, declining to 69\% accuracy at 900k.

Though not shown on the chart, adjusted accuracy for exact match (crediting off-by-one predictions) remains above 94\% through 800k, then dips below 90\% at 900k. A similar pattern is observed for fuzzy match, with adjusted accuracy above 92\% until 900k, at which point it declines to 76\%.

Latency and cost increase nearly linearly with context length, with both doubling from the 400k to the 800k bucket. In sum, Gemini 2.5 Flash sustains high retrieval accuracy across most of its advertised 1M-token context, but begins to show degradation in the upper range.

\begin{figure}[H]
    \centering
    \includegraphics[width=0.8\textwidth]{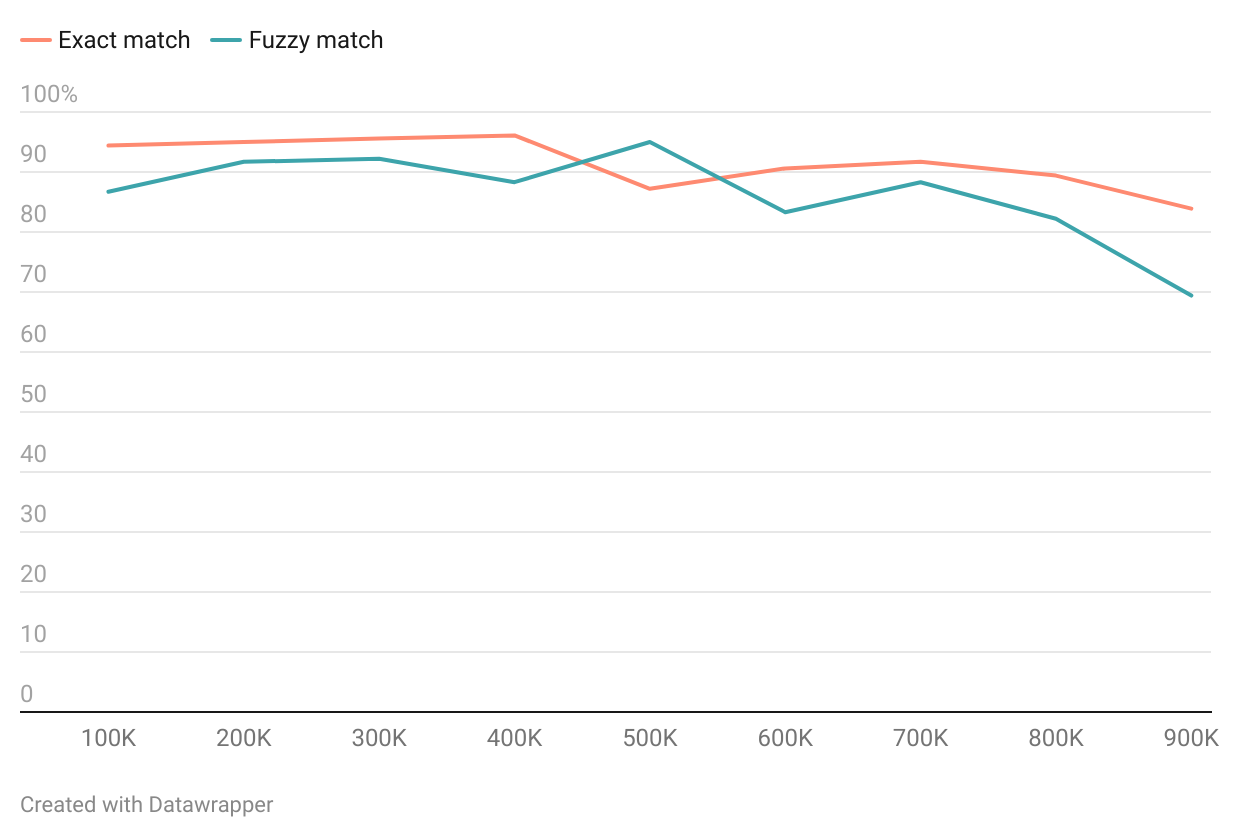}
    \caption{Retrieval accuracy of Gemini 2.5 Flash across transcript lengths from 100k to 900k tokens. Accuracy is reported as exact match and fuzzy match over two independent runs. Performance remains stable through 400k tokens, after which both metrics begin to degrade, with fuzzy match showing earlier volatility.}
    \label{fig:flash25-length-sweep}
\end{figure}

%╔══════════════════════════════════════════════════════════════════════════════╗
%║  ███╗   ██╗███████╗██╗    ██╗    ███████╗███████╗ ██████╗                    ║
%║  ████╗  ██║██╔════╝██║    ██║    ██╔════╝██╔════╝██╔════╝                    ║
%║  ██╔██╗ ██║█████╗  ██║ █╗ ██║    ███████╗█████╗  ██║                         ║
%║  ██║╚██╗██║██╔══╝  ██║███╗██║    ╚════██║██╔══╝  ██║                         ║
%║  ██║ ╚████║███████╗╚███╔███╔╝    ███████║███████╗╚██████╗                    ║
%║  ╚═╝  ╚═══╝╚══════╝ ╚══╝╚══╝     ╚══════╝╚══════╝ ╚═════╝                    ║
%╚══════════════════════════════════════════════════════════════════════════════╝

\section{Full Evaluation: Needle Controls (Task 1)}

This section presents the results of the full set of tests as described in \ref{sec:call-budget-and-task-details} across a range of models.

\subsection{Needle-in-a-Haystack Control Experiments}

To ground our timestamp-retrieval experiments and enable comparison with prior long-context Q\&A literature, we designed a set of control questions that require the model to retrieve key facts from different parts of the Congressional Record transcript. These controls serve three goals:  
\begin{enumerate}[itemsep=2pt,topsep=2pt,leftmargin=14pt]
  \item Test whether models still excel in factual retrieval when transcript formats include timestamps.  
  \item Provide a simple directional baseline for comparison against the main timestamp-retrieval tasks.  
  \item Bridge results to prior long-context Q\&A work, enabling apples-to-apples interpretation across studies.  
\end{enumerate}

\subsection{Control Question Setup and Evidence}
\label{sec:control-question-setup}
Each control question was crafted to reflect the classic ``needle-in-a-haystack'' retrieval paradigm and was mapped to a specific transcript segment (first, middle, or last third):

\begin{enumerate}
    \item \textbf{Speaker Pro Tempore (first third):}\\
    \textit{Who is appointed to act as Speaker Pro Tempore, and what is the date of the proceedings?}
    
    \medskip
    \textbf{Evidence:}
    \begin{jsoncode}
[
  {"start_ms": 42890, "end_ms": 45200, "text": "The speaker's rooms Washington, D.C."},
  {"start_ms": 45340, "end_ms": 47370, "text": "june 11, 2025."},
  {"start_ms": 47690, "end_ms": 52970, "text": "I hereby appoint the Honorable Rudy Yakim III to act as Speaker Pro Tempore on this day."}
]
    \end{jsoncode}

    \textbf{Pydantic output format and correct answer:}
    \begin{pythoncode}
class SpeakerProTempore(BaseModel):
    name: str
    date: str
# Correct answer:
# {'name': 'Rudy Yakim III', 'date': 'June 11, 2025'}
    \end{pythoncode}

    \item \textbf{Funding Amount (middle third):}\\
    \textit{What is the amount mentioned for grant funding to combat the opioid crisis?}

    \medskip
    \textbf{Evidence:}
    \begin{jsoncode}
[
  {"start_ms": 5402480, "end_ms": 5426560, 
   "text": "...halting over 8.6 billion in grant funding for the federal programs to combat the opioid crisis."}
]
    \end{jsoncode}

    \textbf{Pydantic output format and correct answer:}
    \begin{pythoncode}
class FundingAmount(BaseModel):
    amount: float
    units: Literal["hundreds", "thousands", "millions", "billions", "trillions"]
# Correct answer:
# {'amount': 8.6, 'units': 'billions'}
    \end{pythoncode}

    \item \textbf{Solar Ranking (last third):}\\
    \textit{What was the United States ranking in solar panel manufacturing in 2017, and what is it today?}

    \medskip
    \textbf{Evidence:}
    \begin{jsoncode}
[
  {"start_ms": 14338910, "end_ms": 14344270, 
   "text": "In 2017, the United States ranked 14th in the world for solar panel manufacturing."},
  {"start_ms": 14344750, "end_ms": 14345870, 
   "text": "Today we're third."}
]
    \end{jsoncode}

    \textbf{Pydantic output format and correct answer:}
    \begin{pythoncode}
class SolarRanking(BaseModel):
    rank_2017: int
    rank_today: int
# Correct answer:
# {'rank_2017': 14, 'rank_today': 3}
    \end{pythoncode}

\end{enumerate}
We ran three rounds of control experiments on GPT-5–series and Gemini-2.5 models, using different transcript formats. Additionally, in an early set of pilot tests, an additional two rounds on a broader set of models.

In all cases, the needle control question preceded the transcript at the top of the prompt, responses were constrained to a standardized JSON/Pydantic schema (see Section~5.1), and each question was tested three times per model with fresh context windows.

\subsubsection{Full vs. Pilot Tests}
The main test setup evaluated three transcript formats on GPT-5–series and Gemini 2.5 models: 

\begin{itemize}
      \item \textbf{JSON:} Full transcript, structured as a list of sentences with \texttt{start\_ms}, \texttt{end\_ms}, and \texttt{text}.
    \item \textbf{Text-Only:} Plain concatenated transcript text with all timestamp markers removed.
    \item \textbf{Text First Top:} Transcript transformed into a compact text layout that places sentence text before millisecond timestamps (\S\ref{sec:text-field-order-variations}), reducing token count while preserving sentence boundaries.
\end{itemize}

There was also an earlier pilot setup that evaluated only the first two formats. Both setups used the same three control questions (Funding Amount, Speaker Pro Tempore, Solar Ranking), enabling direct comparison across models. See Table~\ref{tab:models-used} for a complete summary.

\subsubsection{Reasoning configuration}
For the new tests, the OpenAI GPT-5 series were explicitly set to use medium reasoning level, while Google Gemini was set to use dynamic thinking. For the pilot test reasoning was enabled for some models and disabled for others. See Section~\ref{sec:reasoning-setup}.

\subsection{Background and Related Work}
\label{sec:needles-literature-review}

Prior work has consistently shown that long-context retrieval presents distinct challenges for LLMs, particularly with respect to the position of the target (``needle'') and the presence of distractors (``needles in a haystack''). Studies such as \citet{liu2024lost} documented a U-shaped performance curve, with information in the middle of the context being hardest to retrieve. \citet{he2024neverlost} and \citet{kuratov2024babilong} also found that facts placed away from the start of the context led to degraded accuracy, and that LLMs effectively use only a fraction of their nominal context window. \citet{hsieh2024ruler} observed that distractor entities in the haystack further reduce retrieval performance, especially for numerical or repetitive facts. Conversely, \citet{wang2025reasoning} noted that for multi-hop reasoning tasks, position sensitivity can be less pronounced. Notably, model providers (e.g., \citet{openai2024gpt41}) have claimed near-perfect recall for needles at all positions within million-token contexts, but real-world retrieval accuracy remains variable.

\subsection{Needle-in-a-Haystack Control Results}
\label{sec:results}

\subsubsection{Results}
New runs show consistently high accuracy across all formats, likely due to a combination of stronger models (e.g., the GPT-5 family of models) and explicitly setting the models' reasoning effort to "medium" as opposed to leaving the default reasoning settings.

\begin{table}[H]
\centering
\small
\begin{tabular}{lccc cc}
\toprule
\textbf{Question} & \multicolumn{3}{c}{\textbf{New test setup}} & \multicolumn{2}{c}{\textbf{Early test setup}} \\
\cmidrule(lr){2-4} \cmidrule(lr){5-6}
 & JSON & Text-Only & Text First Top & JSON & Text \\
\midrule
Funding Amount & 100.0\% & 66.7\% & 100.0\% & 47.2\% & 50.0\% \\
Speaker Pro Tempore & 100.0\% & 94.4\% & 94.4\% & 61.1\% & 75.0\% \\
Solar Ranking & 100.0\% & 100.0\% & 100.0\% & 94.4\% & 94.4\% \\
\midrule
\textbf{Overall} & 100.0\% & 87.0\% & 98.1\% & 67.6\% & 73.1\% \\
\bottomrule
\end{tabular}
\caption{Format comparison across the three control questions.}
\label{tab:format-comparison}
\end{table}

New runs show consistently high accuracy across all formats. JSON Control achieves ceiling-level accuracy, 
Text-Only is slightly lower but faster, and Text First Top performs nearly at ceiling across all three questions. Given only three trials per model, differences between formats likely fall within confidence intervals (\S\ref{sec:gemini-flash-confidence-intervals}). Importantly, the text-only format doesn't perform better than the JSON or Text First Top formats, showing that for factual retrieval the presence of timestamps does not impede performance.

JSON, Text First Top, and Text-Only per-model results respectively in Tables \ref{tab:new-json-control-accuracy}, \ref{tab:new-optimized-text-control-accuracy}, \ref{tab:new-text-only-control-accuracy}. Larger models tend to achieve ceiling accuracy, but smaller models also performed well. The Funding Amount question was the most difficult for smaller models. This was likely due to the presence of distractors ({\S\ref{sec:needles-literature-review}). While only one sentence provides the correct answer of \$8.6 billion in grant funding, there were 8 mentions of \$9.4 billion in rescissions. In all cases where the models were incorrect, their answer was \$9.4 billion. A full replication of factual retrieval is beyond the scope of this paper so we do not discuss the impact of distractors further.

\begin{table}[H]
\centering
\small
\begin{tabular}{lccc}
\toprule
\textbf{Model} & Funding Amount & Speaker Pro Tempore & Solar Ranking \\
\midrule
Gemini-2.5 Flash & 100.0\% & 100.0\% & 100.0\% \\
Gemini-2.5 Flash-Lite & 33.3\% & 100.0\% & 100.0\% \\
Gemini-2.5 Pro & 100.0\% & 100.0\% & 100.0\% \\
GPT-5 & 100.0\% & 100.0\% & 100.0\% \\
GPT-5 mini & 100.0\% & 100.0\% & 100.0\% \\
GPT-5 nano & 100.0\% & 100.0\% & 100.0\% \\
\midrule
\textbf{Total} & 88.9\% & 100.0\% & 100.0\% \\
\bottomrule
\end{tabular}
\caption{Per-model accuracy for JSON Control (target-first). Each cell shows fraction correct out of three trials.}
\label{tab:new-json-control-accuracy}
\end{table}

\begin{table}[H]
\centering
\small
\begin{tabular}{lccc}
\toprule
\textbf{Model} & Funding Amount & Speaker Pro Tempore & Solar Ranking \\
\midrule
Gemini-2.5 Flash & 100.0\% & 100.0\% & 100.0\% \\
Gemini-2.5 Flash-Lite & 100.0\% & 66.7\% & 100.0\% \\
Gemini-2.5 Pro & 100.0\% & 100.0\% & 100.0\% \\
GPT-5 & 100.0\% & 100.0\% & 100.0\% \\
GPT-5 mini & 100.0\% & 100.0\% & 100.0\% \\
GPT-5 nano & 100.0\% & 100.0\% & 100.0\% \\
\midrule
\textbf{Total} & 100.0\% & 94.4\% & 100.0\% \\
\bottomrule
\end{tabular}
\caption{Per-model accuracy for Text First Top (target-first). Each cell shows fraction correct out of three trials.}
\label{tab:new-optimized-text-control-accuracy}
\end{table}

\begin{table}[H]
\centering
\small
\begin{tabular}{lccc}
\toprule
\textbf{Model} & Funding Amount & Speaker Pro Tempore & Solar Ranking \\
\midrule
Gemini-2.5 Flash & 66.7\% & 100.0\% & 100.0\% \\
Gemini-2.5 Flash-Lite & 0.0\% & 66.7\% & 100.0\% \\
Gemini-2.5 Pro & 100.0\% & 100.0\% & 100.0\% \\
GPT-5 & 100.0\% & 100.0\% & 100.0\% \\
GPT-5 mini & 100.0\% & 100.0\% & 100.0\% \\
GPT-5 nano & 33.3\% & 100.0\% & 100.0\% \\
\midrule
\textbf{Total} & 66.7\% & 94.4\% & 100.0\% \\
\bottomrule
\end{tabular}
\caption{Per-model accuracy for Text-Only (target-first). Each cell shows fraction correct out of three trials.}
\label{tab:new-text-only-control-accuracy}
\end{table}

Text formats provide substantial reductions in latency—commensurate with their reduction in total token size—without sacrificing accuracy. JSON Control achieves ceiling-level accuracy but at significantly higher cost and latency. By contrast, both Text-Only and Text First Top are 2.7–2.9$\times$ faster while matching JSON accuracy. Note that these figures include the automatic, if sometimes inconsistent, input caching of GPT-5 and Gemini models.

\begin{table}[H]
\centering
\small
\begin{tabular}{lccc}
\toprule
\textbf{Format} & \textbf{Accuracy} & \textbf{Avg Time (ms)} & \textbf{Relative Speed} \\
\midrule
JSON Control & 96.3\% & 33{,}178 & 1.0x \\
Text-Only & 87.0\% & 12{,}280 & 2.7x faster \\
Text First Top & 98.1\% & 11{,}413 & 2.9x faster \\
\bottomrule
\end{tabular}
\caption{Accuracy and latency comparison across formats in new (target-first) runs.}
\label{tab:latency}
\end{table}

For comparison, Table~\ref{tab:needles-early-results} presents results from the earlier evaluation pass. In line with the newer results presented above, larger models tend to perform better than smaller models, while OpenAI's legacy models generally perform worse than the newer GPT-5 models.

\begin{table}[H]
\centering
\small
\begin{tabular}{lcc cc ccc}
\toprule
\textbf{Question} & \multicolumn{2}{c}{Funding Amount} & \multicolumn{2}{c}{Speaker Pro Tempore} & \multicolumn{2}{c}{Solar Ranking} \\
\cmidrule(lr){2-3} \cmidrule(lr){4-5} \cmidrule(lr){6-7}
 & JSON & Text & JSON & Text & JSON & Text \\
\midrule
Claude-3.5-Haiku & 66.7\% & 33.3\% & 100.0\% & 100.0\% & 100.0\% & 100.0\% \\
Claude-3.7-Sonnet & 33.3\% & 33.3\% & 100.0\% & 100.0\% & 100.0\% & 100.0\% \\
Claude-Sonnet-4.0 & 0.0\% & 33.3\% & 0.0\% & 33.3\% & 33.3\% & 66.7\% \\
Gemini-2.5 Flash & 100.0\% & 100.0\% & 100.0\% & 100.0\% & 100.0\% & 100.0\% \\
Gemini-2.5 Flash-Lite & 0.0\% & 0.0\% & 0.0\% & 0.0\% & 100.0\% & 100.0\% \\
Gemini-2.5 Pro & 100.0\% & 100.0\% & 0.0\% & 0.0\% & 100.0\% & 100.0\% \\
GPT-4.1 & 33.3\% & 66.7\% & 66.7\% & 100.0\% & 100.0\% & 100.0\% \\
GPT-4.1 mini & 100.0\% & 33.3\% & 66.7\% & 100.0\% & 100.0\% & 100.0\% \\
GPT-4.1 nano  & 0.0\% & 0.0\% & 100.0\% & 66.7\% & 100.0\% & 100.0\% \\
o3 & 100.0\% & 100.0\% & 100.0\% & 100.0\% & 100.0\% & 100.0\% \\
o3-Mini & 33.3\% & 100.0\% & 0.0\% & 100.0\% & 100.0\% & 100.0\% \\
o4-Mini & 0.0\% & 0.0\% & 100.0\% & 100.0\% & 100.0\% & 66.7\% \\
\midrule
\textbf{Total} & 47.2\% & 50.0\% & 61.1\% & 75.0\% & 94.4\% & 94.4\% \\
\bottomrule
\end{tabular}
\caption{Per-model accuracy on each control question in the early pilot evaluation pass. Each cell shows fraction correct out of three trials.}
\label{tab:needles-early-results}
\end{table}

\subsection{Comparison with Prior Findings}

Our results both align with and diverge from prior studies. The ``Funding Amount'' question, positioned in the middle third and containing multiple numerical distractors (\$9.4 billion was mentioned eight times), exhibited the lowest retrieval accuracy—consistent with \citet{hsieh2024ruler} and ``lost-in-the-middle'' effects reported by \citet{liu2024lost} and \citet{kuratov2024babilong}. However, the ``Solar Ranking'' fact—placed at the end of the context—was retrieved with perfect accuracy by most models, somewhat at odds with the strictest interpretations of middle/end-of-context degradation. The ``Speaker Pro Tempore'' question (first third) proved moderately challenging.

Interestingly, provider claims of universal ``needle'' recall across context positions did not always hold in these real-world tasks. Models such as GPT-4.1, Claude, and Gemini occasionally failed, with accuracy sometimes sensitive to input format (see Table~\ref{tab:needles-early-results}). This echoes literature noting that published benchmarks and internal provider claims may not generalize to all retrieval settings.

%╔══════════════════════════════════════════════════════════════════════════════╗
%║  ███╗   ██╗███████╗██╗    ██╗    ███████╗███████╗ ██████╗                    ║
%║  ████╗  ██║██╔════╝██║    ██║    ██╔════╝██╔════╝██╔════╝                    ║
%║  ██╔██╗ ██║█████╗  ██║ █╗ ██║    ███████╗█████╗  ██║                         ║
%║  ██║╚██╗██║██╔══╝  ██║███╗██║    ╚════██║██╔══╝  ██║                         ║
%║  ██║ ╚████║███████╗╚███╔███╔╝    ███████║███████╗╚██████╗                    ║
%║  ╚═╝  ╚═══╝╚══════╝ ╚══╝╚══╝     ╚══════╝╚══════╝ ╚═════╝                    ║
%╚══════════════════════════════════════════════════════════════════════════════╝

\section{Full Evaluation: Exact Match Timestamp Retrieval (Task 2)}
\label{sec:timestamps}
Having confirmed through our control experiments that general factual retrieval remains strong—even in timestamped transcript formats, and broadly consistent with the high recall rates advertised by model providers—we now turn to the first of our two timestamp retrieval tasks: exact match. In this task, models are presented with passages of consecutive sentences from the transcript and must return the precise start and end timestamps (in milliseconds) associated with that passage.

\subsection{Task Setup}
\label{sec:timestamps:setup}
We evaluate timestamp retrieval under three transcript formats:

\begin{itemize}
  \item \textbf{JSON Transcript (Query Bottom):} Full transcript, structured as a list of sentences with \texttt{start\_ms}, \texttt{end\_ms}, and \texttt{text}, with the query appearing \emph{before} the transcript.
  \item \textbf{JSON Transcript (Query Top):} Same as above, but with the query appearing \emph{after} the transcript.
  \item \textbf{Text First Top:} Transcript transformed into a compact text layout that places sentence text before millisecond timestamps, reducing token count while preserving sentence boundaries. See \ref{sec:optimized-text-format-description-basic} for details.
\end{itemize}

As a reminder, we construct the exact-match experiment by randomly selecting passages of $\ell$ consecutive sentences. We consider lengths $\ell \in L$, where $L=\{1,2,3,5,10\}$. For each length of consecutive sentences $\ell$, we generate twelve random sets distributed across the transcript: four drawn from the first third, four from the middle third, and four from the last third.

Each passage is passed to the model three times, independently, with a fresh context window. Thus, the total number of trials per model is $12 \times 3 \times 5 = 180$ (12 passages, 3 tries, 5 lengths of consecutive sentences).

Each set of consecutive sentences is associated with a particular timespan within the transcript given by the start and end millisecond identifiers. A prediction is \emph{exact} when both predicted start and end match the target sentence boundaries. Off-by-one errors (start, end, or both) are also recorded.

\subsection{Exact Match Timestamp Retrieval Results}
\label{sec:exact-match-timestamp-results}

Table~\ref{tab:exact-timestamp-by-length} reports exact-match accuracy by passage length across formats. Four patterns stand out.
First, consistent with Gemini 2.5 Flash testing, placing the query \emph{before} the transcript improves accuracy by roughly 3–15 percentage points (pp), depending on passage length.
Second, the Text First Top format outperforms JSON by about 1–15pp.
Third, JSON with the target placed at the bottom remains uniformly difficult across all lengths, hovering in the mid-60s.
Fourth, both formats with the target at the top exhibit an inverted U-shape: accuracy is highest for passages of length 2–5 sentences, but dips for very short (1 sentence) and very long (10 sentence) passages. A likely explanation is that single sentences are less distinctive (and thus easier to confuse with near matches), while long spans require maintaining coherence across broader timestamp ranges; the precise mechanism remains unclear. This U-shape is most visible in smaller models such as GPT-5 nano and Gemini 2.5 Flash-Lite. As Figure~\ref{fig:model-accuracy-on-exact-match-optimized-text} shows, other models also perform slightly worse at lengths 1 and 10, but the curve is shallow, and for some, perfect scores at these extremes may fall within natural variation.

\begin{table}[H]
\centering
\small
\renewcommand{\arraystretch}{1.15}
\begin{tabular}{lccc}
\toprule
\multirow{2}{*}{\textbf{Target Length}} 
  & \multicolumn{1}{c}{\textbf{Text First Top}} 
  & \multicolumn{2}{c}{\textbf{JSON}} \\
\cmidrule(lr){2-2}\cmidrule(lr){3-4}
  & \textbf{Query Bottom} & \textbf{Query Bottom} & \textbf{Query Top} \\
\midrule
1 sentence   & 75.9\% & 66.2\% & 63.4\% \\
2 sentences  & 93.1\% & 77.8\% & 62.0\% \\
3 sentences  & 90.7\% & 81.0\% & 65.3\% \\
5 sentences  & 86.1\% & 81.5\% & 65.7\% \\
10 sentences & 76.4\% & 75.9\% & 63.4\% \\
\bottomrule
\end{tabular}
\caption{Exact-match accuracy by span length across formats (all models aggregated).}
\label{tab:exact-timestamp-by-length}
\end{table}

Even state-of-the-art models are not flawless. Despite the Text First Top format being short (less than 120k tokens and only 30\% of GPT-5’s 400k context window) and the task being unambiguous (exact sentence matching), GPT-5 failed to identify timestamps in some instances. For example, in all three of its attempts it misidentified the timestamps for the sentence, ``I asked to address the House for one minute and then to revise and extend my remarks that,''; though in each case at least one of its timestamps was only off by one. Notably, in the JSON format with the query on top, GPT-5 accurately returned the correct start and end timestamps in all three tries. This demonstrates the natural variability of LLMs still affects even state-of-the-art models at times.

\begin{figure}[H]
    \centering
    \includegraphics[width=0.8\textwidth]{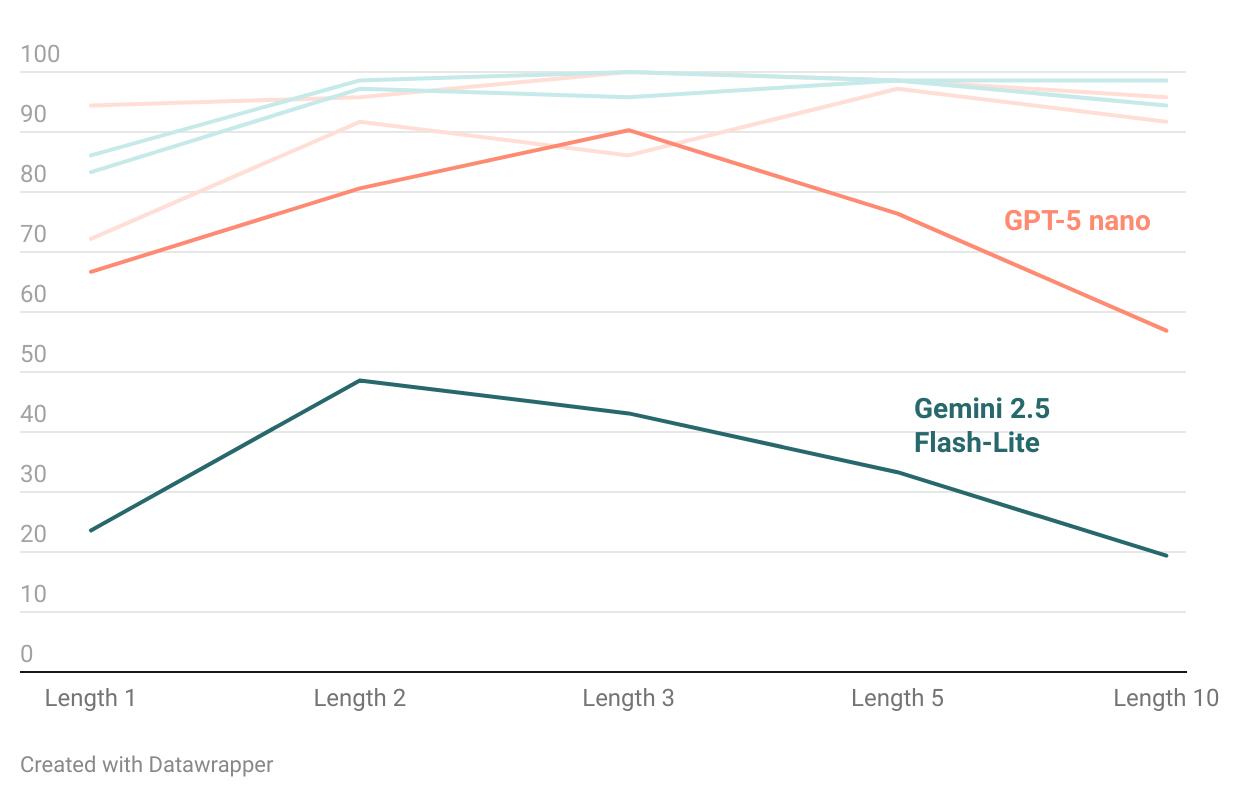}
    \caption{Accuracy of the six models across sentence lengths.}
    \label{fig:model-accuracy-on-exact-match-optimized-text}
\end{figure}

The poor performance of the JSON format with the query at the bottom was largely driven by the particularly poor performance of Google Gemini 2.5 Flash-Lite which had accuracy rates of only between 3\% and 11\% across sentence lengths. If Gemini 2.5 Flash-Lite is removed and averages are recalculated across the remaining 5 models, the JSON format with the query at the bottom is comparable to the JSON performance with the query at the top.

\begin{table}[H]
\centering
\small
\renewcommand{\arraystretch}{1.15}
\begin{tabular}{lccc}
\toprule
\multirow{2}{*}{\textbf{Target Length}} 
  & \multicolumn{3}{c}{\textbf{JSON}} \\
\cmidrule(lr){2-4}
  & \textbf{Query Bottom} & \textbf{Query Top (no Flash-Lite)} & \textbf{Query Top} \\
\midrule
1 sentence   & 66.2\% & 75.0\% & 63.4\% \\
2 sentences  & 77.8\% & 73.9\% & 62.0\% \\
3 sentences  & 81.0\% & 76.1\% & 65.3\% \\
5 sentences  & 81.5\% & 77.8\% & 65.7\% \\
10 sentences & 75.9\% & 73.9\% & 63.4\% \\
\bottomrule
\end{tabular}
\caption{Exact match accuracy by passage length for JSON format, comparing Query Bottom, Query Top (with Flash-Lite removed), and Query Top (all models).}
\label{tab:json-timestamp-by-length}
\end{table}

Performance by sentence length is reported for Text First Top in \autoref{tab:exact-opttext-target-first-by-length}, for JSON with the Query Top in \autoref{tab:json-target-before-by-length}, and for JSON with the Query Bottom in \autoref{tab:json-target-after-by-length}.

The results show a consistent advantage for the larger models: GPT-5 and Gemini-2.5 Pro remain near ceiling across all lengths, while their smaller counterparts degrade substantially as span length increases. GPT-5 mini maintains strong performance up to 5 sentences but drops at 10, whereas GPT-5 nano is less stable overall. Among the Gemini family, Flash performs well on short spans but falls off at 10 sentences, and the Lite variant is uniformly poor. The JSON target-after condition is the hardest: all models decline, yet GPT-5 retains the highest accuracy and Gemini-Pro remains steady, while Flash and especially Lite collapse on longer spans.

The tables below show the model performance by consecutive sentence length for the three tasks. 
\begin{table}[H]
\centering
\renewcommand{\arraystretch}{1.15}
\begin{tabular}{lccccc}
\toprule
\multirow{2}{*}{\textbf{Model}} 
  & \multicolumn{5}{c}{\textbf{Consecutive Sentence Length}} \\ 
\cmidrule(lr){2-6}
  & \textbf{1} & \textbf{2} & \textbf{3} & \textbf{5} & \textbf{10} \\
\midrule
Gemini-2.5 Flash & 91.7\% & 97.2\% & 97.2\% & 100.0\% & 100.0\% \\
Gemini-2.5 Flash-Lite & 38.9\% & 69.4\% & 52.8\% & 38.9\% & 22.2\% \\
Gemini-2.5 Pro & 80.6\% & 100.0\% & 100.0\% & 100.0\% & 100.0\% \\
GPT-5 & 88.9\% & 100.0\% & 100.0\% & 100.0\% & 94.4\% \\
GPT-5 mini & 80.6\% & 94.4\% & 100.0\% & 100.0\% & 88.9\% \\
GPT-5 nano & 75.0\% & 97.2\% & 94.4\% & 77.8\% & 52.8\% \\
\bottomrule
\end{tabular}
\caption{Per-model accuracy by passage length for Text First Top (Query Top).}
\label{tab:exact-opttext-target-first-by-length}
\end{table}

\begin{table}[H]
\centering
\renewcommand{\arraystretch}{1.15}
\begin{tabular}{lccccc}
\toprule
\multirow{2}{*}{\textbf{Model}} 
  & \multicolumn{5}{c}{\textbf{Consecutive Sentence Length}} \\ 
\cmidrule(lr){2-6}
  & \textbf{1} & \textbf{2} & \textbf{3} & \textbf{5} & \textbf{10} \\
\midrule
Gemini-2.5 Flash & 75.0\% & 97.2\% & 94.4\% & 97.2\% & 88.9\% \\
Gemini-2.5 Flash-Lite & 8.3\% & 27.8\% & 33.3\% & 27.8\% & 16.7\% \\
Gemini-2.5 Pro & 91.7\% & 97.2\% & 100.0\% & 97.2\% & 97.2\% \\
GPT-5 & 100.0\% & 91.7\% & 100.0\% & 97.2\% & 97.2\% \\
GPT-5 mini & 63.9\% & 88.9\% & 72.2\% & 94.4\% & 94.4\% \\
GPT-5 nano & 58.3\% & 63.9\% & 86.1\% & 75.0\% & 61.1\% \\
\bottomrule
\end{tabular}
\caption{Per-model accuracy by passage length for JSON (Query Top).}
\label{tab:json-target-before-by-length}
\end{table}

\begin{table}[H]
\centering
\renewcommand{\arraystretch}{1.15}
\begin{tabular}{lccccc}
\toprule
\multirow{2}{*}{\textbf{Model}} 
  & \multicolumn{5}{c}{\textbf{Consecutive Sentence Length}} \\ 
\cmidrule(lr){2-6}
  & \textbf{1} & \textbf{2} & \textbf{3} & \textbf{5} & \textbf{10} \\
\midrule
Gemini-2.5 Flash & 66.7\% & 63.9\% & 69.4\% & 69.4\% & 58.3\% \\
Gemini-2.5 Flash-Lite & 5.6\% & 2.8\% & 11.1\% & 5.6\% & 11.1\% \\
Gemini-2.5 Pro & 86.1\% & 86.1\% & 88.9\% & 86.1\% & 86.1\% \\
GPT-5 & 91.7\% & 91.7\% & 97.2\% & 94.4\% & 88.9\% \\
GPT-5 mini & 63.9\% & 66.7\% & 66.7\% & 80.6\% & 69.4\% \\
GPT-5 nano & 66.7\% & 61.1\% & 58.3\% & 58.3\% & 66.7\% \\
\bottomrule
\end{tabular}
\caption{Per-model accuracy by passage length for JSON (Query Bottom).}
\label{tab:json-target-after-by-length}
\end{table}

\subsection{Accuracy by Target Location}
Unlike Liu et al.\ \citep{liu2024lost}, who find that models perform worst when relevant information is placed in the middle of the context, our results show the steepest degradation in the last third. Accuracy drops most sharply under target--last placement, indicating that end-position targets are more challenging than middle-position targets in our setting.

This effect is driven disproportionately by weaker models. Gemini Flash-Lite collapses almost entirely under JSON conditions (e.g., $0\%$ in the last third for query--first and query--last), while Gemini Flash also shows steep degradation in the last third under query--last ($48.3\%$). By contrast, larger models such as GPT-5 and Gemini Pro maintain high accuracy across thirds, though even here the last third is consistently the weakest segment (e.g., GPT-5 mini falls to $63.3\%$ in query--first and $48.3\%$ in query--last). In short, the overall last-third penalty reflects an interaction between placement and model scale: small models, especially Gemini Flash-Lite, drive the collapse, while larger models only bend but do not break.

\begin{table}[H]
\centering
\small
\renewcommand{\arraystretch}{1.15}
\begin{tabular}{lccc}
\hline
\multirow{2}{*}{\textbf{Third}} 
  & \multicolumn{3}{c}{\textbf{Accuracy (\%)}} \\ 
\cline{2-4}
  & \textbf{Text First Top} & \textbf{JSON Query Top} & \textbf{JSON Query Bottom} \\
\hline
First Third   & 88.1\% & 86.7\% & 79.7\% \\
Middle Third  & 86.1\% & 75.6\% & 59.2\% \\
Last Third    & 79.2\% & 67.2\% & 53.1\% \\
\hline
\end{tabular}
\caption{Accuracy by transcript third, comparing Text First Top placement, JSON Query Top, and JSON Query Bottom.}
\label{tab:thirds-accuracy}
\end{table}

Looking at the per-model breakdown, the substantial gap between first- and last-third accuracy is driven largely by Gemini 2.5 Flash Lite, which shows deltas of $33$ percentage points (pp), $57$pp, and $17$pp on the Text First Top, JSON (Query Bottom), and JSON (Query Bottom) tests, respectively. GPT-5 nano also contributes, with corresponding deltas of $12$pp, $19$pp, and $35$pp. In a few cases models performed slightly worse on the first third, but these drops were generally small and inconsistent across tests.

\begin{table}[H]
\centering
\renewcommand{\arraystretch}{1.15}
\begin{tabular}{lccc}
\toprule
\multirow{2}{*}{\textbf{Model}} 
  & \multicolumn{3}{c}{\textbf{Transcript Third}} \\ 
\cmidrule(lr){2-4}
  & \textbf{First} & \textbf{Middle} & \textbf{Last} \\
\midrule
Gemini-2.5 Flash & 95.0\% & 98.3\% & 98.3\% \\
Gemini-2.5 Flash-Lite & 61.7\% & 43.3\% & 28.3\% \\
Gemini-2.5 Pro & 91.7\% & 96.7\% & 100.0\% \\
GPT-5 & 98.3\% & 98.3\% & 93.3\% \\
GPT-5 mini & 96.7\% & 100.0\% & 81.7\% \\
GPT-5 nano & 85.0\% & 80.0\% & 73.3\% \\
\bottomrule
\end{tabular}
\caption{Per-model accuracy by transcript third for Text First Top.}
\label{tab:opttext-target-first-by-third}
\end{table}

\begin{table}[H]
\centering
\renewcommand{\arraystretch}{1.15}
\begin{tabular}{lccc}
\toprule
\multirow{2}{*}{\textbf{Model}} 
  & \multicolumn{3}{c}{\textbf{Transcript Third}} \\ 
\cmidrule(lr){2-4}
  & \textbf{First} & \textbf{Middle} & \textbf{Last} \\
\midrule
Gemini-2.5 Flash & 93.3\% & 90.0\% & 88.3\% \\
Gemini-2.5 Flash-Lite & 56.7\% & 11.7\% & 0.0\% \\
Gemini-2.5 Pro & 95.0\% & 98.3\% & 96.7\% \\
GPT-5 & 100.0\% & 98.3\% & 93.3\% \\
GPT-5 mini & 95.0\% & 90.0\% & 63.3\% \\
GPT-5 nano & 80.0\% & 65.0\% & 61.7\% \\
\bottomrule
\end{tabular}
\caption{Per-model accuracy by transcript third for JSON (Query Top).}
\label{tab:json-target-before-by-third}
\end{table}

\begin{table}[H]
\centering
\renewcommand{\arraystretch}{1.15}
\begin{tabular}{lccc}
\toprule
\multirow{2}{*}{\textbf{Model}} 
  & \multicolumn{3}{c}{\textbf{Transcript Third}} \\ 
\cmidrule(lr){2-4}
  & \textbf{First} & \textbf{Middle} & \textbf{Last} \\
\midrule
Gemini-2.5 Flash & 80.0\% & 68.3\% & 48.3\% \\
Gemini-2.5 Flash-Lite & 16.7\% & 5.0\% & 0.0\% \\
Gemini-2.5 Pro & 100.0\% & 98.3\% & 61.7\% \\
GPT-5 & 98.3\% & 86.7\% & 93.3\% \\
GPT-5 mini & 96.7\% & 48.3\% & 63.3\% \\
GPT-5 nano & 86.7\% & 48.3\% & 51.7\% \\
\bottomrule
\end{tabular}
\caption{Per-model accuracy by transcript third for JSON (Query Bottom).}
\label{tab:exact-json-target-after-by-third}
\end{table}

\subsection{Off-by-one errors}
Counting off-by-one errors as ``good enough'' matches, a potentially reasonable stance for some workflows (as discussed previously), does not substantially change results when it came to the Text First Top format. Off-by-one rates were generally low across models and providers. Averaged across all models and sentence lengths Gemini averaged an off-by-one error rate of 1.7\% while OpenAI averaged 3.2\%. Though note that OpenAI's models exact-match accuracy was higher, which left less ``headroom'' for errors.

Contrastingly, both JSON formats had higher off-by-one errors, again demonstrating that the Text First Top format achieves greater accuracy by converting off-by-one errors to exact matches. The JSON format, and in particular placing the query at the bottom, induces especially high rates of near misses in Gemini 2.5 Flash-Lite, the mirror effect of its poor exact-match accuracy performance discussed in the previous section. Gemini 2.5 Flash-Lite had off-by-one rates between 20\% and 35\% in the JSON (Query Top) format. Notably, counting off-by-one as success pushes GPT-5 to essentially 100\% performance for all sentence lengths. 

\begin{table}[H]
\centering
\small
\renewcommand{\arraystretch}{1.15}
\begin{tabular}{lccc}
\midrule
\multirow{2}{*}{\textbf{Target Length}} 
  & \multicolumn{1}{c}{\textbf{Text First Top}} 
  & \multicolumn{2}{c}{\textbf{JSON}} \\ 
\cline{2-4}
  & \textbf{Query Top} & \textbf{Query Top} & \textbf{Query Bottom} \\
\midrule
1 sentence   & 2.35\% & 4.6\%  & 13.4\% \\
2 sentences  & 2.35\% & 12.5\% & 13.4\% \\
3 sentences  & 1.40\% & 7.4\%  & 15.7\% \\
5 sentences  & 0.95\% & 4.6\%  & 12.0\% \\
10 sentences & 5.10\% & 4.6\%  & 13.4\% \\
\midrule
\end{tabular}
\caption{Off-by-one rate (\%) by span length and format. Text First Top aggregates across Gemini~2.5 and GPT-5 families (equal weight). JSON rows use the all-model aggregates.}
\label{tab:offby1-by-format}
\end{table}

\subsubsection{Converting Misses to Near Miss, and Near Misses to Exact Matches}
The most interesting pattern to emerge is that off-by-one errors themselves behave differently depending on where the query is placed. When the query is positioned before the transcript, off-by-one rates remain low and largely incidental: small spikes appear at certain lengths, but they never become a distinct stage of model behavior.

By contrast, when the query is placed after the transcript in the JSON format, a striking progression appears. Smaller models often fail outright, middle-tier models convert many of those outright misses into near misses (off-by-one predictions), and the largest models in turn convert those near misses into exact matches. This ladder-like transition is visible across both providers, but with different character: OpenAI’s models exhibit a step-function correction, with off-by-one acting as a temporary holding stage, while Gemini’s models display a more graded reallocation, beginning with extremely high off-by-one rates in Flash-Lite and slowly shifting that mass into exact matches as the models scale up. In both cases, the phenomenon is only consistently present when the query is at the bottom of the transcript—a placement that forces the model to traverse the entire context before producing its answer, and thus highlights the ways in which retrieval difficulty interacts with model scale. This phenomenon can be seen in \autoref{fig:offbyone-transition}.

\begin{figure}[H]
    \centering
    \includegraphics[width=0.8\textwidth]{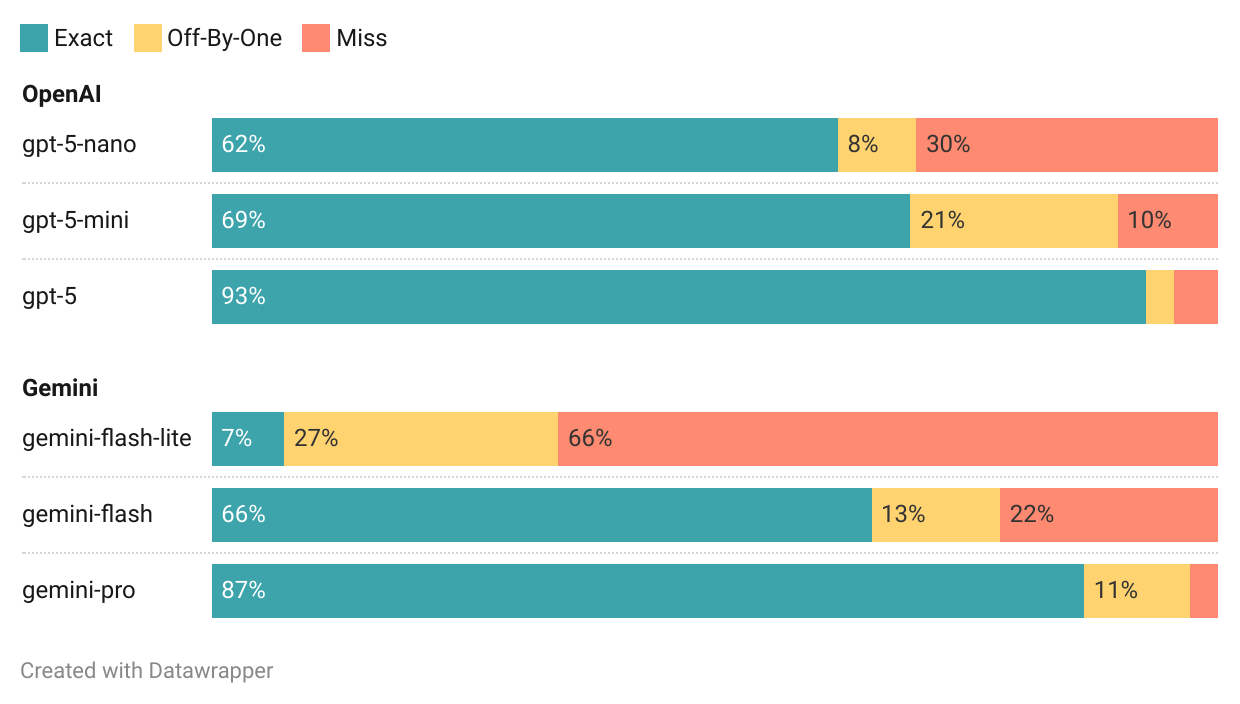}
    \caption{Distribution of exact matches, off-by-one errors, and misses across six models. 
    OpenAI models (Nano, Mini, GPT-5) show a stepwise progression from misses to off-by-one to exact matches, 
    while Gemini models (Flash-Lite, Flash, Pro) exhibit a gradual reallocation of errors into exact matches.}
    \label{fig:offbyone-transition}
\end{figure}

\subsection{Model Response Time, Reasoning Token Usage, and Cost}

\subsubsection{Model Response Time}
In addition to having the lowest accuracy, the JSON (Query Bottom) format also exhibits the highest latency, averaging $21.3$ seconds per response. This compares to $14.4$ seconds for JSON (Query Top) and $13.7$ seconds for Text First Top.
\autoref{tab:exact-match-per-model-latency} ranks models by latency. Gemini 2.5 Flash is the fastest model by a wide margin, returning outputs in about five to six seconds—roughly twice as fast as the next closest model, Gemini 2.5 Pro. In general, Gemini models are faster than their OpenAI counterparts. GPT-5 and GPT-5-mini, in particular, are consistently among the slowest models across all formats.
There was no consistent relationship between consecutive sentence length and latency. Response times varied within each model family without a clear increase with input length.
\begin{table}[H]
\centering
\small
\renewcommand{\arraystretch}{1.15}
\begin{tabular}{lccc}
\midrule
\multirow{2}{*}{\textbf{Model}}
& \multicolumn{1}{c}{\textbf{Text First Top}}
& \multicolumn{2}{c}{\textbf{JSON}} \\
\cline{2-4}
& \textbf{Time (s)} & \textbf{Query Bottom} & \textbf{Query Top} \\
\midrule
Gemini-2.5 Flash & 5.2 & 5.9 & 6.1 \\
Gemini-2.5 Pro & 10.3 & 25.6 & 12.9 \\
GPT-5 nano & 12.1 & 25.6 & 11.9 \\
GPT-5 mini & 14.9 & 29.1 & 17.2 \\
Gemini-2.5 Flash-Lite & 17.0 & 13.2 & 18.5 \\
GPT-5 & 22.6 & 28.0 & 19.8 \\
\midrule
\end{tabular}
\caption{Average response time (seconds, rounded to one decimal) by query format.}
\label{tab:exact-match-per-model-latency}
\end{table}

\subsubsection{Reasoning Token Usage}
We examine models’ reasoning token usage as the length of consecutive sentences increases and as the query is placed either before or after the transcript.

When the query appears before the transcript (Text First Top and JSON Top), median token usage grows roughly linearly with sentence length (\autoref{fig:median-thinking-tokens-by-length}). The slope is about 100 additional tokens per extra sentence, starting from an intercept near 400 tokens. The curve looks slightly convex at longer spans, not because of a true acceleration, but due to the jump in sentence lengths measured (from five directly to ten).

By contrast, when the query appears after the transcript (JSON Bottom), reasoning token medians are consistently higher across all lengths yet remain flat with respect to sentence length. This is consistent with previous research \citep{liu2024lost} showing that placing the query at the top of a context improves performance, and with the decoder-only architecture of both GPT-style and Gemini models, which can only attend to prior tokens at each step. In this setup, the model must process the haystack first and only then integrate the query, leading to a fixed traversal cost. Once this fixed overhead dominates, adding more sentences makes little marginal difference.

It is worth noting that JSON (Query Top) was particularly difficult for GPT-5 nano, which consumed ~2,000 reasoning tokens at all lengths—nearly double the usage of other models and enough to skew global medians upward. Yet the broader pattern still holds: across models, the target-after condition costs between two and five times more tokens than target-before.

\begin{figure}[H]
    \centering
    \includegraphics[width=0.8\textwidth]{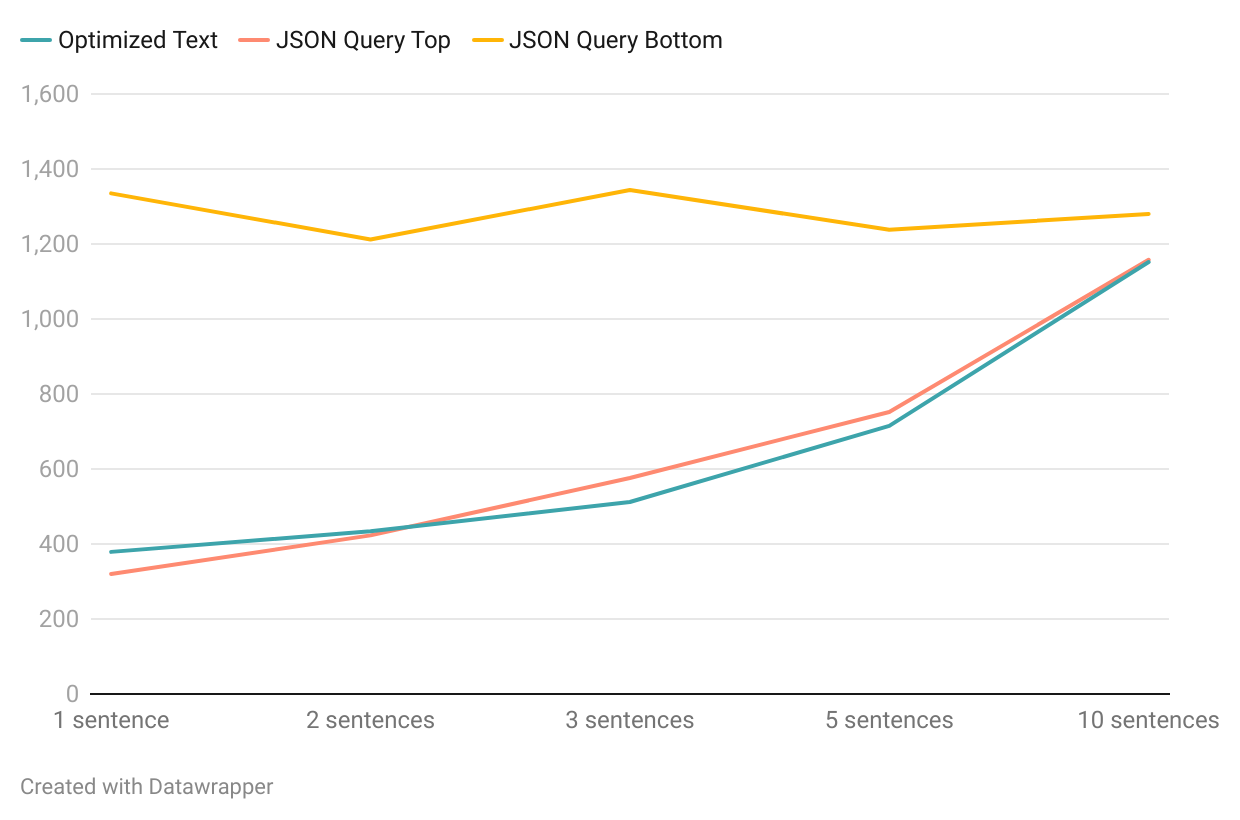}
    \caption{Median number of thinking tokens per test by span length across formats. 
    Text First Top and JSON Query Bottom follow a smooth upward trajectory as sentences lengthen, 
    while JSON Query Top consistently elicits higher median token usage, indicating greater deliberation costs.}
    \label{fig:median-thinking-tokens-by-length}
\end{figure}

\subsubsection{Cost per correct answer}
\autoref{fig:scatter-cost-exact-match} plots total cost against accuracy for each model across the three formats. Cheaper, moderately accurate models cluster in the upper left, while larger, more accurate—and more expensive—models appear in the upper right. Reported costs include prompt auto-caching. Although both providers advertise auto-caching when a portion of the prompt remains fixed across multiple model calls, we find Google Gemini’s implementation more reliable. For example, GPT-5 and Gemini 2.5 Pro have identical input and output token pricing, and Gemini 2.5 Pro produced more output tokens overall. One would therefore expect Gemini 2.5 Pro to be more expensive, yet the opposite holds in practice, owing to Gemini’s more consistent caching.

Caching also helps explain why JSON (Query Top) sometimes costs less than Text First Top (Query Bottom), even though Text First Top is 30\% shorter on input tokens. In both formats the transcript remains entirely constant, but placement matters: when the query is appended at the bottom, the fixed transcript portion sits at the top of the prompt, making it more amenable to caching. By contrast, placing the query at the top forces variation into the very first tokens, which greatly limits the amount of cachable input text. That said, caching is not guaranteed: it does not transfer across models, and whether a given provider caches an input at all can be unpredictable—even when the input is theoretically cacheable.

\begin{figure}[H]
    \centering
    \includegraphics[width=0.8\textwidth]{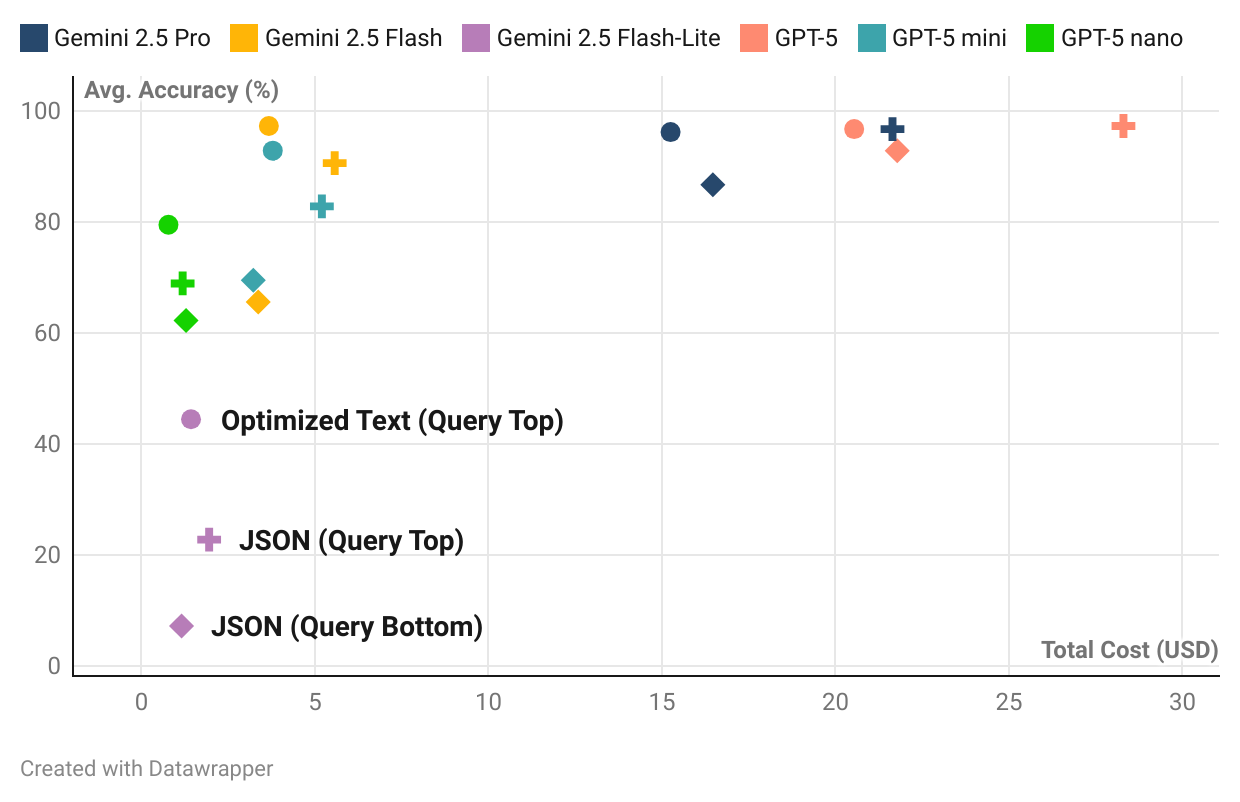}
    \caption{Comparison of model costs and accuracy across JSON (Query Bottom/Bottom) and Text First Top test formats.}
    \label{fig:scatter-cost-exact-match}
\end{figure}

The tradeoff between cost and accuracy naturally leads to another performance metric: cost per correct answer. Unlike accuracy, this metric highlights the efficiency of models rather than the absolute number of correct answers. Here smaller models dominate. This is a natural consequence of model cost structure.  
For example, GPT-5 nano is approximately $250\times$ cheaper than GPT-5 on input tokens, yet its accuracy is not $250\times$ worse.  
Likewise, while GPT-5 nano produces about twice as many reasoning tokens as GPT-5, a $2\times$ increase in output tokens is negligible compared to the $250\times$ reduction in input cost.  

\begin{table}[H]
\centering
\small
\renewcommand{\arraystretch}{1.15}
\begin{tabular}{lccc}
\toprule
\multirow{2}{*}{\textbf{Model}} 
  & \multicolumn{1}{c}{\textbf{Text First Top}} 
  & \multicolumn{2}{c}{\textbf{JSON}} \\ 
\cline{2-4}
  & \textbf{Query Top} & \textbf{Query Top} & \textbf{Query Bottom} \\
\midrule
GPT-5 nano            & \$0.005 & \$0.010 & \$0.011 \\
Gemini 2.5 Flash-Lite & \$0.018 & \$0.048 & \$0.089 \\
Gemini 2.5 Flash      & \$0.021 & \$0.034 & \$0.029 \\
GPT-5 mini            & \$0.023 & \$0.035 & \$0.026 \\
Gemini 2.5 Pro        & \$0.088 & \$0.124 & \$0.106 \\
GPT-5                 & \$0.118 & \$0.162 & \$0.130 \\
\bottomrule
\end{tabular}
\caption{Cost per correct answer (USD) by model and input format. Rows are ranked by average CPC across the three columns (higher is better).}
\label{tab:cpc-by-model}
\end{table}

\subsection{Testing False Positive Rates}
We run a contracted test to see if models properly acknowledge when a requested target sentence is not present in the input transcript. We do this by selecting target sentences from the June 12 2025 House transcript, but using as the input transcript the June 11 2025 House transcript. All input prompts instruct the model to return 0ms-0ms as the start and end millisecond timestamps if the target sentence(s) cannot be identified in the input transcript. The experimental setup used standard lengths {1,2,3,5,10} but only 6 sentences per length with three tries (instead of the normal 12 sentences per length). The Text First Top format was used.

The resulting false positive rates are reported in Table~\ref{tab:null-target-first-by-length}. A false positive here means the model returned a non-zero timestamp range instead of the expected 0ms--0ms. Overall, models were uniformly strong at rejecting absent targets. Notably, the Gemini models showed slightly elevated false positive rates on longer spans. This is somewhat counterintuitive: longer sequences of consecutive sentences should diverge more sharply across two different transcripts, while shorter spans may still overlap or resemble each other.

\begin{table}[H]
\centering
\renewcommand{\arraystretch}{1.15}
\begin{tabular}{lccccc}
\toprule
\multirow{2}{*}{\textbf{Model}} 
  & \multicolumn{5}{c}{\textbf{Consecutive Sentence Length}} \\ 
\cmidrule(lr){2-6}
  & \textbf{1} & \textbf{2} & \textbf{3} & \textbf{5} & \textbf{10} \\
\midrule
Gemini-2.5 Flash & 0.0\% & 0.0\% & 0.0\% & 11.1\% & 16.7\% \\
Gemini-2.5 Flash-Lite & 0.0\% & 0.0\% & 0.0\% & 5.6\% & 22.2\% \\
Gemini-2.5 Pro & 0.0\% & 0.0\% & 0.0\% & 0.0\% & 5.6\% \\
GPT-5 & 0.0\% & 0.0\% & 0.0\% & 0.0\% & 0.0\% \\
GPT-5 mini & 0.0\% & 0.0\% & 0.0\% & 0.0\% & 0.0\% \\
GPT-5 nano & 0.0\% & 0.0\% & 0.0\% & 0.0\% & 0.0\% \\
\bottomrule
\end{tabular}
\caption{Per-model false positive rate by passage length for Text First Top (Query Top).}
\label{tab:null-target-first-by-length}
\end{table}

\subsection{Prior Pilot: JSON (Query Bottom)}
\label{sec:timestamps:prior-pilot}
Before the full evaluation we conducted a smaller pilot restricted to the JSON (Query Bottom) format, with \(L \in \{1,2,3,5,10\}\) consecutive sentences (note the addition of length 4), three attempts per transcript third ($n=9$ per length), and reasoning settings left to defaults (\S\ref{sec:reasoning-setup}). This differs from the main experiment which used $n=36$ per length, a slightly different set of lengths, multiple formats, and reasoning explicitly enabled.

\subsubsection{Accuracy comparison with main run}
Despite the smaller sample and the harder placement condition, the pilot reproduced the central patterns of the full run in Section \ref{sec:exact-match-timestamp-results}. First, query-last is difficult and variance–heavy: weaker and mid-tier models swing from perfect to zero across thirds and lengths, while strong models are steady. Second, length effects are non-monotone: performance generally improves from 1 to 3–5 sentences and then softens by 10, but with the smaller sample size realizations were noisier. Third, last third of the transcript was consistently weakest, with many smaller models collapsing entirely, while stronger systems such as Gemini~2.5 Pro, GPT-4.1, and Claude Sonnet~4.0 maintained high exact-match rates even in this regime.

\subsubsection{Length-by-length Analysis}
At length one top models (e.g., Gemini 2.5-Pro, GPT-4.1, Claude Sonnet 4.0) achieved $100\%$ exact match rates, while lighter variants (e.g., Flash-Lite) showed low exact-match and elevated off-by-one.

At length two Gemini 2.5 Pro remained flawless, Gemin 2.5 Flash delivered mid–50s accuracy at very low cost per correct, and several Claude and GPT mini models dropped to partial success, often concentrated to the first third.

At length three Gemini 2.5 Flash again performed strongly, Pro remained robust, GPT-4.1 reached two-thirds, and smaller models diverged—some producing exact matches only in isolated thirds.

At length four Gemini 2.5 Flash rose to nearly 90\% and GPT-4.1 stayed high, but mid-tier Claude and GPT mini models became brittle, especially in the last third, with some showing complete collapse.

At length five GPT-4.1 achieved perfect performance, Gemini 2.5 Pro and Flash stayed strong, but smaller models were either inconsistent or uniformly poor.

At length ten Gemini 2.5 Pro was again perfect, Flash close behind at nearly 90\%, GPT-4.1 at 44\% but perfect in the first third only, and most others at or near zero, mirroring the long-span fragility we observe in the main study.

\begin{table}[H]
\centering
\renewcommand{\arraystretch}{1.15}
\begin{tabular}{lcccccc}
\toprule
\multirow{2}{*}{\textbf{Model}} 
  & \multicolumn{6}{c}{\textbf{Exact Match Rate (\%) by Span Length}} \\
\cmidrule(lr){2-7}
  & \textbf{1} & \textbf{2} & \textbf{3} & \textbf{4} & \textbf{5} & \textbf{10} \\
\midrule
Gemini-2.5 Pro              & 100 & 100 & 67 & 67 & 89 & 100 \\
Gemini-2.5 Flash            & 44  & 56  & 78 & 89 & 78 & 89 \\
Gemini-2.5 Flash-Lite & 33 & 0   & 0  & 0  & 0  & 0  \\
GPT-4.1                     & 100 & 33  & 67 & 78 & 100 & 44 \\
GPT-4.1 mini                & 67  & 56  & 33 & 33 & 0   & 22 \\
GPT-4.1 nano                 & 33  & 0   & 0  & 0  & 0   & 0  \\
Claude Sonnet 4.0           & 100 & 11  & 44 & 44 & 11  & 0  \\
Claude 3.7 Sonnet           & 67  & 0   & 0  & 0  & 0   & 0  \\
Claude 3.5 Haiku            & 56  & 22  & 0  & 11 & 33  & 0  \\
o3                          & 67  & 33  & 44 & 33 & 33  & 33 \\
o3-mini                     & 0   & 0   & 11 & 22 & 11  & 0  \\
o4-mini                     & 33  & 33  & 33 & 33 & 33  & 33 \\
\bottomrule
\end{tabular}
\caption{Pilot exact-match rates (\%, $n=9$ per length) for JSON (Query Top) with no reasoning tokens. Values rounded from raw counts (e.g., 9/9 = 100\%).}
\label{tab:pilot-json-after}
\end{table}

\subsubsection{Region effects}
Region breakdowns reinforced the main study’s findings. Consistent with Table~\ref{tab:thirds-accuracy}, last-third accuracy was the most brittle. In the pilot, multiple mid and small models posted $0/3$ in the final third at several lengths, while the strongest systems maintained $3/3$ across all thirds. This replicates our scaled finding that placement (Query Bottom) and tail position interact: small models break, larger models bend.

\begin{table}[H]
\centering
\renewcommand{\arraystretch}{1.15}
\begin{tabular}{lccc}
\toprule
\multirow{2}{*}{\textbf{Model}} 
  & \multicolumn{3}{c}{\textbf{Transcript Third (Exact Match \%)}} \\
\cmidrule(lr){2-4}
  & \textbf{First} & \textbf{Middle} & \textbf{Last} \\
\midrule
Gemini-2.5 Pro              & 100 & 100 & 100 \\
Gemini-2.5 Flash            & 100 & 44  & 67  \\
Gemini-2.5 Flash-Lite & 100 & 0   & 0   \\
GPT-4.1                     & 67  & 78  & 44  \\
GPT-4.1 mini                & 67  & 22  & 0   \\
GPT-4.1 nano                 & 33  & 0   & 0   \\
Claude Sonnet 4.0           & 67  & 33  & 22  \\
Claude 3.7 Sonnet           & 33  & 0   & 0   \\
Claude 3.5 Haiku            & 44  & 11  & 33  \\
o3                          & 78  & 11  & 11  \\
o3-mini                     & 11  & 11  & 0   \\
o4-mini                     & 67  & 11  & 11  \\
\bottomrule
\end{tabular}
\caption{Pilot exact-match rates (\%, $n=3$ per region) for JSON (Query Top) with reasoning settings left on default. Values are aggregated across passage lengths.}
\label{tab:pilot-json-after-region}
\end{table}

\subsubsection{Off-by-one errors}
The pilot exhibits the same \emph{miss $\rightarrow$ off-by-one $\rightarrow$ exact} ladder we highlight in Section~\ref{sec:exact-match-timestamp-results}: weaker models failed outright, mid-tier models converted some misses into near misses, and large models in turn converted those into exact matches. This staging behavior is less visible when the query sits on top (main study) because the task is easier and near-misses are rarer.

\subsubsection{Cost per correct}
Cost efficiency in the pilot echoed the main results. Although prices spanned an older lineup and a single placement, the efficiency ranking resembled \autoref{tab:cpc-by-model}: small and efficient models such as Gemini~2.5 Flash often delivered the lowest \$\,/correct at modest accuracy, while larger models bought reliability at a premium. Pro-tier Gemini paired high accuracy with reasonable cost, and older large models like GPT-4.1 were accurate but expensive. Flash and Pro consistently offered strong value, large GPT models traded cost for reliability, and older or lighter models oscillated between occasional low-cost successes and complete failure.

\begin{table}[H]
\centering
\renewcommand{\arraystretch}{1.15}
\begin{tabular}{lcc}
\toprule
\textbf{Model} & \textbf{Cost (USD)} & \textbf{Cost per Correct (USD)} \\
\midrule
Gemini-2.5 Flash            & 0.19 & 0.048 \\
Gemini-2.5 Pro              & 0.81 & 0.090 \\
Claude Sonnet 4.0           & 0.40 & 0.099 \\
GPT-4.1                     & 2.21 & 0.245 \\
GPT-4.1 mini                & 0.44 & 0.074 \\
GPT-4.1 nano                 & 0.11 & 0.037 \\
Claude 3.5 Haiku            & 0.58 & 0.116 \\
Claude 3.7 Sonnet           & 0.65 & 0.653 \\
Gemini-2.5 Flash-Lite & 0.08 & 0.026 \\
o3                          & 2.26 & 0.377 \\
o3-mini                     & 1.16 & 1.131 \\
o4-mini                     & 1.21 & 0.405 \\
\bottomrule
\end{tabular}
\caption{Pilot costs and cost-per-correct (USD) for JSON (Query Top), no reasoning tokens. Values rounded from raw totals across all lengths.}
\label{tab:pilot-json-after-cost}
\end{table}

\subsubsection{Takeaways relative to the full run.}
The pilot—though smaller and noisier—acted as a stress-biased confirmation of the main study, showing that moving the query to the bottom inflates difficulty, exacerbates last-third penalties, and magnifies the progression from miss to near miss to exact match across model scales.

%╔══════════════════════════════════════════════════════════════════════════════╗
%║  ███╗   ██╗███████╗██╗    ██╗    ███████╗███████╗ ██████╗                    ║
%║  ████╗  ██║██╔════╝██║    ██║    ██╔════╝██╔════╝██╔════╝                    ║
%║  ██╔██╗ ██║█████╗  ██║ █╗ ██║    ███████╗█████╗  ██║                         ║
%║  ██║╚██╗██║██╔══╝  ██║███╗██║    ╚════██║██╔══╝  ██║                         ║
%║  ██║ ╚████║███████╗╚███╔███╔╝    ███████║███████╗╚██████╗                    ║
%║  ╚═╝  ╚═══╝╚══════╝ ╚══╝╚══╝     ╚══════╝╚══════╝ ╚═════╝                    ║
%╚══════════════════════════════════════════════════════════════════════════════╝

\section{Full Evaluation: Fuzzy Match Timestamp Retrieval (Task 3)}
\label{sec:fuzzy}
Having established in Section~\ref{sec:timestamps} that models can return millisecond boundaries for \emph{verbatim} passages we now move on to the task of fuzzy matching. In the fuzzy matching task, the target quote may differ from the transcript via ordinary lexical drift (contractions, punctuation, light paraphrase), Speech-To-Text (STT) artifacts (insertions/deletions, segmentation), or editorial edits between a (possibly edited) written record and the associated spoken source. See Section~\ref{sec:why-exact-match-fails} for a detailed discussion of these challenges. The model must still identify the \texttt{start\_ms}–\texttt{end\_ms} of the best matching passage despite these perturbations.

In this section we first bench mark models on a synthetic dataset of perturbed sentences. We show that fuzzy matching is more challenging than exact matches, especially for smaller models. We then introduce a straightforward hybrid matching algorithm that drastically improves performance.

\subsection{Experimental setup and data creation}
Fuzzy sentences were generated synthetically using a structured prompt with six few-shot demonstrations in Gemini 2.5 Pro. The prompt specified a menu of perturbations: punctuation edits, abbreviation expansions/contractions, light word substitutions, capitalization shifts, filler insertions/removals, and minor rephrasings that leave the meaning intact. In short, the goal was to create variants that look like natural Speech-To-Text or editorial drift, not paraphrases. 

The base set of passages was the same as Task~2 (exact matching). Across five sentence lengths (1, 2, 3, 5, 10), we generated twelve perturbed examples per length, for a total of sixty sentence. Like in Task 2 each model was given three attempts per sentence. By design each fuzzy sentence was still unambiguous: recognizable as the same passage but with realistic surface differences. On average the normalized Levenshtein similarity between original and perturbed was 91\%, which reflects mild but non-trivial changes.  

An example of a perturbed sentence and its true fuzzy match is shown in the table below. These two sentences have a normalized Levenshtein similarity of 91\%.

\begin{table}[H]
\centering
\caption{Semantically identical but lexically divergent sentences. A quote–to–timestamp retriever must tolerate all highlighted differences.}
\vspace{0.3em}
\begingroup\setlength{\emergencystretch}{2em}\sloppy
\begin{tabularx}{\textwidth}{@{}P P@{}}
\toprule
\textbf{Original (fuzzy match)} & \textbf{Perturbed } \\
\midrule
``PBS and NPR will continue to pivot their response to this cut back and forth as they have for some time. They tell us that their taxpayer funded gift is just a drop in the bucket, not worth our attention.''  & ``\hl{Public Broadcasting Service} and NPR will continue to pivot their response to this cut\hl{, back and forth, as they have for some time now}. \hl{Now,} they tell us that their \hl{taxpayer-funded} gift is just \hl{`a drop in the bucket'}, not worth our attention.'' \\
\bottomrule
\end{tabularx}
\endgroup
\label{tab:failure-transcript-fuzzy}
\end{table}

\subsection{Results}
Performance fell on fuzzy matching, but the decline was modest—roughly 4–10 percentage points depending on model. This is consistent with expectations: given the confidence intervals already seen in the Gemini 2.5 Flash deep dive, fuzzy accuracy lines up with the lower bound of the exact-match distribution. Flash Lite actually improved slightly (+2 points), though this bump is small and likely noise. The takeaway is that modern models tolerate light perturbation fairly well; degradation exists but is nowhere near catastrophic.  

Overall OpenAI models had higher accuracy on the Fuzzy Matching task, 83.7\% compared to 75.2\%. The relative ordering of models remained unchanged, with GPT-5 and Gemini-2.5 Pro still clustered at the top, though both dropped a few points. Gemini-2.5 Flash continues to show strong performance, but its margin over GPT-5 mini disappeared under fuzziness. GPT-5 nano saw the steepest decline, reflecting greater brittleness in smaller-capacity models.  These results suggest that scale and training depth both play a role in robustness to noisy inputs.

\begin{table}[H]
\centering
\renewcommand{\arraystretch}{1.15}
\begin{tabular}{lcc}
\toprule
\textbf{Model} & \textbf{Exact Match} & \textbf{Fuzzy Match} \\
\midrule
Gemini-2.5 Pro                & 96.1\% & 91.7\% \\
Gemini-2.5 Flash              & 97.2\% & 87.8\% \\
Gemini-2.5 Flash-Lite & 44.4\% & 46.1\% \\
GPT-5                         & 96.7\% & 92.8\% \\
GPT-5 mini                    & 92.8\% & 87.8\% \\
GPT-5 nano                    & 79.4\% & 70.6\% \\
\bottomrule
\end{tabular}
\caption{Per-model accuracy: exact match (averaged across lengths) vs.\ unAssisted Fuzzy match.}
\label{tab:opttext-exact-vs-fuzzy}
\end{table}

\subsubsection{Pilot results}
The pilot experiment with older models shows a sharper drop. The setup was different from the main fuzzy task: Text First Top, enablement of reasoning left at defaults, only six handcrafted fuzzy examples (again three tries each, for $n=18$). This lighter, noisier design is less generalization, but we present the results here for completeness.

In the pilot some models lost 20–30 points when moving from exact to fuzzy. Larger models (Gemini Pro, GPT-4.1 full) generally held up better, staying above ~40–80\%, while their mini/nano counterparts had accuracy as low as single digits. Provider effects were clear as well: Gemini models degraded more gracefully, usually giving up 10–15 points, while the Claude line was brittle—Haiku and Sonnet often fell to zero under fuzziness. OpenAI’s mid-tier (GPT-4.1 mini) dropped 20 points, but GPT-4.1 full was at least competitive with Pro and Flash. In short, size and generation mattered: big models carried resilience, while small ones showed fragility once edits entered the mix.

\begin{table}[H]
\centering
\renewcommand{\arraystretch}{1.15}
\begin{tabular}{lcc}
\toprule
\textbf{Model} & \textbf{Exact Match (Pilot)} & \textbf{Fuzzy Match (Pilot)} \\
\midrule
Gemini-2.5 Pro                      & 87.0\% & 77.8\% \\
Gemini-2.5 Flash                    & 72.2\% & 61.1\% \\
GPT-4.1                              & 70.4\% & 38.9\% \\
o3                                   & 40.7\% & 16.7\% \\
GPT-4.1 mini                         & 35.2\% & 11.1\% \\
Claude Sonnet 4.0                    & 35.2\% & 16.7\% \\
o4-mini                              & 33.3\% & 5.6\% \\
Claude 3.5 Haiku                     & 20.4\% & 0.0\% \\
Claude 3.7 Sonnet                    & 11.1\% & 0.0\% \\
Gemini-2.5 Flash-Lite        & 5.6\%  & 16.7\% \\
GPT-4.1 nano                          & 5.6\%  & 0.0\% \\
o3-mini                              & 7.4\%  & 0.0\% \\
\bottomrule
\end{tabular}
\caption{Pilot comparison by model: exact-match accuracy (aggregated across lengths $L \in\{1,2,3,4,5,10\}$; $n=54$ per model = 6 lengths $\times$ 3 sentences $\times$ 3 trials) vs.\ fuzzy accuracy ($n=18$ per model = 6 sentences $\times$ 3 trials). Note the pilot used Text First Top format with reasoning enablement left at defaults.}
\label{tab:pilot-exact-vs-fuzzy}
\end{table}

\subsubsection{Reasoning Token Usage}
Median reasoning token usage rose under the fuzzy match task. Across models the typical increase was about 160 tokens per response, or roughly a 25\% bump. The scale of the jump varied: Gemini~2.5 Flash rose from a median of 566 to 802 tokens (a 42\% increase), while GPT-5 mini barely moved, from 448 to 512 (a 14\% rise). The Gemini 2.5 Flash-Lite showed the steepest change, nearly doubling from 450 to 846 (88\%).
It is notable that even small perturbations in sentence wording force models to deliberate more, in some cases 50\% - 90\% more.

\begin{table}[H]
\centering
\renewcommand{\arraystretch}{1.15}
\begin{tabular}{lcc}
\toprule
\textbf{Model} & \textbf{Text First Top (median)} & \textbf{Fuzzy (median)} \\
\midrule
Gemini-2.5 Flash              & 566  & 802 \\
Gemini-2.5 Flash-Lite         & 450  & 846 \\
Gemini-2.5 Pro                & 578  & 957 \\
GPT-5                         & 640  & 800 \\
GPT-5 mini                    & 448  & 512 \\
GPT-5 nano                    & 704  & 832 \\
\bottomrule
\end{tabular}
\caption{Median thinking tokens per response under Text First Top vs.\ unAssisted Fuzzy retrieval. Medians across n=180 trials per model.}
\label{tab:thinking-tokens-exact-vs-fuzzy-median}
\end{table}

\subsubsection{Latency}
Across all models, the introduction of fuzzy retrieval generally increased latency by about three seconds relative to Text First Top. An exception was GPT-5-nano, where latency increased by less than 2\% despite an increase in reasoning tokens of almost 20\%. Gemini-2.5 Flash remained the clear leader in speed, sustaining latencies roughly half those of the next fastest models. At the other end, GPT-5 continued to lag with the slowest responses overall. When averaged across variants, Gemini models achieved a mean response time of 13.7 seconds, compared to 18.5 seconds for GPT-5 models, underscoring a consistent advantage in latency.  Beyond raw speed, the results suggest that Gemini models scale more predictably under added complexity, while GPT-5 variants show higher variance and slower baseline performance. However, note the tradeoff in latency and accuracy. As previously discussed, OpenAI models are 8.5 percentage points more accurate on average than Gemini models.

\begin{table}[H]
\centering
\renewcommand{\arraystretch}{1.2}
\begin{tabular}{lcc}
\toprule
\multirow{2}{*}{\textbf{Model}} 
  & \textbf{Text First Top} & \textbf{Fuzzy} \\
\cmidrule(lr){2-3}
  & Time (s) & Time (s) \\
\midrule
Gemini-2.5 Flash & 5.2 & 7.3 \\
Gemini-2.5 Pro & 10.3 & 13.8 \\
GPT-5 nano & 12.1 & 12.3 \\
GPT-5 mini & 14.9 & 17.7 \\
Gemini-2.5 Flash-Lite & 17.0 & 20.1 \\
GPT-5 & 22.6 & 25.5 \\
\bottomrule
\end{tabular}
\caption{Average response time (seconds, rounded to one decimal) for Text First Top and fuzzy retrieval tasks.}
\label{tab:opt-vs-fuzzy}
\end{table}

%-----------------------------------------------------------------------------
% Hybrid Fuzzy Matching
%-----------------------------------------------------------------------------
\subsection{Hybrid Fuzzy Matching}
\label{sec:fuzzy-matching}
We now move on to an improved fuzzy matching technique that vastly improves performance while reducing cost and latency. The approach has two stages:
\begin{enumerate}
\item Using off-the-shelf machine learning algorithms to reduce the full transcript to a set of probable snippets.
\item Send this reduced list of snippets to an LLM for final verification.
\end{enumerate}

We describe the details of the approach below and then show that the method allows for extremely high accuracy, surpassing even that found in the Exact Match task.

\subsubsection{Hybrid Algorithm}
We propose a hybrid timestamp retrieval method (Algorithm~\ref{alg:hybrid-matching-algo}), combining fuzzy matching and dynamic snippet sizing with downstream refinement by a Large Language Model (LLM). The hybrid timestamp retrieval algorithm balances efficient fuzzy matching with the semantic precision of LLMs, which can resolve meaning when multiple near-matches exist.

Initially, the target quote undergoes basic text cleaning to remove trivial differences (e.g., honorifics like ``Mr. Speaker''). Fuzzy matches are then identified using the RapidFuzz Python Library. Dynamic offsets are then applied, creating expanded snippets from the original matches to ensure complete coverage while offering a buffer for safety. Overlapping snippets are merged to reduce redundancy and increase token efficiency. The merged snippets—now compact—are then passed to a downstream LLM, which returns exact millisecond-precision timestamps. There are numerous nuances to the algorithm which are shown in Algorithm~\ref{alg:hybrid-matching-algo} and described in more accessible language below. There are five main steps.

\begin{enumerate}
  \item Build sentence-only and joined views. We first construct a sentence-only Python list by stripping timestamps and metadata from the transcript so that each index in this list corresponds one-to-one with the original position of that sentence in the Text First Top transcript format. We also create a single joined string by concatenating all sentence texts without separators (no timestamps are included in this either). For each sentence in the sentence-only list, we then record its starting and ending character offsets within the joined string. We therefore end up with four objects:
    \begin{enumerate}
    \item The original Text First Top transcript complete with both spoken sentences and millisecond-level timestamps.
      \item A sentence-only list: a Python list of strings, one per transcript sentence, aligned 1:1 with the Text First Top transcript, but without timestamp markers.
      \item Character spans: a Python list of (start\_char, end\_char) offsets corresponding to the sentence-only list.
      \item Joined string: a single string formed by concatenating all entries in S without separators.
    \end{enumerate}
    The index of objects a to c above all correspond, making it easy to, for example, look up character spans given an index from the sentence-only list, or convert from the charcter spans or sentence-only list back to the original Text First Top transcirpt.

  \item Clean the target quote. We normalize the input quote with conservative text cleaning that removes superficial variations (for example, honorifics and punctuation mismatches) while preserving the semantic content that matters for matching. This reduces false positives without injecting bias or losing important distinctions.

  \item We produce two kinds of match candidates and combine them.
  \begin{enumerate}
  \item First, we run sentence-level matching over the sentence-only list and keep the top three sentence indices with scores above a threshold (70\% fuzzy ratio); this path favors longer quotes that resemble complete sentences. However, because it looks at the transcript sentence-by-sentence it does poorly on shorter sentences that may be similar to others in the transcript. This is done using the ratio function in the RapidFuzz Python library.
  \item Second, we run a match over the joined string to capture short or fragmentary quotes. This is done using the partial alignment function in the RapidFuzz Python library. Because the partial alignment function outputs starting and ending character spans from the joined string, we then map these back to the coorsponding sentence using the character span list described in Step 1.
  \end{enumerate}
  This dual approach was discovered empirically after in-depth, freeform testing and noticing that using either method alone produced failure points.
  
\item We then prepare the model input for Step 5. We compute a sentence-based radius from the estimated quote length with a small safety buffer so the model sees full context even if the best RapidFuzz match lands very near the start or end of a span of sentences. We always include at least three sentences of context and cap extra padding at eight sentences; longer quotes get proportionally more context so the window scales with the length of the target passage. We map each candidate snippet from Step 3 back to the original Text First Top transcript to regain the sentence timestamps that were removed in Step 1 for matching. We then union these candidate sets, deduplicate, and continue only if at least one candidate snippet remains. This may result in a single snippet if all individually matched snippets happen to overlap, or conversely could result in up to four separate snippets, three from Step 3a and one from Step 3b. Thus in Step 5 the model receives up to four snippets from the Text First Top transcript, each typically ranging from 3 to 18 sentences in length.

  \item The model receives the transcript snippets from Step 4 with a prompting indicating it is meant to conduct a fuzzy match and return the start and end millisecond timestamps. This final step aims to transform probabilistic timestamp markers that might have been found using a purely fuzzy matching approach into more deterministic timestamp retrieval using an LLM, reducing off-by-one errors while keeping latency low because the model only sees the relevant, merged context rather than the entire transcript.
\end{enumerate}

\begin{algorithm}[H]
\DontPrintSemicolon
\SetKwInput{KwIn}{Input}\SetKwInput{KwOut}{Output}

\KwIn{$q$ (target), transcript $T=\big[\{start_{ms},end_{ms},text\}_1,\dots,\{.\}_n\big]$, model $\mathcal{M}$}
\KwOut{$(\widehat{start}_{ms}, \widehat{end}_{ms})$}

$\tilde{q} \gets \texttt{Clean}(q)$ \tcp*{Clean target}
$S \gets [\,T[i].\text{text}\,]_{i=1}^n$ \tcp*{Sentence list}
$\texttt{joined} \gets \texttt{concat}(S)$ \tcp*{Joined text}
$\texttt{charSpans} \gets \texttt{CharacterSpansOf}(S)\ \text{in}\ \texttt{joined}$ \tcp*{Char spans}

$\texttt{matches}_{\text{ratio}} \gets \texttt{TopKRatio}(\tilde{q}, S)$ \tcp*{Sentence matches}
$\texttt{match}_{\text{aln}} \gets \texttt{AlignToRange}(\tilde{q}, \texttt{joined}, \texttt{charSpans})$ \tcp*{Alignment match}
$\texttt{matches} \gets \texttt{matches}_{\text{ratio}} \cup \big(\texttt{match}_{\text{aln}} \text{ if not } \varnothing\big)$ \tcp*{Combine candidates}
\If{$|\texttt{matches}| = 0$}{\textbf{return} failure} \tcp*{Fail fast}

$\hat{m} \gets \texttt{EstimateSentenceCount}(\tilde{q})$ \tcp*{Estimate length}
$\Delta \gets \hat{m} + \min\!\big(8,\ \max(3,\ \lfloor 0.25 \cdot \hat{m} \rfloor)\big)$ \tcp*{Dynamic radius}

$\texttt{snippets} \gets \texttt{ExpandAroundMatches}(T, \texttt{matches}, \Delta)$ \tcp*{Expand windows}

Sort $\texttt{snippets}$ by $start\_time$ ascending \tcp*{Order windows}
$\texttt{merged} \gets [\,]$;\quad $\texttt{cur} \gets \texttt{snippets}[1]$ \tcp*{Init merge}
\For{$s$ in $\texttt{snippets}[2..]$}{
  \uIf{$s.start\_time \le \texttt{cur}.end\_time$}{
    $\texttt{cur.start\_index} \gets \min(\texttt{cur.start\_index}, s.start\_index)$ \tcp*{Extend left}
    $\texttt{cur.end\_index} \gets \max(\texttt{cur.end\_index}, s.end\_index)$ \tcp*{Extend right}
    $\texttt{cur.snippet} \gets T[\texttt{cur.start\_index} : \texttt{cur.end\_index}]$ \tcp*{Rebuild text}
    $\texttt{cur.start\_time} \gets \texttt{cur.snippet}[1].start_{ms}$;\quad $\texttt{cur.end\_time} \gets \texttt{cur.snippet}[-1].end_{ms}$ \tcp*{Update times}
  }
  \Else{
    append $\texttt{cur}$ to $\texttt{merged}$;\quad $\texttt{cur} \gets s$ \tcp*{Emit and reset}
  }
}
append $\texttt{cur}$ to $\texttt{merged}$ \tcp*{Flush last}

$\texttt{snippetsText} \gets [\,\texttt{CreateTextFirstFormat}(s.\text{snippet})\ :\ s \in \texttt{merged}\,]$ \tcp*{Text-first format}
\textbf{return} $\mathcal{M}(\texttt{snippetsText}, q)$ \tcp*{LLM refinement}

\caption{Hybrid: sentence-level ratio + all-text alignment → dynamic expansion → merging → LLM refinement}
\label{alg:hybrid-matching-algo}
\end{algorithm}

\subsubsection{Proving Out The Approach}
To isolate the effect of context size from input fuzziness, we repeated the exact-match timestamp retrieval task under a lighter setup using the dynamic snippet method. Instead of constructing twelve passages per length, we reduced to three in order to quickly assess whether the assisted approach of sending smaller transcript windows would prove viable. The experiment therefore has a narrower scope than the main exact-match task, but still provides a useful directional signal.

Results show that accuracy rises to near ceiling performance for nearly all models and sentence lengths. Even the weakest baseline—Gemini 2.5 Flash-Lite, which had achieved just 22\% accuracy at length ten under the Exact Match task (Table~\ref{tab:exact-match-opttext-target-first-by-length-repeat}})—improved to 100\% accuracy under the snippet condition (Table~\ref{tab:assisted-control-target-first-by-length}). Larger models, already strong in the full-context condition, likewise reached or approached ceiling. GPT-5 nano remained the most variable, but even here performance at shorter spans corrected to 100\%, only declining modestly at ten sentences.

Taken together these findings suggest that the primary challenge in exact-match retrieval lies not in the matching task itself but in the interaction with prompt length. Reducing the context to a small, targeted snippet eliminates almost all errors, demonstrating that input token length exerts a first-order effect on performance. In theory this establishes a clear path forward: if fuzzy matching can be paired with snippet construction, the same improvements observed here should extend to the harder Fuzzy Match task as well.

\begin{table}[H]
\centering
\renewcommand{\arraystretch}{1.15}
\begin{tabular}{lccccc}
\toprule
\multirow{2}{*}{\textbf{Model}} 
  & \multicolumn{5}{c}{\textbf{Consecutive Sentence Length}} \\ 
\cmidrule(lr){2-6}
  & \textbf{1} & \textbf{2} & \textbf{3} & \textbf{5} & \textbf{10} \\
\midrule
Gemini-2.5 Flash & 91.7\% & 97.2\% & 97.2\% & 100.0\% & 100.0\% \\
Gemini-2.5 Flash-Lite & 38.9\% & 69.4\% & 52.8\% & 38.9\% & 22.2\% \\
Gemini-2.5 Pro & 80.6\% & 100.0\% & 100.0\% & 100.0\% & 100.0\% \\
GPT-5 & 88.9\% & 100.0\% & 100.0\% & 100.0\% & 94.4\% \\
GPT-5 mini & 80.6\% & 94.4\% & 100.0\% & 100.0\% & 88.9\% \\
GPT-5 nano & 75.0\% & 97.2\% & 94.4\% & 77.8\% & 52.8\% \\
\bottomrule
\end{tabular}
\caption{Reprint from Section~\ref{sec:exact-match-timestamp-results}. Per-model accuracy by span length for the full Exact Match setup using Text First Top (Query Top). 
Each entry averages across 12 sampled spans per length, drawn from different transcript thirds and repeated across three trials (540 total per model).}
\label{tab:exact-match-opttext-target-first-by-length-repeat}
\end{table}

\begin{table}[H]
\centering
\renewcommand{\arraystretch}{1.15}
\begin{tabular}{lccccc}
\toprule
\multirow{2}{*}{\textbf{Model}} 
  & \multicolumn{5}{c}{\textbf{Consecutive Sentence Length}} \\ 
\cmidrule(lr){2-6}
  & \textbf{1} & \textbf{2} & \textbf{3} & \textbf{5} & \textbf{10} \\
\midrule
Gemini-2.5 Flash & 100.0\% & 100.0\% & 100.0\% & 100.0\% & 100.0\% \\
Gemini-2.5 Flash-Lite & 100.0\% & 100.0\% & 100.0\% & 100.0\% & 100.0\% \\
Gemini-2.5 Pro & 100.0\% & 100.0\% & 100.0\% & 100.0\% & 100.0\% \\
GPT-5 & 100.0\% & 100.0\% & 100.0\% & 100.0\% & 100.0\% \\
GPT-5 mini & 100.0\% & 100.0\% & 100.0\% & 100.0\% & 100.0\% \\
GPT-5 nano & 100.0\% & 100.0\% & 100.0\% & 88.9\% & 77.8\% \\
\bottomrule
\end{tabular}
\caption{Per-model accuracy by span length for the Assisted Control (Snippet, Target First). 
This lighter setup samples only 3 spans per length (n=3) rather than 12, sufficient to test the effect of snippet-based prompting.}
\label{tab:assisted-control-target-first-by-length}
\end{table}

\subsection{Fuzzy Matching Results}
The Assisted Fuzzy results provide a strong complement to the exact-match findings in Section 8.2. Whereas fuzzy matching introduces additional ambiguity by requiring the model to tolerate lexical divergence rather than rely on verbatim spans, the assisted condition shows that this difficulty is not insurmountable. Accuracy improves sharply when snippet windows are substituted for full transcripts, with even the weakest models closing much of the gap. The pattern echoes what we observed for assisted exact match (Table~\ref{tab:assisted-control-target-first-by-length}): reducing context length effectively removes retrieval barriers that otherwise appear when models are forced to traverse long transcripts.

The improvement is especially striking for Gemini 2.5 Flash-Lite, which under baseline fuzzy matching hovered around 40\% accuracy, but reached 90\% under the snippet condition (Table~\ref{tab:fuzzy-vs-assisted-fuzzy}). This mirrors the off-by-one ladder effect documented in Section 8.4, where small models often stalled at near misses while larger models converted them into exact matches. Here, however, the shift is more categorical: trimming the haystack allows Flash-Lite to behave more like its larger counterparts. GPT-5 and Gemini Pro, already near ceiling in the unassisted condition, move to essentially perfect performance once context is shortened, underscoring that residual errors at full length were driven by input format rather than by true task ambiguity.

Taken together, these results point to input token length as the decisive factor shaping both exact-match and fuzzy retrieval. The last-third degradation documented in Section 8.3, the off-by-one transitions in Section 8.4, and now the dramatic gains from snippet prompting all suggest that scaling context directly affects whether models miss, nearly miss, or succeed. In other words, the retrieval difficulty is not rooted in aligning target sentences to timestamps, but in maintaining coherence across extended transcripts. The Assisted Fuzzy trial demonstrates that when this burden is removed, even weaker models can achieve accuracies comparable to state-of-the-art systems.

In addition to improved accuracy, the average model response time was cut in half (16 seconds vs. 8 seconds), input token size was reduced by around 99\% (the full transcript was 2,272 sentences compared to just 20 sentences for the average snippet under the Assisted Fuzzy Matching approach), and cost per correct answer dropped by an order of magnitude, from \$0.0547 to \$0.0045. See detailed breakdown in sections \ref{sec:assisted-fuzzy-model-response-times} and \ref{sec:cpc-assisted-fuzzy}.

\begin{table}[H]
\centering
\renewcommand{\arraystretch}{1.15}
\begin{tabular}{lccc}
\toprule
\multirow{2}{*}{\textbf{Model}} 
  & \multicolumn{2}{c}{\textbf{Accuracy (\%)}} & \multirow{2}{*}{\textbf{$\Delta$ (pp)}} \\ 
\cmidrule(lr){2-3}
  & \textbf{Fuzzy Match} & \textbf{Assisted Fuzzy Match} &  \\
\midrule
Gemini-2.5 Flash        & 82.2\% & 97.8\% & +15.6 \\
Gemini-2.5 Flash-Lite   & 41.1\% & 90.0\% & +48.9 \\
Gemini-2.5 Pro          & 91.1\% & 100.0\% & +8.9 \\
GPT-5                   & 95.6\% & 100.0\% & +4.4 \\
GPT-5 mini              & 84.4\% & 97.8\% & +13.4 \\
GPT-5 nano              & 72.2\% & 93.3\% & +21.1 \\
\bottomrule
\end{tabular}
\caption{Per-model accuracy on fuzzy vs.\ Assisted Fuzzy retrieval, with absolute percentage-point ($\Delta$) improvement shown in the rightmost column. Both conditions use the same 60 sentence spans as the exact-match task, with three independent trials each (180 queries per model per condition).}
\label{tab:fuzzy-vs-assisted-fuzzy}
\end{table}

\subsubsection{Pilot Fuzzy vs.\ Assisted Fuzzy Retrieval}
Assisted Fuzzy Matching was also part of the early Pilot experiments. However, the Pilot used just six manually generated fuzzy sentences.

As Table~\ref{tab:pilot-fuzzy-assisted} shows, accuracy increased sharply under the Assisted Fuzzy approach. Models that previously failed outright, such as Claude 3.5 Haiku and GPT-4.1 nano , recovered to double-digit or even mid-range accuracies. Mid-tier systems like Gemini Flash-Lite Preview and Claude Sonnet 4.0 nearly tripled their accuracy, while higher-end models such as GPT-4.1 and Gemini Pro shifted from the 60–80\% range to above 80\%. The strongest improvements were observed in models with low baseline fuzzy accuracy: o3-mini rose from 0\% to 88.9\%, a gain of nearly 90 percentage points (pp), while Gemini Flash-Lite improved by almost 28 pp.

These findings are consistent with the broader pattern documented in Section 8.4: smaller or weaker models that struggle with long-context traversal benefit disproportionately from shortened inputs, converting many outright misses into exact matches. Even though the pilot involved fewer queries, the results reinforce the conclusion from Section 8.2 that input length, rather than the retrieval task itself, is the primary driver of model variability. Instructively, the same ``ladder effect'' seen for off-by-one errors appears here as a categorical shift when snippets are introduced: weak models leap into performance ranges that, under full transcripts, seemed unattainable.

\begin{table}[H]
\centering
\renewcommand{\arraystretch}{1.15}
\begin{tabular}{lccc}
\toprule
\multirow{2}{*}{\textbf{Model}} 
  & \multicolumn{2}{c}{\textbf{Accuracy (\%)}} & \multirow{2}{*}{\textbf{$\Delta$ (pp)}} \\ 
\cmidrule(lr){2-3}
  & \textbf{Fuzzy} & \textbf{Assisted Fuzzy} &  \\
\midrule
Gemini-2.5 Pro                  & 77.8\% & 83.3\% & +5.5 \\
Gemini-2.5 Flash                & 61.1\% & 88.9\% & +27.8 \\
GPT-4.1                         & 38.9\% & 61.1\% & +22.2 \\
Claude Sonnet 4.0               & 16.7\% & 66.7\% & +50.0 \\
Gemini-2.5 Flash-Lite (Preview) & 16.7\% & 44.4\% & +27.7 \\
o3                              & 16.7\% & 88.9\% & +72.2 \\
GPT-4.1 mini                    & 11.1\% & 55.6\% & +44.5 \\
o4-mini                         & 5.6\%  & 83.3\% & +77.7 \\
Claude 3.5 Haiku                & 0.0\%  & 44.4\% & +44.4 \\
Claude 3.7 Sonnet               & 0.0\%  & 61.1\% & +61.1 \\
GPT-4.1 nano                    & 0.0\%  & 16.7\% & +16.7 \\
o3-mini                         & 0.0\%  & 88.9\% & +88.9 \\
\bottomrule
\end{tabular}
\caption{Pilot comparison of Fuzzy Matching vs.\ Assisted Fuzzy Matching across models. Both conditions used the same 60 sentence spans as the exact-match task, with three independent trials each (180 queries per model per condition). While the setup is limited in scope, the consistent improvements across diverse model families indicate that snippet prompting substantially mitigates retrieval difficulty.}
\label{tab:pilot-fuzzy-assisted}
\end{table}

\subsubsection{Model Response time}
\label{sec:assisted-fuzzy-model-response-times}
Average model response time dropped by about half using the assisted methodology compared to the unAssisted Fuzzy Matching task: 16.1 seconds for Fuzzy Matching, just 8.1 seconds for Assisted Fuzzy. GTP-5 dropped by half while Gemini 2.5 Flash-Lite dropped by 85\% (recall Gemini 2.5 Flash-Lite also had an increase in accuracy of nearly 50 percentage points). Smaller drops were seen for GPT-5 nano and Gemini 2.5 Flash-Lite, those both were among the lowest latency models in the Fuzzy Matching task.

\begin{table}[H]
\centering
\renewcommand{\arraystretch}{1.15}
\begin{tabular}{lccc}
\toprule
\multirow{2}{*}{\textbf{Model}} 
  & \multicolumn{2}{c}{\textbf{Avg Response Time (s)}} & \multirow{2}{*}{\textbf{$\Delta$ (s)}} \\ 
\cmidrule(lr){2-3}
  & \textbf{Fuzzy} & \textbf{Assisted Fuzzy} &  \\
\midrule
Gemini-2.5 Flash                & 7.3   & 5.0   & $-$2.3  \\
Gemini-2.5 Flash-Lite & 20.1  & 3.1   & $-$17.0 \\
Gemini-2.5 Pro                  & 13.8  & 9.8   & $-$4.0  \\
GPT-5                           & 25.5  & 12.4  & $-$13.1 \\
GPT-5 mini                      & 17.7  & 8.0   & $-$9.7  \\
GPT-5 nano                      & 12.3  & 10.1  & $-$2.2  \\
\bottomrule
\end{tabular}
\caption{Average response time in seconds for Fuzzy Matching vs.\ Assisted Fuzzy Matching. 
Assisted Fuzzy consistently reduces latency, with the largest gain (17.0s) for Gemini-2.5 Flash-Lite 
and more modest but still meaningful improvements across other models.}
\label{tab:latency-fuzzy-assisted-seconds}
\end{table}

\subsubsection{Cost per Correct}
\label{sec:cpc-assisted-fuzzy}
The Assisted Fuzzy Marching approach also drastically reduced the cost per correct answer. Averaged across all models, cost per correct answer dropped by more than an order of magnitude, from \$0.0547 to \$0.0045. Reductions were largely uniform across models, with cost of correct reductions between 87\% and 96\%.

This reduction in cost was driven by both input and output cost reductions. On the input size, the number of input tokens was reduced by roughly 99\% as only an average of 20 sentences were sent to the model in the final snippets, as opposed to the full 2{,}772-sentence transcript under the standard Fuzzy matching condition. Output tokens were also reduced due to fewer reasoning tokens being used. The Fuzzy matching task used a mean of 1,230 tokens as opposed to just 810 in the Assisted Fuzzy case. While ``only'' a reduction of around 400 tokens, this is still a substantial contributor to cost reduction at scale as output tokens tend to be around 8 times more expensive than input tokens. This 400 token reduction in output tokens then is equivalent to roughly 3,200 input tokens, or about 70 sentences (using the average of 44 tokens per sentence calculated using our transcript).

\begin{table}[H]
\centering
\renewcommand{\arraystretch}{1.15}
\begin{tabular}{lccc}
\toprule
\multirow{2}{*}{\textbf{Model}} 
  & \multicolumn{2}{c}{\textbf{Cost per Correct (\$)}} & \multirow{2}{*}{\textbf{Reduction (\%)}} \\ 
\cmidrule(lr){2-3}
  & \textbf{Fuzzy} & \textbf{Assisted Fuzzy} &  \\
\midrule
Gemini-2.5 Flash                & 0.0256 & 0.0024 & 91\% \\
Gemini-2.5 Flash-Lite (Preview) & 0.0154 & 0.0006 & 96\% \\
Gemini-2.5 Pro                  & 0.0987 & 0.0127 & 87\% \\
GPT-5                           & 0.1244 & 0.0093 & 93\% \\
GPT-5 mini                      & 0.0240 & 0.0011 & 95\% \\
GPT-5 nano                      & 0.0062 & 0.0005 & 92\% \\
\bottomrule
\end{tabular}
\caption{Cost per correct comparison between fuzzy and Assisted Fuzzy retrieval in the new run. 
All models show dramatic cost efficiency gains, with reductions of 87--96\%.}
\label{tab:new-run-cost-fuzzy-assisted-pct}
\end{table}

Results from the early pilot test were even more extreme, though again the pilot had only 6 total fuzzy sentences.

\begin{table}[H]
\centering
\renewcommand{\arraystretch}{1.15}
\begin{tabular}{lccc}
\toprule
\multirow{2}{*}{\textbf{Model}} 
  & \multicolumn{2}{c}{\textbf{Cost per Correct (\$)}} & \multirow{2}{*}{\textbf{Reduction (\%)}} \\ 
\cmidrule(lr){2-3}
  & \textbf{Fuzzy} & \textbf{Assisted Fuzzy} &  \\
\midrule
Gemini-2.5 Pro                  & 0.0935  & 0.0057 & 93.9\% \\
Gemini-2.5 Flash                & 0.0287  & 0.0011 & 96.2\% \\
GPT-4.1                         & 0.6051  & 0.0088 & 98.5\% \\
Claude Sonnet 4.0               & 0.4337  & 0.0089 & 97.9\% \\
Gemini-2.5 Flash-Lite (Preview) & 0.0350  & 0.0006 & 98.3\% \\
o3                              & 1.3594  & 0.0127 & 99.1\% \\
GPT-4.1 mini                    & 0.4173  & 0.0019 & 99.5\% \\
o4-mini                         & 2.6666  & 0.0095 & 99.6\% \\
Claude 3.5 Haiku                & ---     & 0.0046 & --- \\
Claude 3.7 Sonnet               & ---     & 0.0023 & --- \\
GPT-4.1 nano                     & ---     & 0.0016 & --- \\
o3-mini                         & ---     & 0.0136 & --- \\
\bottomrule
\end{tabular}
\caption{Comparison of cost efficiency (\$/Correct) between Fuzzy Matching and Assisted Fuzzy Matching across models. 
Percentage reduction is relative to fuzzy cost per correct. 
For models with 0\% accuracy in the Fuzzy condition, cost per correct is undefined and left blank.}
\label{tab:pilot-cost-fuzzy-assisted-pct}
\end{table}

\section{Conclusion}
\label{sec:conclusion}

This paper asked a narrow, production‑driven question: given a long, sentence‑timestamped transcript and a short quote that may not appear verbatim, how reliably can contemporary LLMs return the correct \texttt{start\_ms}–\texttt{end\_ms} pair? We answered it in three parts. First, we defined a simple, reusable evaluation—\emph{TimeStampEval}—that makes timestamp prediction the output, not an afterthought. Second, we showed that small, concrete design choices dominate outcomes: put the query \emph{before} the transcript; use a compact text layout that places the sentence \emph{before} its timestamps (Text First Top, TFT); budget a little reasoning for weak layouts; and treat \emph{off‑by‑one} as a distinct, transitional error class rather than noise. Third, we moved beyond exact match to the fuzzier world that production systems actually face, and demonstrated that a lightweight hybrid—cheap fuzzy narrowing followed by LLM verification on a short snippet—yields large accuracy gains while cutting both latency and dollars‑per‑correct by an order of magnitude.

\paragraph{What the measurements say.}
Across models and setups, four empirical regularities stand out.

\begin{enumerate}[leftmargin=1.5em,itemsep=4pt,topsep=2pt]
\item \textbf{Prompt placement and format are first‑order effects.} Moving the query to the top and swapping the native STT JSON for TFT consistently improved exact‑match accuracy (typically +3–20 pp for query placement; +1–15 pp for TFT) while trimming 30–42\% of prompt tokens. These same changes convert many near‑misses into exact hits and shave seconds from response time (\S\ref{sec:accuracy-improv-tests}, \S\ref{sec:timestamps}).
\item \textbf{Off‑by‑one is a real regime, not just error bars.} Under harder layouts—especially JSON with query‑after—mistakes organize themselves: \emph{miss} $\rightarrow$ \emph{off‑by‑one} $\rightarrow$ \emph{exact} as model scale rises (\S\ref{sec:off-by-one-error-description}, Fig.~\ref{fig:offbyone-transition}). Under TFT with query‑before, the off‑by‑one mass largely collapses to exact.
\item \textbf{Reasoning budgets rescue weak prompts and nudge strong ones to ceiling.} With a few hundred “thinking” tokens, the worst configuration jumped from $\sim$37\% to $\sim$72–77\% exact; the best moved from high‑80s to mid‑90s (\S\ref{sec:thinking-budget}). Because models self‑regulate and typically use $\sim$600–850 tokens, the marginal cost is negligible.
\item \textbf{Fuzziness is manageable with a hybrid.} Unassisted fuzzy retrieval lags exact by $\sim$4–10 pp on average. Pre‑narrowing the haystack (RapidFuzz) and sending only a short snippet to the model adds +4 to +50 pp (larger in the pilot), halves latency, and cuts cost‑per‑correct by $\ge!10\times$ (\S\ref{sec:fuzzy}). In ablations where we feed \emph{exact} snippets, even small models go near‑ceiling—evidence that input length, not task semantics, is the dominant difficulty.
\end{enumerate}

Other patterns are consistent with prior long‑context work but sharpened here for timestamp retrieval: an inverted‑U by span length (2–5 easiest), last‑third penalties in the full run (attenuated in the multi‑transcript sweep), and robustness through $\sim$400k tokens with gradual tapering nearer a million (\S\ref{sec:accuracy-improv-tests}). Provider differences largely express as a speed–accuracy trade: Gemini families are faster and scale latency more smoothly; OpenAI models run slower but averaged higher fuzzy accuracy in our full run. Finally, absent‑target rejection worked well (Table~\ref{tab:null-target-first-by-length}), a practical guardrail when the quote simply is not there.

\paragraph{What to use in practice.}
The measurements reduce to a short recipe you can ship:

\begin{enumerate}[leftmargin=1.5em,itemsep=3pt,topsep=2pt]
\item \textbf{If the quote is verbatim,} don’t use an LLM. Deterministic methods are cheaper, faster, and perfectly accurate (\S\ref{sec:timestamps}). Keep exact‑match benchmarking around only to probe layout and placement effects.
\item \textbf{If the quote is fuzzy (real life),} run a two‑stage hybrid:
\begin{enumerate}[label=(\alph*), leftmargin=1.25em, align=left]
  \item RapidFuzz (or equivalent) on a sentence-only view to produce a small set of candidate snippets with dynamic padding.
  \item LLM verification over \emph{just} those snippets to return \texttt{start\_ms}/\texttt{end\_ms}.
\end{enumerate}
Escalate only on low confidence or candidate disagreement (small $\rightarrow$ base $\rightarrow$ large model).
\item \textbf{Always structure prompts for retrieval:} query‑before, TFT, and a small reasoning budget for hard layouts. If latency matters, prefer TFT; if budgets are tight, lean on the hybrid to avoid sending the full transcript (~99% fewer sentences in our runs).
\item \textbf{Interpret errors operationally.} Decide upfront whether off‑by‑one is “good enough” for your downstream (clipper, player, audit). Use adjusted accuracy for planning and strict accuracy for automation thresholds.
\item \textbf{Treat caching as opportunistic.} Provider prefix caches can make query‑after \emph{appear} cheaper when the fixed transcript sits at the top, but cache hits are not portable across models and are not guaranteed (\S\ref{sec:caching}, Fig.~\ref{fig:scatter-cost-exact-match}).
\end{enumerate}

\paragraph{Conceptual contributions.}
Beyond engineering guidance, two small conceptual points may travel:

\begin{itemize}[leftmargin=1.25em,itemsep=3pt,topsep=2pt]
\item \textbf{Timestamp retrieval is a distinct long‑context task.} Prior “needle” setups stop at text recall; many production systems need \emph{temporal} coordinates. Treating timestamps as the output surfaces position‑sensitive errors (off‑by‑one) that ordinary QA metrics hide.
\item \textbf{Layout is a capability amplifier.} Swapping JSON for TFT did more for accuracy, cost, and latency than switching among several strong models. Long‑context behavior is not only about model scaling—it is about making traversal cheap and alignment unambiguous.
\end{itemize}

\paragraph{Limitations and scope.}
Our main full run used one House‑day transcript for cross‑model comparability; we then ran a ten‑transcript sweep (House and Senate, 1989–2025, 44k–860k tokens) to put error bars around the TFT + reasoning recipe (\S\ref{sec:gemini-flash-confidence-intervals}). Still, results may shift with other STT engines, languages, or domains heavy in numerics and acronyms. Our fuzzy perturbations are synthetic but designed to mimic common editorial and STT drift; real‑world noise can be harsher. We also measured at sentence granularity; word‑level alignment (e.g., forced alignment) is a different regime with its own tools. Finally, observed provider costs reflect contemporary pricing and imperfect, provider‑specific caching; these are moving targets.

\paragraph{Outlook.}
Three directions feel immediately useful. First, \textbf{robust fuzzing} grounded in real paired corpora (official text vs.\ spoken transcripts) would pressure‑test beyond synthetic edits. Second, \textbf{uncertainty signals}—confidence from both the fuzzy stage and the LLM—would enable principled escalation and batch‑time budgeting. Third, \textbf{format‑aware training} (or lightweight adapters) that natively “speak” TFT could compress prompts further and suppress off‑by‑one without extra thinking tokens.

\medskip
\noindent
In short, the timestamp problem is solvable with simple parts. If passages are verbatim, don’t reach for a model. If they aren’t, shrink the haystack first, then ask the model to be precise. Put the question before the data, put the text before the timestamps, and give the model a little space to think when the layout fights you. The result is a system that is fast, cheap, and—by the metrics that matter to downstream tools—accurate enough to trust.

\bibliographystyle{plainnat}
\bibliography{references}
\end{document}